\documentclass[compsoc,conference,a4paper,10pt,times]{IEEEtran}

\IEEEoverridecommandlockouts

\usepackage{cite}
\usepackage{amsmath,amssymb,amsfonts}
\usepackage{dblfloatfix}
\usepackage{graphicx}
\usepackage{textcomp}
\usepackage{xcolor}
\usepackage[colorlinks=true,urlcolor=black]{hyperref}

\usepackage{algpseudocode}

\usepackage{booktabs}  
\usepackage{multirow}
\usepackage{siunitx}   
\usepackage{amssymb}
\usepackage{amsthm}

\newcounter{MYtempeqncnt}

\ifCLASSOPTIONcompsoc
  \usepackage[caption=false,font=normalsize,labelfont=rm,textfont=rm]{subfig}%
\else
  \usepackage[caption=false,font=footnotesize]{subfig}%
\fi

\usepackage{float}      
\usepackage{placeins}   
\usepackage{afterpage}
\usepackage{needspace}

\usepackage{enumitem}   

\usepackage{tikz}
\usepackage{xfp}
\begin{document}

\title{Auditing Fairness-Privacy Trade-offs: Subpopulation-Level Effects of Fairness-Enhancing Algorithms}

\author{%
\hspace*{\fill}%
\begin{minipage}[t]{0.30\textwidth}\centering
Umid Suleymanov*\\
\textit{Virginia Tech}\\
Blacksburg, USA\\
umids@vt.edu
\end{minipage}%
\hspace*{\fill}%
\begin{minipage}[t]{0.32\textwidth}\centering
Ilhama Novruzova*\\
\textit{ADA University}\\
Baku, Azerbaijan\\
inovruzova16235@ada.edu.az
\end{minipage}%
\hspace*{\fill}%
\begin{minipage}[t]{0.34\textwidth}\centering
Khalid Mammadov\\
\textit{University of Potsdam}\\
Potsdam, Germany\\
khalid.mammadov@uni-potsdam.de
\end{minipage}%
\hspace*{\fill}%
\\[2ex]
\hspace*{\fill}%
\begin{minipage}[t]{0.34\textwidth}\centering
Natavan Hasanova\\
\textit{University of Passau}\\
Passau, Germany\\
hasano04@ads.uni-passau.de
\end{minipage}%
\hspace*{\fill}%
\begin{minipage}[t]{0.30\textwidth}\centering
Murat Kantarcioglu\\
\textit{Virginia Tech}\\
Blacksburg, USA\\
muratk@vt.edu
\end{minipage}%
\hspace*{\fill}%
}


\maketitle
\begin{abstract}
Machine learning (ML) models deployed in sensitive domains such as healthcare, law enforcement, and finance must satisfy not only utility requirements but also fairness and privacy guarantees. While prior work has largely examined how privacy-preserving techniques affect fairness, the inverse question-\emph{how fairness-enhancing algorithms influence privacy leakage}-remains underexplored. We present the first comprehensive study of how fairness interventions affect membership inference privacy risks at the subpopulation level. By adapting the Likelihood Ratio Attack (LiRA) for subgroup auditing, we uncover privacy disparities that aggregate evaluations obscure. We further analyze how Differential Privacy (DP) interacts with fairness-enhancing methods across different categories, showing that DP's privacy benefits and utility costs are \emph{unevenly distributed across subpopulations}. Our results demonstrate that fairness interventions do not uniformly increase privacy risk; their impact depends on model architecture, subgroup size, and mitigation strategy. These findings reveal that fairness, privacy, and utility must be \textbf{jointly evaluated at the subpopulation level}, and we introduce the first unified empirical framework to support such auditing in practice.
\end{abstract}


\section{Introduction}

ML models are increasingly deployed in sensitive domains such as healthcare, law enforcement, and finance, where addressing fairness and privacy concerns is essential. These models often rely on datasets that are not only sensitive but also unequally distributed across subpopulations, intensifying both fairness and privacy challenges~\cite{murakonda2020mlprivacymeteraiding, 9014384}. Yet, these concerns are typically studied at the \emph{aggregate} level, obscuring disparities across subpopulations. In practice, risks are rarely uniform: minority or less-represented groups may experience disproportionately poor fairness, weaker privacy protections, or severe utility losses. This motivates a shift from aggregate evaluation toward \emph{subpopulation-level auditing}.

Prior work has primarily examined how privacy-preserving mechanisms affect fairness, with particular focus on Differential Privacy (DP)~\cite{tran2022differentiallyempiricalriskminimization,fioretto2024decisionmakingdifferentialprivacy,makhlouf2024systematicformalstudyimpact,mangold2023differentialprivacyboundedimpact}. These studies ask whether DP amplifies or reduces group disparities under standard training pipelines. In contrast, the inverse question, \textbf{how fairness-enhancing algorithms influence privacy risk}, remains largely underexplored. Widely used fairness interventions span three categories: (i) \textit{pre-processing} methods such as Disparate Impact Remover (DIR)~\cite{feldman2015certifying} and Reweighing (REW)~\cite{kamiran2012data}; (ii) \textit{in-processing} techniques such as Exponentiated Gradient Reduction (EGR)~\cite{pmlr-v80-agarwal18a}; and (iii) \textit{post-processing} approaches such as Calibrated Equalized Odds (CPP)~\cite{pleiss2017fairness}. Despite their widespread adoption, we do not yet understand whether these interventions \emph{raise or reduce privacy leakage}, nor how these effects vary across demographic subgroups.

In this work, we fill this gap by conducting the first comprehensive analysis of how fairness-enhancing methods impact membership inference risk at the subpopulation level. We evaluate three state-of-the-art membership inference attacks (MIAs), including the Likelihood Ratio Attack (LiRA)~\cite{carlini2022membership}, and extend LiRA to produce subgroup-specific leakage estimates. We further analyze how Differential Privacy interacts with fairness interventions across mitigation categories. Our findings show that while DP reliably reduces privacy risk, its utility costs are \emph{highly uneven across subgroups}: some retain acceptable performance, while others, often underrepresented groups, collapse to near-random accuracy.

Motivated by the lack of evidence regarding how fairness-enhancing interventions influence privacy leakage, we pose the following research questions:

\begin{itemize}
    \item \textbf{RQ1:} How do fairness-enhancing algorithms affect membership inference risk and utility across subpopulations?
    \item \textbf{RQ2:} How does Differential Privacy, alone or combined with fairness methods, alter the privacy--fairness--utility trade-off at the subgroup level?
    \item \textbf{RQ3:} How do model architecture and subgroup representation shape these interactions?
\end{itemize}

Our results demonstrate that \textit{fairness and privacy are not inherently in conflict}. Rather, their interaction depends on factors such as model architecture, the prevalence of subpopulations in the training data, and the choice of fairness technique. By illuminating these dependencies, our work promotes a more nuanced understanding of the trade-offs between fairness enhancing techniques and their privacy implications in ML systems.

Our experiments span ten variants of six widely used datasets (plus a synthetic benchmark) in the fairness and privacy domains. To quantify privacy risks, we employ state-of-the-art membership inference attacks under both black-box and white-box threat models. The main contributions of this paper are summarized as follows:

\begin{itemize}[topsep=0pt, partopsep=0pt]
    \item We provide the first systematic audit of how pre-, in-, and post-processing fairness algorithms affect subgroup-level privacy risk and utility.
    \item We extend the Likelihood Ratio Attack (LiRA)~\cite{carlini2022membership} to enable subgroup-specific privacy auditing, allowing fine-grained measurement of membership inference vulnerability across demographic groups.
    \item We systematically characterize how Differential Privacy (DP) interacts with fairness interventions, revealing uneven privacy–utility trade-offs across models and subpopulations.
    \item We uncover that fairness and privacy are not inherently conflicting objectives; their interaction depends critically on model architecture, subgroup representation, and the chosen fairness mitigation strategy.
\end{itemize}

\section{Related Work}

\textbf{Disparate vulnerability and subgroup privacy risk.} Average-based generalization metrics can mask large differences across protected subgroups. To address this, researchers have proposed subgroup-level notions of generalization and disparate vulnerability. Kulynych et al.~\cite{KulynychYCVT22} introduce a subgroup-specific distributional generalization gap, $R(\pi,d)\triangleq d(\mu_1^\pi,\mu_0^\pi)$, and show how subgroup shifts can affect vulnerability to MIAs. Empirical studies likewise report that underrepresented groups are often more susceptible to membership inference~\cite{9014384,zhong2022understanding}, motivating audits at the subpopulation level rather than aggregate reporting.

\textbf{How Differential Privacy alters fairness and utility.} A large body of work investigates how Differential Privacy (DP) training affects fairness outcomes and model utility. Several studies find that DP-SGD can amplify disparities in accuracy and fairness for underrepresented groups~\cite{NEURIPS2019_fc0de4e0}. Jagielski et al.~\cite{pmlr-v97-jagielski19a} provide one of the first formal treatments of the joint design of fairness and Differential Privacy, proposing algorithms that satisfy both constraints and identifying fundamental trade-offs between privacy, accuracy, and fairness. More recent work reveals nuanced effects: Hansen et al.~\cite{hansen-etal-2024-impact} show that DP can sometimes reduce disparities when combined with group-robust optimization. Makhlouf et al.~\cite{makhlouf2024systematicformalstudyimpact} analyze local DP and derive theoretical bounds for fairness metrics under randomized response, while Mangold et al.~\cite{mangold2023differentialprivacyboundedimpact} prove that deviations in fairness measures under central DP diminish with larger margins. Tran et al.~\cite{tran2022differentiallyempiricalriskminimization,fioretto2024decisionmakingdifferentialprivacy} empirically assess DP’s impact on group-level utility and provide mitigation guidelines. Bullwinkel et al.~\cite{bullwinkel2022evaluating} extend this discussion to DP synthetic data generation, showing that privacy-preserving synthesis can maintain overall accuracy but often worsens fairness for minority groups. Collectively, these works focus on how DP affects fairness and utility, but \textit{they do not directly examine how fairness-enhancing interventions, in turn, influence privacy leakage.}

\textbf{Fairness interventions and privacy leakage: an open question.} 
Prior research often treats fairness and privacy as a strict trade-off~\cite{chang2021privacyrisksalgorithmicfairness, fioretto2022survey}, with improvements in one presumed to harm the other. However, many existing evaluations remain narrow in scope, typically limited to a single model family or a restricted set of fairness methods (often in-processing), and they do not consider the interactions of fairness-enhancing algorithms with Differential Privacy (DP). For example, Chang and Shokri~\cite{chang2021privacyrisksalgorithmicfairness} and more recently Tian et al.~\cite{inproceedings} study membership inference risks in fairness-enhanced models, but focus on specific mitigation strategies and do not analyze interactions with DP or multiple fairness mitigation categories. As a result, the combined effect of fairness techniques, privacy protections, and model choice across subpopulations remains poorly understood. In this paper, we address this gap through a broad, systematic study. 

\textbf{Gap and our focus.} In summary, while there is substantial research on (i) disparate privacy vulnerability across subgroups, (ii) fairness and utility under DP, there remains a  gap: \textbf{how widely used fairness-enhancing algorithms (pre-, in-, and post-processing) affect empirical privacy leakage at the subpopulation level, and how these effects interact with Differential Privacy.} In this paper, we fill this gap by (a) adapting modern membership inference attacks (LiRA, OQTA, OTA) for subgroup auditing, (b) evaluating a broad set of fairness mitigators (DIR, REW, EGR, CPP, SYN) across multiple model families and datasets, and (c) quantifying the joint fairness–privacy–utility trade-offs at subpopulation granularity.

\begin{figure*}[!t]
\normalsize
\setcounter{MYtempeqncnt}{\value{equation}}
\setcounter{equation}{1} 

\begin{equation}
\label{eq:stat_par_diff}
\text{stat\_par\_diff} = \Pr(\hat{Y} = 1 \mid \text{unprivileged}) - \Pr(\hat{Y} = 1 \mid \text{privileged})
\end{equation}

\begin{equation}
\label{eq:disp_imp}
\text{disp\_imp} = 1 - \min\Bigl(\mathcal{R}, \frac{1}{\mathcal{R}}\Bigr), \text{ where}
\end{equation}

\begin{equation}
\label{eq:ratio_r}
\mathcal{R} = \frac{\Pr(\hat{Y} = 1 \mid \text{unprivileged})}{\Pr(\hat{Y} = 1 \mid \text{privileged})}
\end{equation}

\begin{equation}
\label{eq:avg_odds_diff}
\text{avg\_odds\_diff} = \frac{1}{2} \Big[ \big| \text{FPR}_{\text{unpriv}} - \text{FPR}_{\text{priv}} \big| + \big| \text{TPR}_{\text{unpriv}} - \text{TPR}_{\text{priv}} \big| \Big]
\end{equation}

\setcounter{equation}{\value{MYtempeqncnt}}
\hrulefill
\vspace*{4pt}
\end{figure*}

\section{Background}

\subsection{Membership Inference Attack}
MIAs pose a significant privacy threat in machine learning. These attacks seek to identify whether a specific data point was included in the training set used to build a model. If successful, such attacks can expose sensitive or personal information about individuals in the dataset, raising serious privacy concerns. In a typical MIA scenario, the attacker has black-box access to the trained model, meaning they can provide inputs and observe the corresponding outputs (e.g., predicted labels or probabilities). By carefully analyzing the model's responses, the attacker attempts to infer whether a given data point was used during training.

Formally, let \( \Omega \) represent the overall population of possible data points, and \( D \) be the data-generating distribution over \( \Omega \). A training set \( S \subset \Omega \) of size \( n \) is drawn from \( D \), and a learning algorithm \( A(\cdot) \) produces the model \( A(S) \). Given a challenge example \( x \in \Omega \), the attacker’s objective is to decide if \( x \in S \). To achieve this, the attacker designs a membership inference function:
\begin{equation}
\hat{m} = \mathcal{A}\bigl(A(S), x\bigr),
\end{equation}
where \( \hat{m} \in \{0, 1\} \) is the membership prediction (\( \hat{m} = 1 \) indicates that \( x \) is inferred to be in \( S \), while \( \hat{m} = 0 \) implies otherwise).

The attack leverages the intuition that models often behave differently for data points they have seen during training (members) compared to those they have not (non-members). For example, models may assign higher confidence scores to training points, making it easier for an attacker to distinguish members from non-members.

These attacks pose serious privacy risks, particularly when dealing with sensitive datasets such as medical records, financial information, or social data. By successfully guessing membership, an attacker can infer private details about individuals, emphasizing the need for robust defenses against MIAs.

\subsection{Fairness}
Fairness in machine learning seeks to ensure that models treat different groups without causing undue benefit or harm based on sensitive attributes (e.g., gender, race, age). In our approach, we assess fairness by measuring performance gaps or differences in outcomes across subpopulations defined by these attributes. Specifically, the data is analyzed by identifying privileged and unprivileged groups based on the sensitive attribute under consideration. Privileged group consists of individuals who receive systematically better outcomes or treatment from the model while the unprivileged group includes individuals who are more likely to face unfavorable outcomes or treatment from the model. By distinguishing between these groups, we can quantify and address disparities in model behavior, ensuring that outcomes are distributed equitably across all subpopulations. Our analysis uses several fairness metrics to evaluate these disparities:

\begin{itemize}
    \item \textbf{Balanced Accuracy:} 
    \addtocounter{equation}{4} 
    \begin{equation}
    \text{bal\_acc} = \frac{1}{2}\Bigl(\text{TPR} + \text{TNR}\Bigr),
    \end{equation}
    where $\text{TPR}$ is the true positive rate and $\text{TNR}$ is the true negative rate. This metric captures how well the model classifies both positive and negative instances across all groups.
    
    \item \textbf{Statistical Parity Difference:} Shown in (\ref{eq:stat_par_diff}), where $\hat{Y}$ denotes the model's prediction, a value of $0$ indicates equal rates of positive prediction for both groups.
    
    \item \textbf{Disparate Impact:} Formulated as in (\ref{eq:disp_imp}) using the ratio $\mathcal{R}$ from (\ref{eq:ratio_r}). If $\mathcal{R}$ is close to $1$, the model assigns positive outcomes at similar rates to both groups, indicating low disparate impact. Our definition of $\text{disp\_imp}$ transforms this ratio such that lower values correspond to better fairness, with $\text{disp\_imp} = 0$ representing perfect parity.

    \item \textbf{Average Odds Difference:} Defined in (\ref{eq:avg_odds_diff}), where $\text{FPR}$ is the false positive rate, this measures how differently the model classifies positives and negatives across privileged and unprivileged groups.
    
    \item \textbf{Equal Opportunity Difference:}
    \begin{equation}
    \text{eq\_opp\_diff} = \text{TPR}_{\text{unpriv}} - \text{TPR}_{\text{priv}},
    \end{equation}
    evaluating the gap in true positive rates between groups.
\end{itemize}
\subsection{Subpopulation Privacy}

Subpopulation privacy examines how much membership information is leaked for specific subgroups under membership inference attacks. If certain subgroups have higher privacy risk values than others, it indicates that membership information is easier to detect for those subgroups, a phenomenon referred to as \emph{disparate vulnerability}. This disparity can be quantified by the difference between the highest and lowest privacy risk values across all subgroups:
\begin{equation}
\max_{j} \left (\text{PrivacyRisk}(G_j) \right )\;-\; \min_{j} \left ( \text{PrivacyRisk}(G_j) \right ).
\end{equation}
A large gap suggests that the model leaks membership information more readily for some subgroups, which may arise due to imbalanced training sets, model overfitting on certain groups, or fairness constraints that unintentionally expose membership data about underrepresented groups.

To assess privacy risk at the subgroup level, we compute per-example losses $\ell_i$ for both training and test data, analyzing how well an attacker can distinguish group members from non-members. For a subgroup \( G_j \), we define:
\begin{equation}
\begin{aligned}
\text{TPR}_{G_j} &= \Pr(\text{Attack says ``member''} \mid x \in G_j \cap \text{train}), \\
\text{TNR}_{G_j} &= \Pr(\text{Attack says ``non-member''} \mid x \in G_j \cap \text{test}).
\end{aligned}
\end{equation}
The subgroup-level privacy risk is given by:
\begin{equation}
\text{PrivacyRisk}(G_j) = \frac{\text{TPR}_{G_j} + \text{TNR}_{G_j}}{2}.
\end{equation}

A higher \(\text{PrivacyRisk}(G_j)\) indicates that membership inference is more successful for that subgroup, implying greater privacy exposure. This analysis helps identify whether certain fairness strategies inadvertently increase inference accuracy in underrepresented groups and provides insights into mitigating such risks.

\subsection{Fairness-Enhancing Algorithms}
\label{sec:mitigators}

In this work, we evaluate five fairness mitigators to assess their impact on both fairness and privacy: Synthetic Mitigator (SYN), Disparate Impact Remover (DIR), and Reweighing (REW) as pre-processing techniques; Exponentiated Gradient Reduction (EGR) as an in-processing method; and Calibrated Equalized Odds (CPP) as a post-processing approach. Each mitigator addresses fairness through a distinct mechanism, which may, in turn, influence privacy risk.

Regarding pre-processing techniques, SYN generates synthetic data to balance favorable outcome distributions across privileged and unprivileged groups~\cite{doi:10.1137/1.9781611977653.ch98, 10.5555/1622407.1622416}. It modifies the training dataset through oversampling or undersampling to equalize positive base rates. In our implementation, the transformed dataset is used only for training the target model and is excluded from the attack setup (i.e., only real data is used in privacy evaluation). This setup ensures a realistic assessment of privacy risk. DIR modifies feature values to reduce the correlation between sensitive attributes and predictions, aiming to lower disparate impact while maintaining data utility~\cite{feldman2015certifying}. In our setup, DIR is applied to both training and testing data for all models, including both shadow and target models used in the LiRA attack framework (Section~\ref{subsec:lira-implementation}). REW assigns weights to training samples based on group membership and class labels, enabling the model to learn from all subgroups in a balanced way~\cite{kamiran2012data}. In our experiments, REW is applied only to the training data; test datasets remain unchanged. This mitigator addresses fairness without altering the dataset itself.

Moving to the in-processing method, EGR enforces fairness constraints-such as demographic parity or equal opportunity-during model training~\cite{pmlr-v80-agarwal18a}. It operates by iteratively optimizing a constrained loss function. In our implementation, EGR is applied directly during model training with a learning rate set to 0.001. Finally, for the post-processing approach, CPP adjusts predicted probabilities to satisfy fairness criteria like equalized odds~\cite{pleiss2017fairness}. This method is particularly useful for high-accuracy models, as it modifies output probabilities without retraining. In our experiments, the CPP mitigator is applied to model predictions, calibrating probabilities for each subgroup to meet fairness constraints while leaving the underlying model unchanged.

\section{Membership Inference Attacks for Subpopulations}
\label{sec:mia_attacks}
In this section, we describe the attack setups used to evaluate the privacy risks associated with fairness algorithms across subpopulations. 
We examine three membership inference attacks: the \textbf{Optimal-Quantile-Threshold Attack (OQTA)}~\cite{chang2021privacyrisksalgorithmicfairness}, the \textbf{Optimal-Threshold Attack (OTA)}~\cite{KulynychYCVT22,8429311}, and the \textbf{LiRA}~\cite{carlini2022membership}. These attacks are adapted to measure privacy leakage under different fairness algorithms. We specifically chose OQTA, OTA, and LiRA because of their methodological differences e.g. determining thresholds based on population dataset and shadow models, suitability for subgroup-specific analysis, and relevance to real-world adversarial scenarios.

We consider subpopulation privacy risk as an additional dimension of fairness. To address this, we provide a detailed description of the attack setups and their adaptations for subpopulations. This evaluation framework allows the privacy leakage of fairness algorithms to be measured and reported alongside other fairness metrics.

\subsection{Likelihood Ratio Attack Implementation}
\label{subsec:lira-implementation}

In the original LiRA attack~\cite{carlini2022membership}, the \emph{in}- and \emph{out}-distributions are sampled uniformly from the entire training and reference datasets. In contrast, our modified LiRA restricts both distributions to specific fairness-relevant subpopulations. For a given subgroup, both the target sample and its reference comparison sets are drawn exclusively from that subgroup. This adaptation reveals how membership inference vulnerabilities vary across groups, providing a finer granularity necessary for auditing fairness-privacy trade-offs.

\textbf{Group-Specific Likelihood Ratio Attack.} 
For each subgroup \(g\), we filter the dataset to include only samples belonging to that subgroup and perform the following steps:
\begin{enumerate}[noitemsep]
    \item Extract the confidence scores \(\phi_{\text{in}}^g\) and \(\phi_{\text{out}}^g\) for the shadow models trained on \(\mathcal{D}_{\text{in}}\).
    \item Compute the mean and variance of the subgroup-specific in- and out-distributions:
    \begin{equation}
    \mu_{\text{in}}^g = \text{mean}(\phi_{\text{in}}^g), \quad \sigma_{\text{in}}^{g^2} = \text{var}(\phi_{\text{in}}^g),
    \end{equation}
    \begin{equation}
    \mu_{\text{out}}^g = \text{mean}(\phi_{\text{out}}^g), \quad \sigma_{\text{out}}^{g^2} = \text{var}(\phi_{\text{out}}^g).
    \end{equation}
    \item Compute the likelihood ratio for the subgroup \(g\):
    \begin{equation}
    \Lambda^g = \frac{p(\phi_{\text{obs}}^g \mid \mathcal{N}(\mu_{\text{in}}^g, \sigma_{\text{in}}^{g^2}))}{p(\phi_{\text{obs}}^g \mid \mathcal{N}(\mu_{\text{out}}^g, \sigma_{\text{out}}^{g^2}))}.
    \end{equation}
\end{enumerate}

\textbf{Robustness for Small Subpopulations.} 
Estimating Gaussian parameters for small subgroups (e.g., in the German Credit dataset, $N=1{,}000$) can introduce statistical instability. We address this through two primary safeguards. First, all reported metrics are averaged over 20 independent experimental runs; as shown in our results  (e.g., Figures~\ref{fig:dt_lira_mia}--\ref{fig:nn_lira_mia_part2}), the standard deviations (represented by error bars) remain tightly bounded even for minority subgroups, indicating that the identified disparate vulnerability is a stable characteristic rather than a statistical artifact.  Second, we cross-validate our findings against the non-parametric OQTA and OTA attacks. Because these methodologies do not rely on Gaussian parameter estimation yet yield consistent leakage patterns across all models (e.g., Random Forests), we confirm that our subpopulation-level LiRA adaptation is robust to the constraints of small-scale data.

\medskip
We implemented the LiRA attack using the TensorFlow Privacy library to assess the membership privacy risks of models trained on sensitive datasets.

\subsection{Optimal-Quantile-Threshold Attack}
\label{subsec:optimal-quantile-threshold-attack}

The \textbf{OQTA} is a threshold-based membership inference attack inspired by the (OTA)~\cite{KulynychYCVT22,8429311}, with a key difference: it selects per-subgroup thresholds by sweeping quantiles over the loss distribution of a separate population dataset. This quantile-based strategy is based on the implementation in Chang and Shokri~\cite{chang2021privacyrisksalgorithmicfairness}, and we adopt their setup to ensure reproduction of their results.

We use Decision Tree classifier, Random Forest classifier, Neural Network, Differentially Private Random Forest, or Differentially Private Neural Network as the target model. The original dataset is split into three disjoint parts: training set, test set (together forming the TARGET\_MEMBER and TARGET\_NON\_MEMBER sets, respectively), and a larger population dataset (REFERENCE\_MEMBER). Loss values are computed on all samples using the target model.

To perform inference, we define subgroup-specific thresholds \(\tau^{(g,y)}\), where \(g\) denotes group membership (e.g., privileged/unprivileged) and \(y\) the true label (favorable/unfavorable). We compute a range of thresholds by sweeping quantiles (logarithmically spaced between \(10^{-5}\) and 1) over the population loss distribution, and select the threshold that maximizes balanced accuracy on the target dataset.

Membership is then inferred as follows: if the model's loss on a sample \(z = (x, g, y)\) is below the threshold \(\tau^{(g,y)}\), it is labeled as a member; otherwise, a non-member. We use cross-entropy loss with clipping for numerical stability. Subgroups (\(G_0^-\): unprivileged unfavorable, \(G_0^+\): unprivileged favorable, \(G_1^-\): privileged unfavorable, \(G_1^+\): privileged favorable) are defined by sensitive attribute and true label (see figures and Appendix~\ref{appendix:oqta_algorithm} for full details). ROC curves are available in the code.
\subsection{Optimal-Threshold Attack}
\label{subsec:optimal-threshold-attack}

In our setting, we evaluate the membership inference (MI) vulnerability of different subpopulations by comparing the model’s per-example losses on training and test points, and then applying a threshold-based attack. Suppose we have:
\begin{itemize}
    \item $\boldsymbol{\ell}_{\text{train}} = (\ell_{x_1}, \ldots, \ell_{x_{n_{\text{train}}}})$, the losses for $n_{\text{train}}$ training examples,
    \item $\boldsymbol{\ell}_{\text{test}} = (\ell_{x'_1}, \ldots, \ell_{x'_{n_{\text{test}}}})$, the losses for $n_{\text{test}}$ test examples.
\end{itemize}
(Here, a smaller loss often indicates the point was likely in the training set.)

We pick a threshold $c$ and label examples with loss $\le c$ as ``\emph{member}'' (i.e., from the training data) and those with loss $> c$ as ``\emph{non-member}.'' One way to find $c$ is to scan possible thresholds to maximize a measure of attack success, such as the overall balanced accuracy:
\begin{equation}
\max_c \; \frac{1}{2}
\biggl(\frac{\sum_{x \in \text{train}} \mathbf{1}\{\ell_x \le c\}}{n_{\text{train}}}
\;+\;
\frac{\sum_{x' \in \text{test}} \mathbf{1}\{\ell_{x'} > c\}}{n_{\text{test}}}
\biggr).
\end{equation}
Here, $\mathbf{1}\{\cdot\}$ is the indicator function. Once the threshold is chosen, we define:
\begin{equation}
\text{TPR} 
= \frac{\sum_{x \in \text{train}} \mathbf{1}\{\ell_x \le c\}}{n_{\text{train}}}, 
\quad
\text{TNR} 
= \frac{\sum_{x' \in \text{test}} \mathbf{1}\{\ell_{x'} > c\}}{n_{\text{test}}}.
\end{equation}

\textbf{Subpopulation Analysis} To see how privacy differs across subgroups, we label each data point with a subgroup identifier (e.g., a protected attribute) and a class label. Let $G_j$ represent a particular subgroup of the training set. Then we compute:
$\boldsymbol{\ell}_{\text{train}, G_j} 
\quad\text{and}\quad 
\boldsymbol{\ell}_{\text{test}, G_j},$
the training and test losses for that subgroup. We apply the threshold-based approach within each subgroup:
\begin{align}
\text{TPR}_{G_j} 
&= \frac{\sum_{x \in G_j} \mathbf{1}\{\ell_x \le c_j\}}{|G_j|}, \nonumber \\
\text{TNR}_{G_j} 
&= \frac{\sum_{x' \in \text{non-}G_j} \mathbf{1}\{\ell_{x'} > c_j\}}{|\text{non-}G_j|}
\label{eq:tpr_tnr}
\end{align}
for a threshold $c_j$ chosen to optimize the attack over subgroup $G_j$. By comparing $\text{TPR}_{G_j}$ and $\text{TNR}_{G_j}$ across subgroups, we identify which subpopulations are at higher risk of membership exposure. Understanding this is crucial, as certain fairness interventions or data imbalances can lead to increased memorization for specific subgroups, potentially exposing more information about their membership status. Additionally, ROC curves are available in the code.

\section{Datasets and Models}
Our study focuses on tabular data as it remains the primary format in high-stakes domains, such as healthcare, finance, and criminal justice where auditing fairness-privacy trade-offs is most critical~\cite{8843908, grinsztajn2022tree}. We evaluate our framework on ten versions of six real-world datasets and one synthetic dataset.  
Each dataset is partitioned into subpopulations defined by (i) a binary sensitive attribute (privileged vs.\ unprivileged) and (ii) a binary label (favorable vs.\ unfavorable), following standard fairness literature. Table~\ref{tab:datasets} summarizes all datasets, subgroup definitions, and sample sizes.

\begin{table*}[t]
\caption{Summary of datasets used in this study. Each dataset is binarized into four subpopulations based on (privileged/unprivileged) $\times$ (favorable/unfavorable). Synthetic data enables controlled mechanism analysis; real datasets provide external validity across diverse domains.}
\label{tab:datasets}
\centering
\small
\begin{tabular}{lcccc}
\toprule
\textbf{Dataset} & \textbf{\#Samples} & \textbf{\#Features} & \textbf{Protected Attribute} & \textbf{Privileged Group} \\
\midrule
\textbf{Synthetic} & 2{,}500 & 2 &
$G \in \{0,1\}$ &
$G{=}1$ (80\%) \\
\textbf{Bank Marketing} & 30{,}448 & 57 &
Age & Age $\ge 25$ \\
\textbf{COMPAS (Race)} & 6{,}172 & 11 &
Race & Caucasian \\
\textbf{COMPAS (Gender)} & 6{,}172 & 11 &
Gender & Female \\
\textbf{Law School Admissions (Race)} & 20{,}798 & 13 &
Race & White \\
\textbf{Law School Admissions (Gender)} & 20{,}798 & 13 &
Gender & Male \\
\textbf{German Credit (Age)} & 1{,}000 & 57 &
Age & Age $>25$ \\
\textbf{German Credit (Sex)} & 1{,}000 & 57 &
Sex & Male \\
\textbf{Law School GPA} & 22{,}342 & 3 &
Race & White \\
\textbf{MEPS~19} & 15{,}830 & 138 &
Race & White \\
\bottomrule
\end{tabular}
\end{table*}

\textbf{Real-world Datasets.} We use six benchmark datasets from AIF360~\cite{8843908}: \textbf{Bank Marketing} (age as protected attribute; 57 features, 30,448 samples), \textbf{COMPAS} (race/gender; 11 features, 6,172 samples), \textbf{Law School Admissions}\footnote{From https://github.com/jjgold012/lab-project-fairness (Bechavod and Ligett, 2017). Preprocessed by dropping duplicates, null values, and highly correlated features ($>0.95$).} (race/gender; 13 features, 20,798 samples), \textbf{German Credit} (age/sex; 57 features, 1,000 samples), \textbf{Law School GPA} (race; binarized regression target at threshold 0.6; 3 features, 22,342 samples), and \textbf{MEPS 19} (race; 138 features, 15,830 samples). All datasets follow standard AIF360 preprocessing pipelines to ensure reproducibility.

\textbf{Synthetic Dataset}
To complement real-world datasets and enable controlled analysis of subgroup privacy leakage, we generate a synthetic binary classification dataset following the setup of Chang and Shokri~\cite{chang2021privacyrisksalgorithmicfairness}. The dataset contains $N=2{,}500$ samples with two continuous features, a binary sensitive attribute $G \in \{0,1\}$, and a binary label $Y \in \{0,1\}$. The sensitive attribute is intentionally imbalanced, with $\Pr(G{=}0)=0.2$ and $\Pr(G{=}1)=0.8$, reflecting real-world population disparities. Labels are generated conditional on $G$ such that the disadvantaged group exhibits higher base-rate skew: for $G=0$, $\Pr(Y{=}1)=0.9$, whereas for $G=1$, $\Pr(Y{=}1)=0.5$. Feature vectors are sampled from group–label–specific Gaussians, producing four subpopulations $(G,Y)$ with distinct means and covariance structures:
\[
X \sim \mathcal{N}(\mu_{g,y}, \Sigma_{g,y}),
\]
where the parameters are:
\begin{itemize}[noitemsep, leftmargin=*]
\item $(G{=}0, Y{=}0)$: $\mu=[0,-1]$, $\Sigma=\begin{bmatrix}7 & 1\\ 1 & 7\end{bmatrix}$
\item $(G{=}1, Y{=}0)$: $\mu=[-5, 0]$, $\Sigma=\begin{bmatrix}5 & 1\\ 1 & 5\end{bmatrix}$
\item $(G{=}0, Y{=}1)$: $\mu=[1,2]$, $\Sigma=\begin{bmatrix}5 & 2\\ 2 & 5\end{bmatrix}$
\item $(G{=}1, Y{=}1)$: $\mu=[2,3]$, $\Sigma=\begin{bmatrix}10 & 1\\ 1 & 4\end{bmatrix}$
\end{itemize}

This construction produces (i) unequal subgroup sizes, (ii) asymmetric class balance across protected groups, and (iii) distinct feature distributions - three core drivers of disparate vulnerability demonstrated in prior theory~\cite{KulynychYCVT22}. Because ground-truth generative structure is known, this dataset enables mechanistic examination of how fairness-enhancing algorithms and Differential Privacy alter subgroup-specific memorization and membership inference risk. All synthetic experiments use the same threat model, classifier families, and fairness interventions as the real datasets to ensure comparability.

\noindent\textbf{Models:}
Five distinct classifier architectures were employed to assess across different model families: 

\textbf{Decision Trees (DT)} were implemented using scikit-learn's default parameters. We set the maximum depth to 10, following prior work (e.g., Shokri et al.~\cite{chang2021privacyrisksalgorithmicfairness}), where this parameter is commonly used as a standard benchmark to balance between underfitting and overfitting. 
    
\textbf{Random Forest (RF)} classifiers were implemented using scikit-learn with constrained complexity (maximum depth set to $\lceil \frac{d}{2} \rceil$ where $d$ is the number of features), retaining default parameters for tree counts (100 estimators). This constrained complexity approach is supported by prior work such as~\cite{NADI2019112801} showing that shallower trees (with appropriate ensemble size) often suffice for strong performance without the risk of overly deep, overfit trees. Because the synthetic data task is 2-dimensional with smooth class boundaries, we regularize Random Forests by capping tree depth at 6 to prevent memorization.
    
\textbf{Neural Networks (NN)} were implemented using scikit-learn's MLPClassifier with ReLU activation functions and the Adam optimizer. The architecture was adapted to each dataset: a single hidden layer was used for moderate-dimensional datasets (Bank, COMPAS, Law School GPA); two hidden layers for the Law dataset; and three hidden layers for high-dimensional datasets such as German and MEPS19. Training employed an adaptive learning rate schedule and early stopping to reduce overfitting. The number of layers and overall setup were informed by existing literature to ensure alignment with expected test accuracies~\cite{bank-dataset-nn, pagliari2024comprehensivesustainableframeworkmachine, dressel2018accuracy, aithal2019credit, du2021fairness}. Our results closely replicate prior findings, with the exception of the Law School GPA dataset with classification setup, for which no clear benchmark values were available in the literature.
    
\textbf{Differentially Private Random Forest (DPRF)} classifiers were implemented using IBM’s Differential Privacy library. The model mirrors the configuration of scikit-learn’s RF, with the same maximum depth logic applied (i.e., $\lceil \frac{d}{2} \rceil$). In addition, it includes privacy-specific parameters. We tested the model under three different privacy levels, using $\epsilon$ values of 1, 10, and 50. As the literature commonly explores $\epsilon$ values in the range of 1 to 100, as noted in Holohan et al.~\cite{holohan2019diffprivlib}, our choices are consistent with these practices.
    
\textbf{Differentially Private Stochastic Gradient Descent Neural Network (DP-SGD NN)} was implemented in PyTorch as a feedforward neural network, with the number of hidden layers and units matched to the dataset-specific architectures described for the non-private Neural Networks (see \textbf{Neural Networks (NN)} above). Each hidden layer used ReLU activation, followed by a final linear output layer producing logits. Privacy was enforced using the Opacus \texttt{PrivacyEngine}, which applies per-sample gradient clipping and Gaussian noise addition during training to satisfy $(\epsilon, \delta)$-differential privacy. We set the target privacy budget to $\epsilon = 8.0$ and $\delta = 10^{-5}$ to balance privacy protection and model utility, following common practice in differentially private ML experiments. We used the Opacus default maximum gradient clipping norm of $1.0$, which bounds per-sample sensitivity while avoiding overly aggressive clipping. Models were trained with the Adam optimizer, a learning rate of $10^{-3}$, a batch size of 64, and for 20 epochs. All other hyperparameters and training settings were kept consistent with the non-private NN baselines, with the only difference being the use of DP-SGD to provide formal privacy guarantees. Our parameter settings follow the conventions established in the original DP-SGD paper~\cite{10.1145/2976749.2978318}. 

Because all six datasets in our study are tabular and of moderate dimensionality, we focus on models that are standard and widely deployed in tabular decision-making pipelines (tree-based ensembles and feedforward neural networks), rather than architectures tailored to images or long sequences such as convolutional networks or transformers. This ensures that our findings reflect realistic fairness-privacy behavior in the domains where these datasets are typically used.

These models provide a comparative analysis across simple non-parametric models, ensemble methods, and neural network architectures, while controlling for overfitting through architectural constraints and early stopping.


\textbf{Application of Differential Privacy}. We apply Differential Privacy (DP) using IBM’s Differential Privacy Library for Random Forests (configured to match \texttt{scikit-learn}) and Opacus for differentially private stochastic gradient descent (SGD) when training neural networks. Below, we describe how DP is integrated with fairness-enhancing techniques in our experiments:

\begin{enumerate}
    \item \textbf{Data Preparation:} We begin by uploading the dataset and splitting it into training and test sets.

    \item \textbf{Fairness Intervention (Pre-Processing):} If the fairness method is a pre-processing technique, it is applied 
    Pre-processing methods are applied to the training set only, except for Disparate Impact Remover (DIR), which operates on the full dataset. This ensures that fairness interventions do not interfere with the separation of training and test data.

    \item \textbf{Differentially Private Training:} We then train the model using its differentially private version (Random Forest or Neural Network), maintaining the same hyperparameters as in the non-private version to enable direct comparison. If the fairness method is \emph{in-processing} (e.g., Exponentiated Gradient Reduction), it is applied during this training phase. DP is applied solely during training, ensuring that all learned parameters and outputs satisfy the DP guarantee with respect to the training data.
    
    \item \textbf{Evaluation and Post-Processing:} After training, the model is evaluated on the test set. For subgroup-level analysis, the test set is partitioned into subpopulations, and metrics are reported per subgroup. If the fairness technique is \emph{post-processing} (e.g., Calibrated Equalized Odds), it is applied at this stage to model predictions-\emph{after} training, without access to the training data, thereby preserving DP guarantees.
\end{enumerate}

\section{Theoretical Characterization of Subpopulation Privacy Risk}
\label{sec:theory_synthetic}

\begin{figure*}[!ht]
    \captionsetup{font=footnotesize}
    \includegraphics[width=\textwidth]{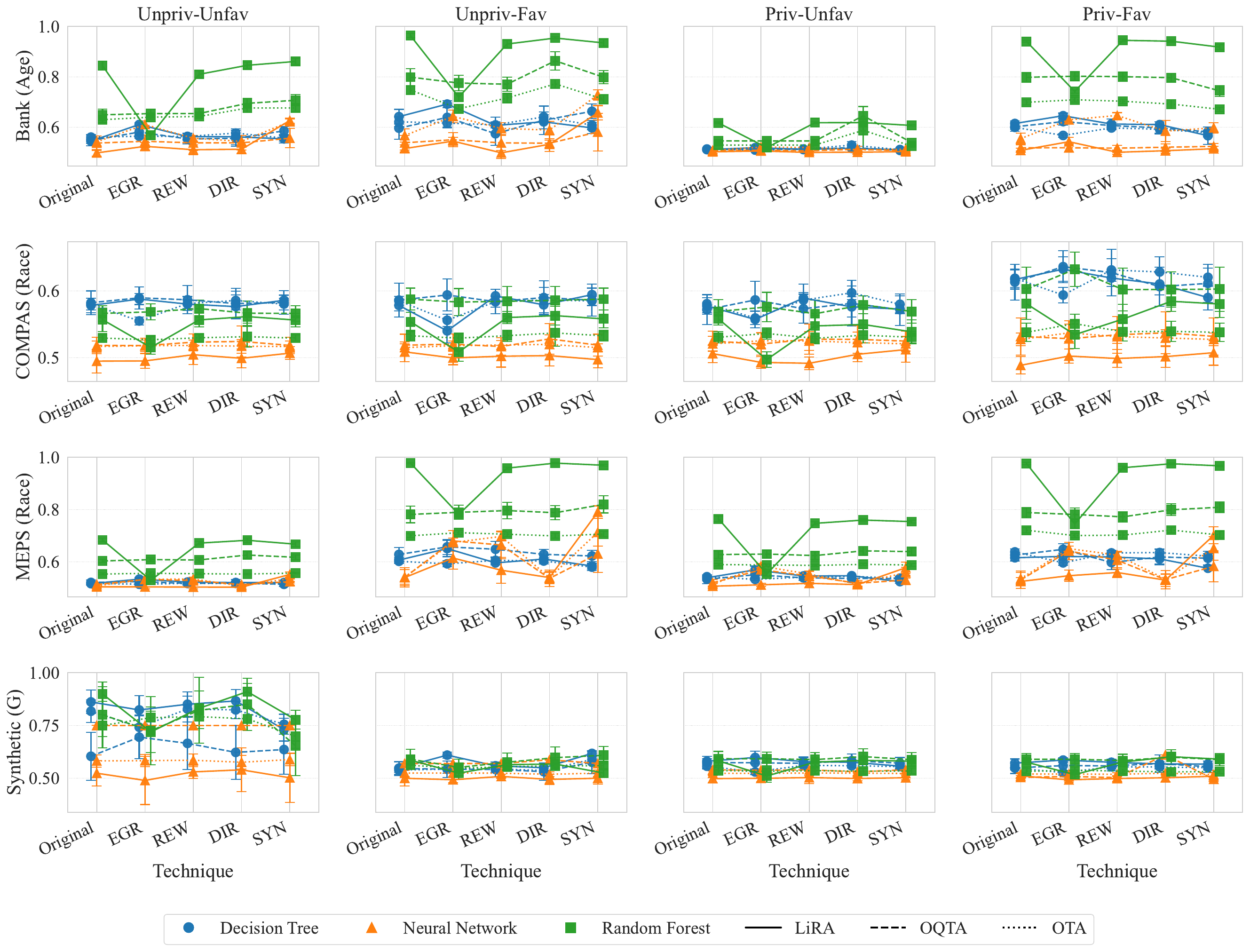} 
    \caption{Subpopulation-level membership inference vulnerability across datasets and fairness interventions. Plots show MIA success rates for every subgroup, dataset, and fairness mitigation strategy. Across all datasets minority subgroups exhibit consistently higher privacy leakage, while some fairness methods (e.g., egr) reduce subgroup-specific risk and others shift vulnerability unevenly.}
    \label{fig:all_models_attacks}
\end{figure*}

Before presenting our full empirical analysis, we provide a concise theoretical explanation for the main phenomena observed on the synthetic dataset and later confirmed in real-world experiments (see Figure~\ref{fig:all_models_attacks}): (i) smaller / rarer subpopulations are more vulnerable to membership inference attacks (MIAs); (ii) Differential Privacy (DP) reliably reduces membership-inference risk but can severely degrade subgroup utility, especially for underrepresented groups; and (iii) different fairness-enhancing algorithms (reweighting, pre-processing, in-processing, post-processing) modify subgroup risk via changes to effective sample size, class separability, and Fisher information. All derivations given below are carried out under simplifying assumptions (binary class label, Gaussian class-conditional features, and a smooth parametric classifier such as logistic regression). These assumptions match the synthetic data construction used in our experiments (see the synthetic-data specification in Section~\ref{sec:dataset_model} and results in Figure~\ref{fig:all_models_attacks}).

\textbf{Setup and notation}  
Let data points be \(z=(x,y,g)\) where \(x\in\mathbb{R}^d\) are features, \(y\in\{0,1\}\) the class label, and \(g\in\{0,1\}\) the sensitive attribute (group). The training set \(S\) contains \(n\) samples partitioned into \(n_g\) samples from subgroup \(g\). For clarity we write \(n_0\) and \(n_1\) for the two subgroup sizes. Assume a parametric classifier \(p_\theta(y\mid x)\) trained by empirical risk minimization with loss \(\ell(\theta;z)\) (e.g., negative log-likelihood for logistic regression). Denote by \(\hat\theta\) the estimator trained on the full dataset and by \(\hat\theta^{-i}\) the leave-one-out estimator excluding sample \(z_i\).

We study MIAs whose test statistic is (or is correlated with) the per-sample score difference between the model trained with and without that sample. Writing the per-sample logit \(s_\theta(x)\) (e.g., raw model output), a generic membership signal is:
\[
\Delta_i \triangleq s_{\hat\theta}(x_i) - s_{\hat\theta^{-i}}(x_i).
\]
Large (positive) \(\Delta_i\) indicates that including point \(i\) makes the model more confident on \(x_i\), which increases membership-inference success.


\textbf{Assumptions}: The following derivation relies on the ability to approximately isolate the Hessian contribution of each group. We emphasize that the conditions below are alternative sufficient conditions; the decomposition holds if any single one is satisfied:
\begin{enumerate}[label=(\roman*), noitemsep, leftmargin=*]
    \item \textbf{Distinct feature distributions}: Groups occupy different regions of the feature space. This is a standard prerequisite in fairness settings; if protected groups had identical distributions, disparate impact would generally not occur.
    \item \textbf{Group-specific parameters}: In overparameterized models, different subsets of parameters (e.g., specific neurons or ensemble branches) effectively specialize for different subgroups.
    \item \textbf{Local decision boundaries}: The boundary near group $g$ is shaped primarily by within-group samples, which often occurs as a natural consequence of condition (i). In our study of tabular fairness datasets, sensitive attributes (e.g., race, gender) often correlate with distinct feature distributions, making these conditions realistic for providing explanatory intuition for our empirical findings.
\end{enumerate}

We now present the theoretical analysis that may explain the observed effects.

\subsection{Influence function framework}

Under standard regularity conditions and for smooth loss \(\ell\), the leave-one-out parameter difference admits the first-order influence-function approximation~\cite{cook1982residuals, hampel1986robust, koh2017understanding}:
\[
\hat\theta - \hat\theta^{-i} \approx \frac{1}{n} H^{-1}\nabla_\theta \ell(\hat\theta; z_i),
\]
where \(H = \frac{1}{n}\sum_{j=1}^n \nabla^2_\theta \ell(\hat\theta; z_j)\) is the empirical Hessian at \(\hat\theta\). This is the classical influence function result from M-estimation theory.

Applying a first-order Taylor expansion to the score difference:
\[
\Delta_i \approx \nabla_\theta s_{\hat\theta}(x_i)^\top (\hat\theta - \hat\theta^{-i}) \approx \frac{1}{n} \nabla_\theta s_{\hat\theta}(x_i)^\top H^{-1}\nabla_\theta \ell(\hat\theta; z_i).
\]

\subsection{Why small groups are more vulnerable}

Prior empirical work has documented that minority subgroups exhibit higher vulnerability to membership inference attacks~\cite{yaghini2019disparate, NEURIPS2019_fc0de4e0, 9014384, zhong2022understanding}. Here we provide a theoretical explanation for this phenomenon through influence function analysis.

To understand the subgroup-size dependence, we decompose the empirical Hessian by group:
\[
H = \frac{1}{n}\sum_{j=1}^n \nabla^2_\theta \ell(\hat\theta; z_j) = \frac{n_0}{n}H_0 + \frac{n_1}{n}H_1,
\]
where \(H_g = \frac{1}{n_g}\sum_{j:g_j=g} \nabla^2_\theta \ell(\hat\theta; z_j)\) is the average Hessian within group \(g\).

For many learning problems, particularly when groups have distinct distributions or when the model has group-specific parameters, the curvature information about group \(g\) is primarily captured by \(H_g\) and contributes to \(H\) with weight \(n_g/n\). 

Consider the simplified case where the loss decomposes across groups (or parameters affecting each group are approximately independent). For a sample \(z_i\) from group \(g\), the relevant second-order information is dominated by the group-specific component. In this regime, \(H^{-1} \approx (n/n_g)H_g^{-1}\), giving:
\[
H^{-1}\nabla_\theta \ell(\hat\theta; z_i) \approx \frac{n}{n_g}H_g^{-1}\nabla_\theta \ell(\hat\theta; z_i).
\]

Substituting back:
\begin{align}
\Delta_i &\approx \frac{1}{n} \nabla_\theta s_{\hat\theta}(x_i)^\top \cdot \frac{n}{n_g}H_g^{-1}\nabla_\theta \ell(\hat\theta; z_i) \nonumber\\
&= \frac{1}{n_g} \nabla_\theta s_{\hat\theta}(x_i)^\top H_g^{-1}\nabla_\theta \ell(\hat\theta; z_i).
\end{align}

\textbf{Result and Interpretation}: The membership signal \(\Delta_i\) scales as \(O(1/n_g)\). Each sample from group \(g\) represents fraction \(1/n_g\) of the group's information, so removing it eliminates a larger share of group-specific knowledge when \(n_g\) is small. This amplifies the observable change in model predictions, linking smaller group size to larger \(|\Delta_i|\) and higher MIA vulnerability. This aligns with theoretical work on deep learning memorization, which establishes that models must effectively "memorize" examples from rare subpopulations (where $n_g$ is small) to achieve low training error, resulting in high influence scores and privacy vulnerability~\cite{NEURIPS2020_1e14bfe2, feldman2021doeslearningrequirememorization}.

\subsection{Effect of Differential Privacy: noise reduces signal-to-noise ratio}

\textbf{DP lowers membership signal but may disproportionately harm small groups.}
DP-SGD adds calibrated noise to gradients during training, utilizing mechanisms such as those introduced by Abadi et al.~\cite{10.1145/2976749.2978318}. This has the cumulative effect of perturbing the learned parameters. As a first-order approximation, we can model this as additive Gaussian noise on the final parameters: \(\hat\theta_{\text{DP}} \approx \hat\theta + \eta\) where \(\eta \sim \mathcal{N}(0, \sigma^2 I)\) and \(\sigma\) depends on the privacy budget.

Critically, the DP-trained model \(\hat\theta_{\text{DP}}\) and its leave-one-out counterpart \(\hat\theta_{\text{DP}}^{-i}\) receive \emph{independent} noise realizations (since they are trained on different datasets with independent randomness). For a linear score function \(s_\theta(x) = \theta^\top x\), the noisy membership signal becomes:
\[
\widetilde{\Delta}_i = s_{\hat\theta_{\text{DP}}}(x_i) - s_{\hat\theta_{\text{DP}}^{-i}}(x_i) \approx \Delta_i + (\eta - \eta^{-i})^\top x_i,
\]
where \(\Delta_i = s_{\hat\theta}(x_i) - s_{\hat\theta^{-i}}(x_i)\) is the non-private signal and \((\eta - \eta^{-i})^\top x_i\) is the additive noise term with variance \(2\sigma^2\|x_i\|^2\).

The signal-to-noise ratio for distinguishing members from non-members degrades as:
\[
\text{SNR} \propto \frac{(\mathbb{E}[\Delta_i | \text{member}] - \mathbb{E}[\Delta_i | \text{non-member}])^2}{\operatorname{Var}(\Delta_i) + 2\sigma^2\|x_i\|^2}.
\]
When \(2\sigma^2\|x_i\|^2 \gg \operatorname{Var}(\Delta_i)\), the DP noise dominates and the membership signal becomes undetectable, reducing MIA success.

\textbf{Disproportionate impact on small groups.}
While DP noise uniformly affects all samples, its impact on model utility is heterogeneous across subgroups. Recall that for small subgroups with size \(n_g\), the model has less statistical information to learn group-specific patterns. If we view the learning problem for group \(g\) as distinct, the effective signal strength for 
classification in group \(g\) scales roughly with \(\sqrt{n_g}\) (from standard learning theory~\cite{10.5555/2621980}), while DP noise remains constant at level \(\sigma\).

For minority groups where $n_g$ is small, the learned decision boundary exhibits higher variance and is less reliable, while the effective margin between classes is smaller. Consequently, DP noise of magnitude $\sigma$ can overwhelm this weak group-specific signal, manifesting as catastrophic utility degradation (near-zero accuracy).

In contrast, majority groups with large \(n_g\) have stronger learned signals that remain above the noise floor even with DP. This explains the empirical observations in Tables~\ref{tab:utility_privacy_core_Lira}--\ref{tab:utility_privacy_core_oqta} and Figure~\ref{fig:dprf_lira_test_accuracies}, where DP training preserves utility for majority groups while collapsing accuracy for underrepresented groups.

This creates a tension: DP successfully reduces MIA vulnerability (good for privacy), but the uniform noise injection disproportionately harms groups that have weaker representation in the training data (bad for fairness).

\subsection{How fairness interventions change subgroup MIA risk}

Below we analyze how common fairness strategies modify the learned model and consequently affect MIA vulnerability across subgroups. The key insight is that fairness interventions change the decision boundaries, margins, and loss landscape in group-specific ways.

\textbf{Reweighting (REW)} multiplies each sample's loss by a weight \(w_{g,y}\) that upweights underrepresented subgroup-label combinations~\cite{kamiran2012data}. The weighted objective becomes:
\[
\hat\theta_{\text{REW}} = \arg\min_\theta \frac{1}{n}\sum_{i=1}^n w_{g_i,y_i}\ell(\theta;z_i).
\]

This changes MIA vulnerability through two mechanisms. First, the learned \(\hat\theta_{\text{REW}}\) differs from the unweighted \(\hat\theta\), typically improving classification margins for upweighted groups. Better margins mean smaller per-sample gradients \(\|\nabla_\theta \ell(\hat\theta_{\text{REW}}; z_i)\|\) for correctly classified samples, which reduces the membership signal \(\Delta_i\). Second, the influence function for weighted ERM becomes:
\[
\hat\theta_{\text{REW}} - \hat\theta_{\text{REW}}^{-i} \approx \frac{w_{g_i,y_i}}{n} H_w^{-1}\nabla_\theta \ell(\hat\theta_{\text{REW}}; z_i),
\]
where \(H_w = \frac{1}{n}\sum_j w_{g_j,y_j}\nabla^2_\theta \ell(\hat\theta_{\text{REW}}; z_j)\). While individual sample influence is scaled by \(w_{g_i,y_i}\), the Hessian \(H_w\) also changes, making the net effect depend on the balance between upweighting in numerator vs. denominator.

Empirically, REW often reduces MIA vulnerability for minority groups, primarily because the improved model quality (better margins, lower loss) reduces the distinguishability between member and non-member samples.

\textbf{Disparate Impact Remover (DIR)} reduces correlation between features and sensitive attributes by projecting data to remove disparate impact~\cite{feldman2015certifying}. This changes the class-conditional distributions $P(x|y,g)$, potentially reducing distributional differences between groups, and changing the feature space geometry (affecting $\|x_i\|$ and thus noise terms). The impact on MIA depends on whether DIR makes the learning problem easier (better generalization $\to$ lower vulnerability) or introduces artifacts (potential for different types of memorization).

\textbf{Synthetic data augmentation (SYN)} increases training set size by adding synthetic samples for minority groups~\cite{doi:10.1137/1.9781611977653.ch98, 10.5555/1622407.1622416}. This effectively increases \(n_g\), which should reduce per-sample influence (\(\Delta_i \propto 1/n_g\)). However, if synthetic samples are low-quality or highly correlated with real samples, they may not provide independent information, limiting the privacy benefit.

In summary, fairness interventions modify MIA risk primarily by changing the learned model \(\hat\theta\) and the resulting classification margins, loss values, and gradient magnitudes for each subgroup. The direction and magnitude of privacy impact depend on the specific intervention and how it reshapes the loss landscape.

We would like to stress that the above derivations use linearized / first-order approximations (influence functions, additive DP noise). Nonlinear model behavior (deep nets), adaptive training, and more complex fairness algorithms can produce richer dynamics; nevertheless the arguments identify the dominant mechanisms that explain the empirical patterns reported in Section~\ref{sec:Empirical} and illustrated in Figure~\ref{fig:all_models_attacks}

\section{Empirical Results}
\label{sec:Empirical}

\subsection{Impact of Fairness-Enhancing Algorithms}
\label{sec:fairness_impact}
\begin{figure}[!t]
    \captionsetup{font=footnotesize}
    \centering
    \includegraphics[width=\columnwidth]{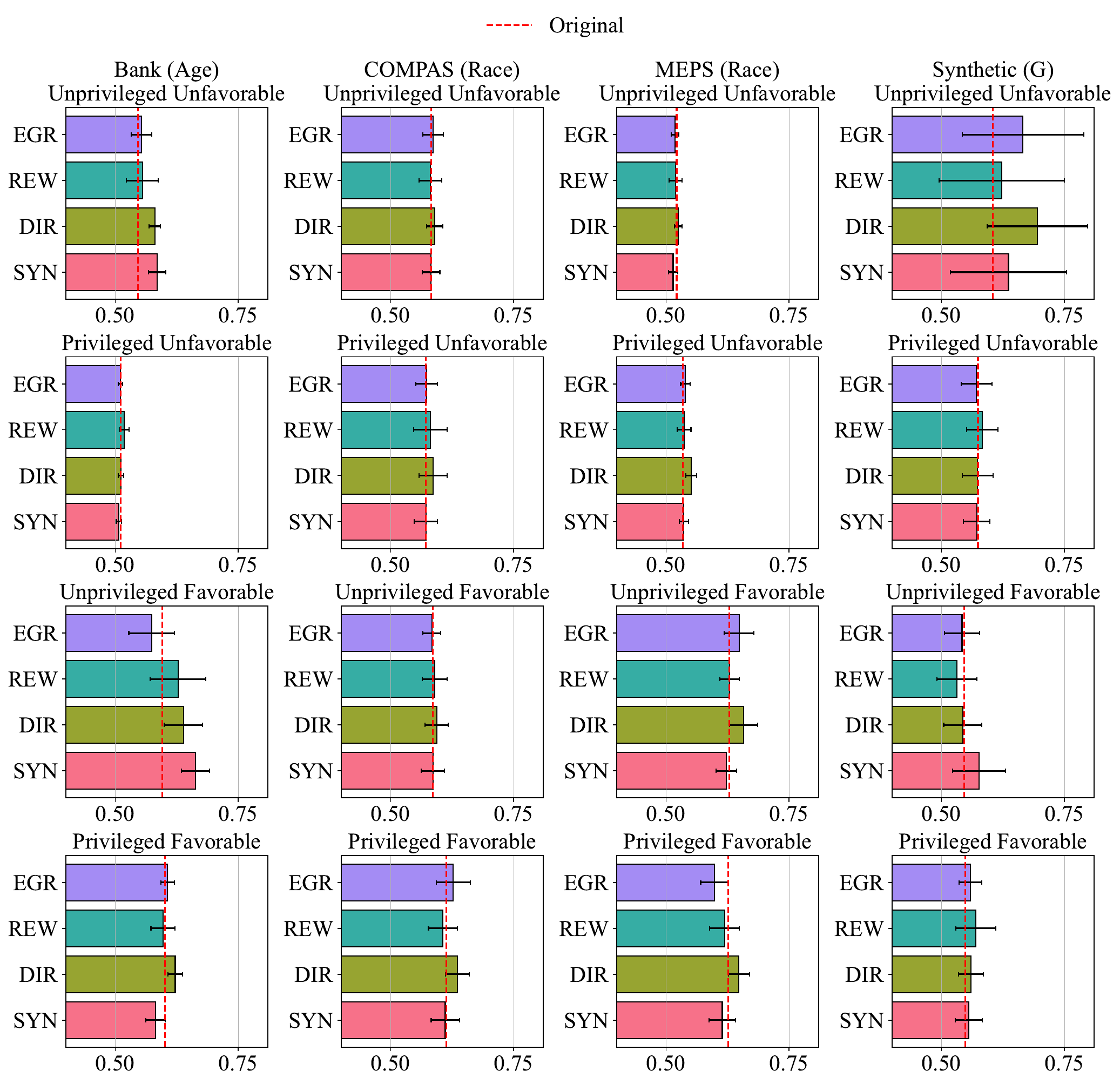} 
    
    \caption{Decision Tree privacy risk results under OQTA for: Bank (Age), COMPAS (Race), MEPS (Race), and Synthetic (G) datasets. 
    It visualizes subpopulation privacy risks across different fairness mitigation techniques. The red dashed line indicates baseline risk from the original unmitigated model. Values are averaged over 20 runs, while the standard deviation is shown with error bars. The horizontal range is dynamic for better visibility.}
    \label{fig:dt_oqta_mia_1}
\end{figure}

Mitigation strategies reveal several notable patterns. As shown in Figure~\ref{fig:all_models_attacks}, \textbf{EGR} generally achieves the strongest reductions in privacy risk for the more complex RF model. In contrast, \textbf{REW} and \textbf{SYN} yield only modest privacy-risk reductions for simpler models such as DT and NN, and these differences are typically small (Figures~\ref{fig:all_models_attacks},~\ref{fig:dt_oqta_mia_1} and Appendix~\ref{appendix:pr_graphs}). \textbf{DIR} exhibits inconsistent behavior, sometimes increasing and other times decreasing privacy risk, while \textbf{EGR} itself performs poorly on DT and NN despite its strong performance on RF. While \textbf{SYN} performs well on Bank and COMPAS, it does not exhibit consistent advantages across other datasets, indicating that its effectiveness is dataset-dependent. Notably, these fairness algorithms preserve overall accuracy within $\pm$1--3\% across all datasets, but subpopulation-level utility shifts can be dramatic, reaching up to $\pm$33\% for individual subgroups (details in Appendix \ref{app:subpop-utility}). No single method consistently improves utility for underrepresented groups across all datasets, further confirming that fairness--privacy--utility trade-offs are dataset-dependent.

\begin{figure}[!htbp]
    \captionsetup{font=footnotesize}
    \centering
    \includegraphics[width=\columnwidth]{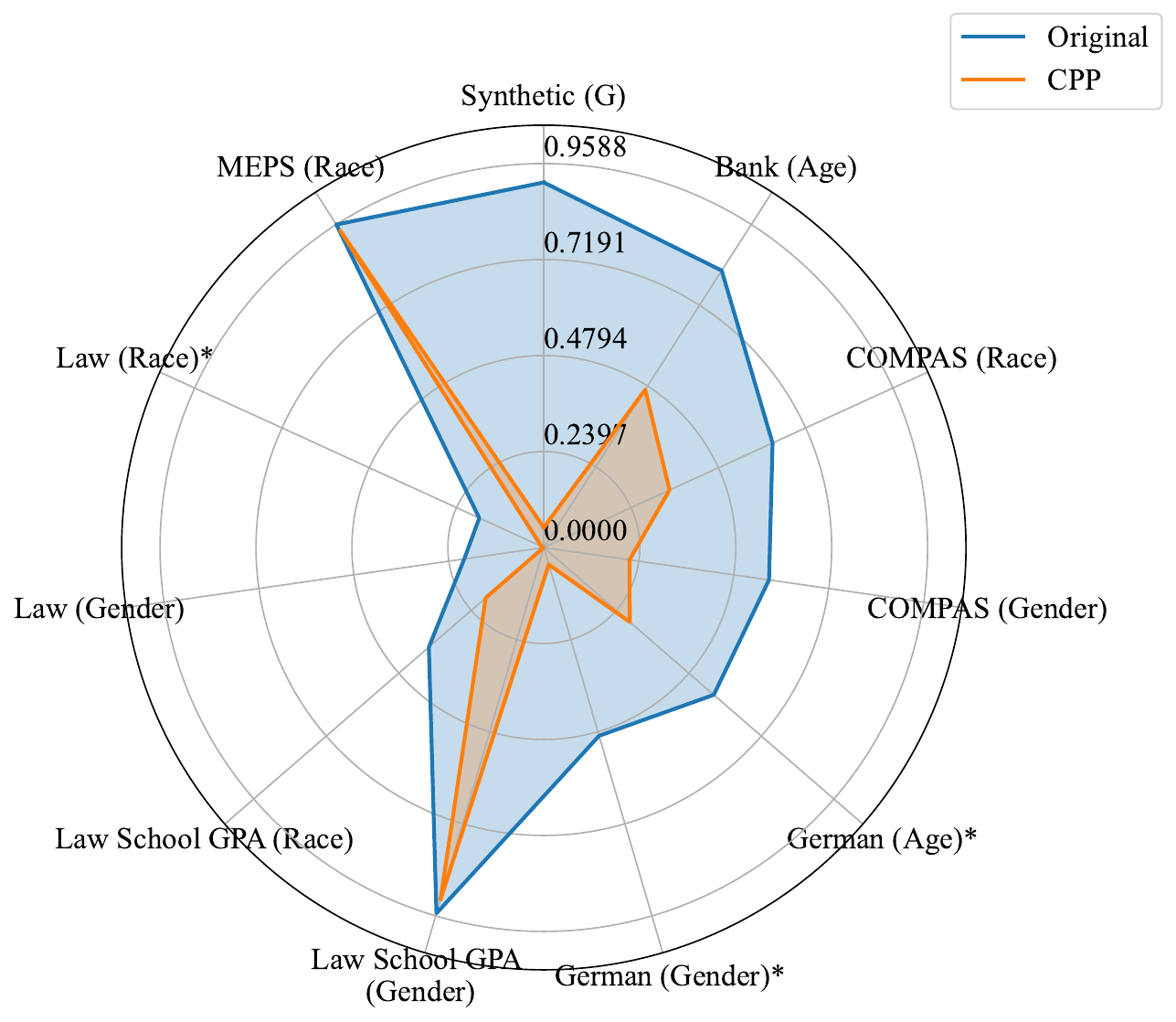} 
    \caption{Radial plot comparing the subpopulation most affected by the CPP mitigator in each dataset for the Decision Tree (DT) model under the OTA attack. For every dataset, the plotted value corresponds to the subpopulation with the largest positive gap between the original and CPP test accuracies, indicating where CPP reduces accuracy the most. Dataset labels marked with an asterisk (*) denote cases where this most-affected subpopulation is the underrepresented group. The outer polygon shows the original model’s accuracy, while the inner polygon shows the CPP-mitigated accuracy, highlighting the extent of subgroup-level utility degradation introduced by the mitigator.}
    \label{fig:cpp_test_accuracies}
\end{figure}

\label{sec:dataset_model}
\begin{figure*}[ht]
    \captionsetup{font=footnotesize}
    \centering
    \includegraphics[width=0.9\textwidth]{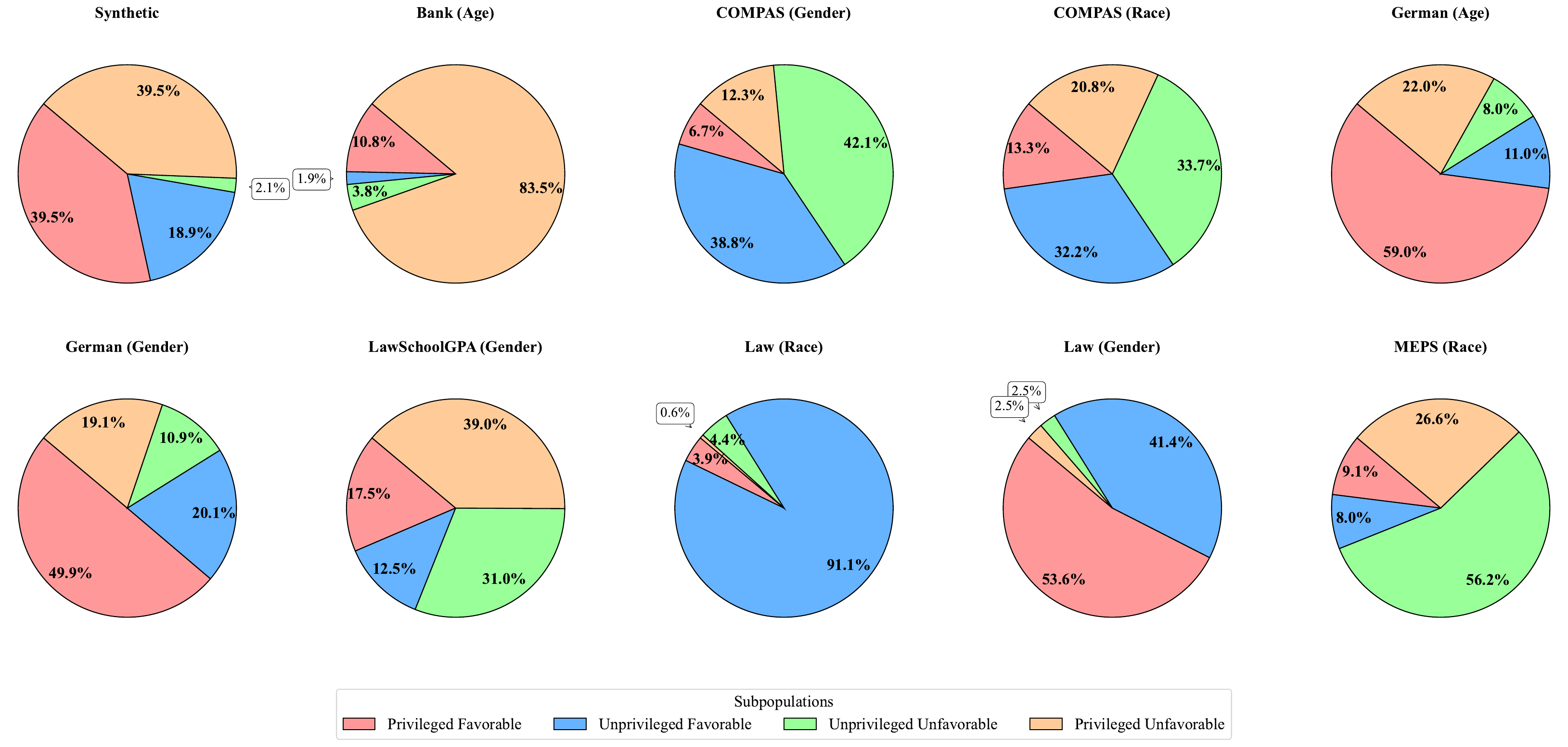}
    \caption{Distribution of data samples across all subpopulations for each dataset.}
    \label{fig:subpopulation_sizes}
\end{figure*}

\begin{figure}[!t]
    \captionsetup{font=footnotesize}
    \centering
    \includegraphics[width=\columnwidth]{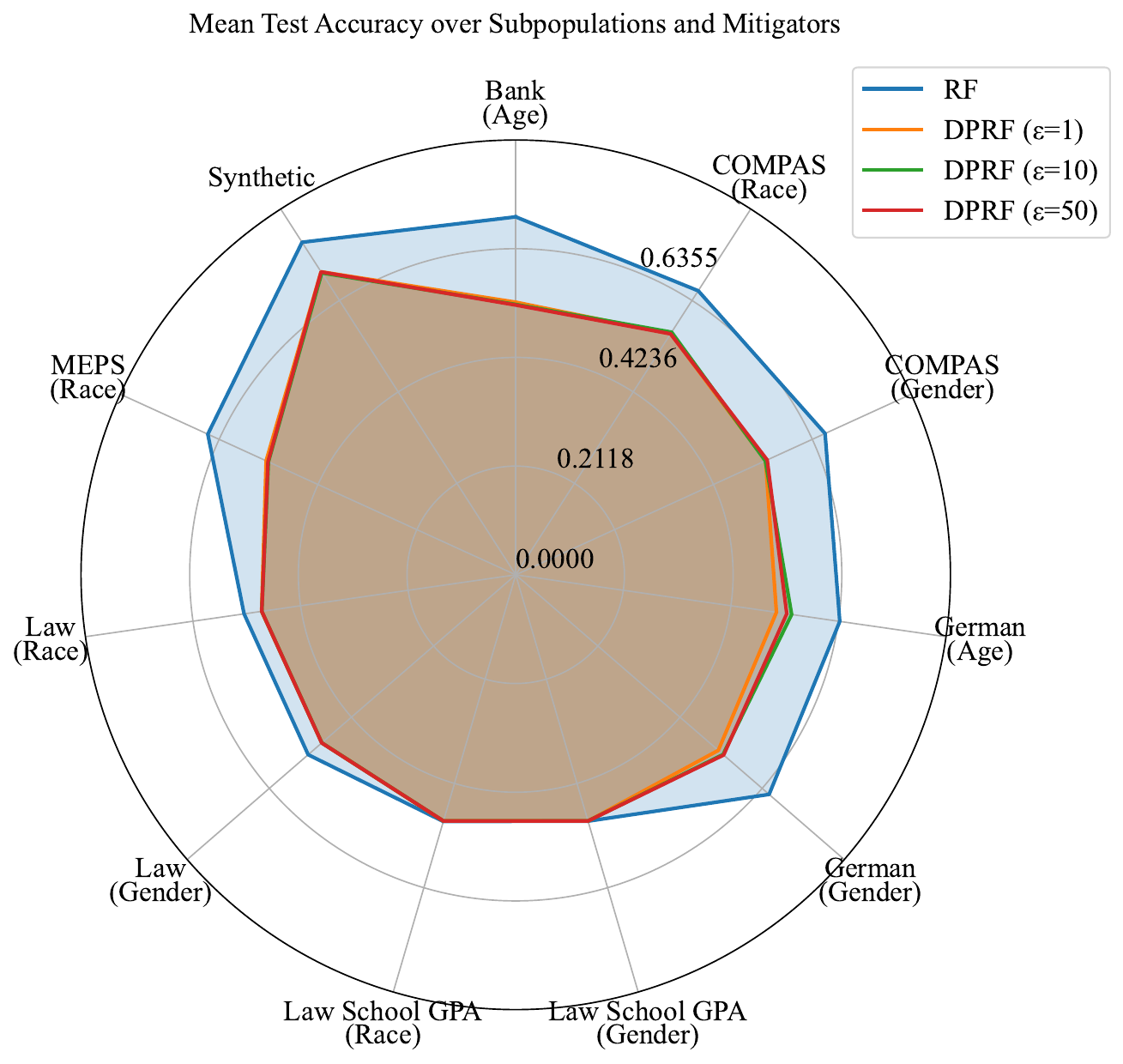} 
    \caption{Comparison of mean test accuracies across subpopulations and fairness mitigation methods for different models under LiRA setup (RF and DPRF with varying privacy epsilons). Each point on the radar represents the average accuracy across subpopulations and methods for a dataset.}

    \label{fig:dprf_lira_test_accuracies}
\end{figure}

\begin{figure*}[t]
    \centering

    \subfloat[{\fontfamily{ptm}\selectfont\footnotesize Decision Tree}]{
        \includegraphics[width=0.45\textwidth]{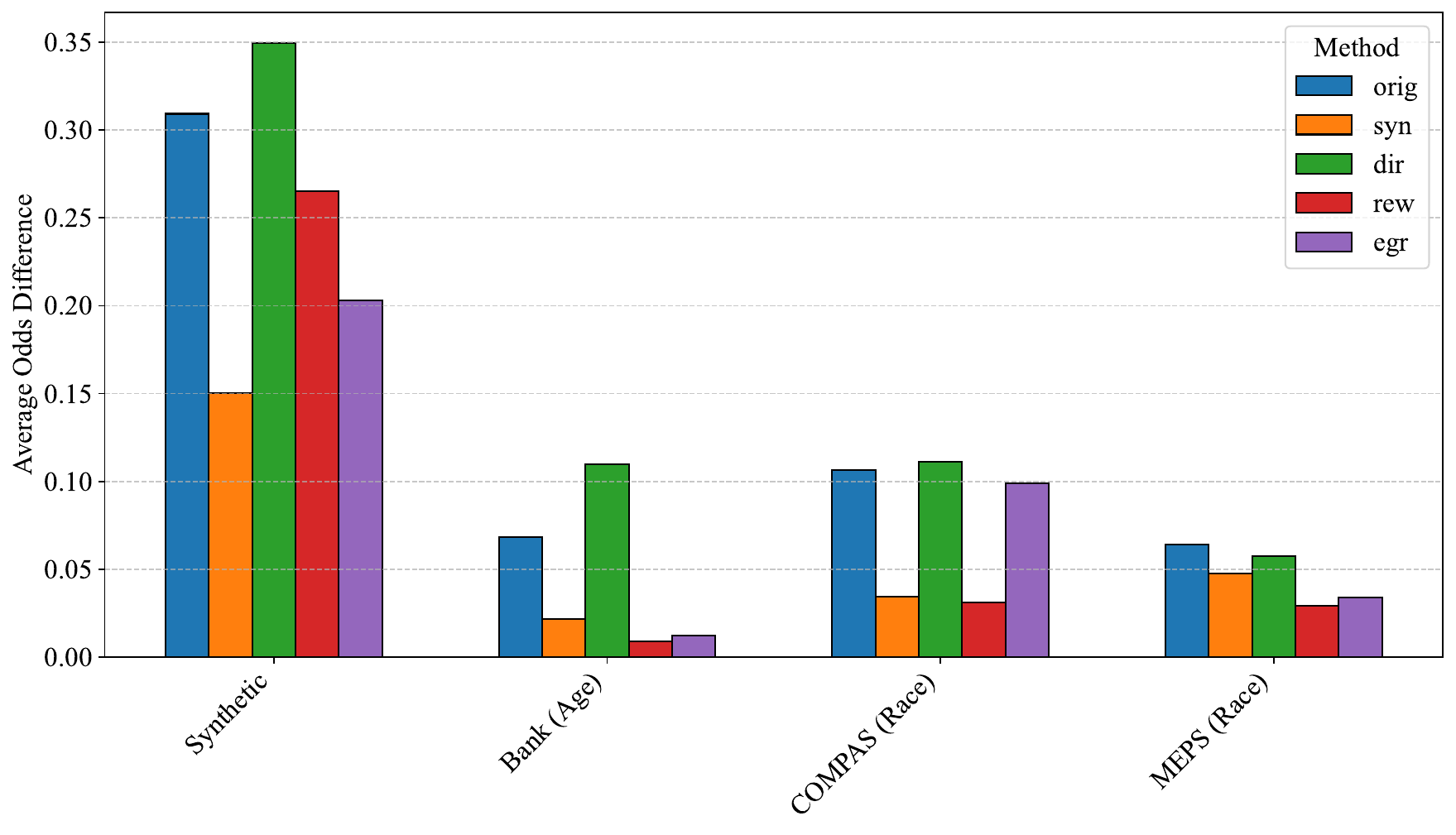}
        \label{fig:lira_fairness_dt}
    }
    \hfill
    \subfloat[{\fontfamily{ptm}\selectfont\footnotesize Random Forest}]{
        \includegraphics[width=0.45\textwidth]{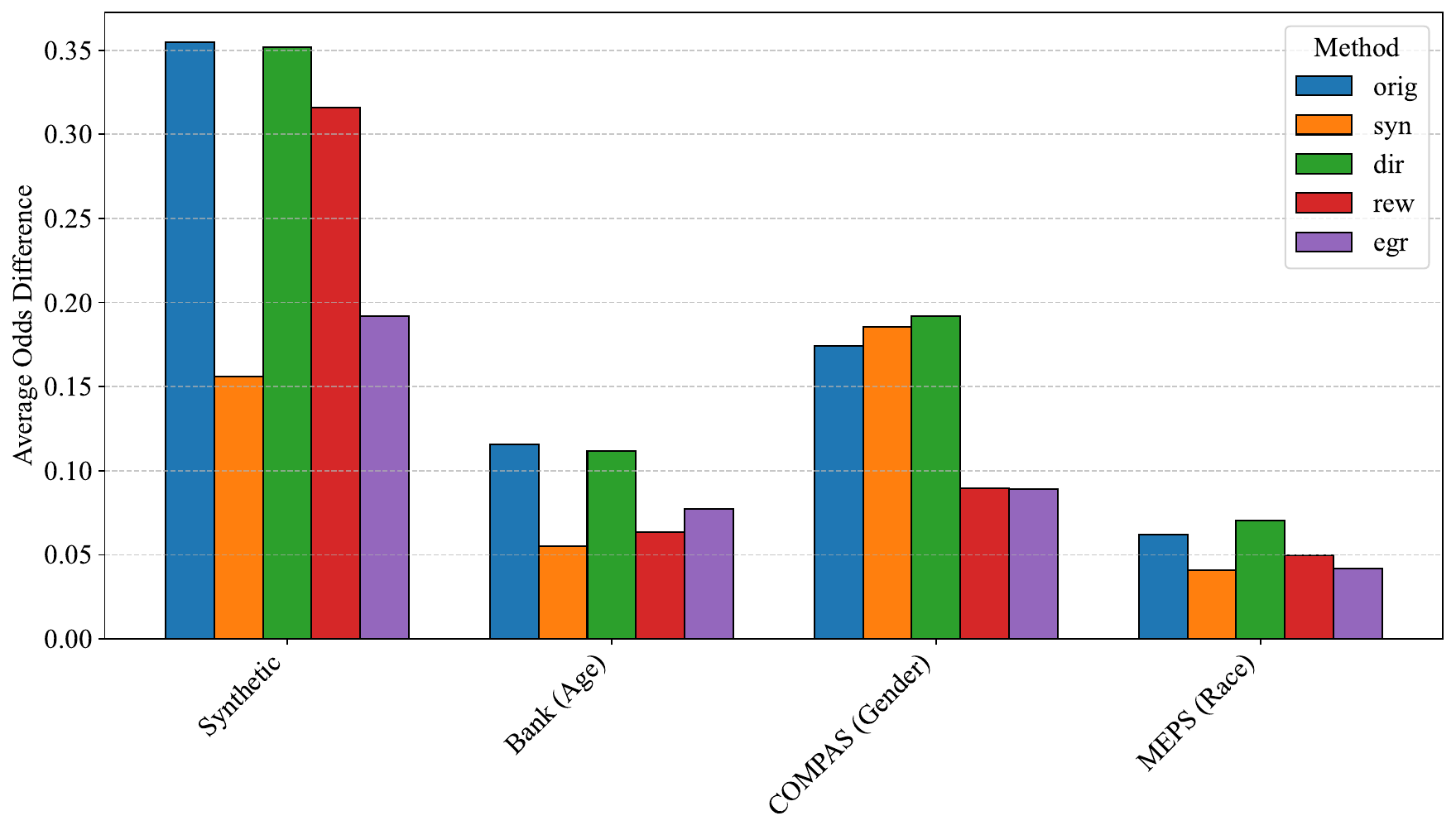}
        \label{fig:lira_fairness_rf}
    }


    \subfloat[{\fontfamily{ptm}\selectfont\footnotesize Neural Network}]{
        \includegraphics[width=0.45\textwidth]{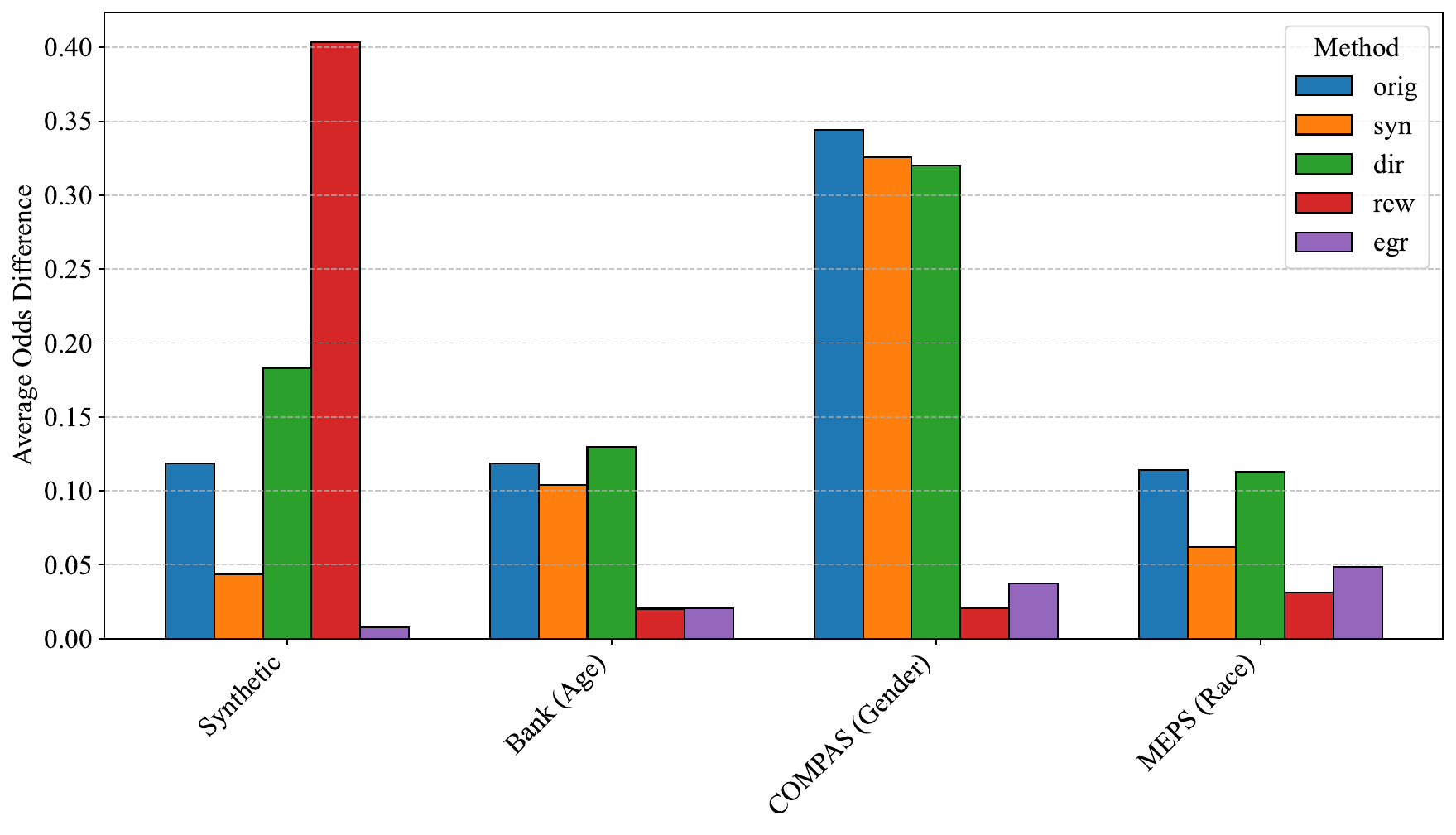}
        \label{fig:lira_fairness_nn}
    }
    \hfill
    \subfloat[{\fontfamily{ptm}\selectfont\footnotesize DP Random Forest ($\varepsilon{=}1$)}]{
        \includegraphics[width=0.45\textwidth]{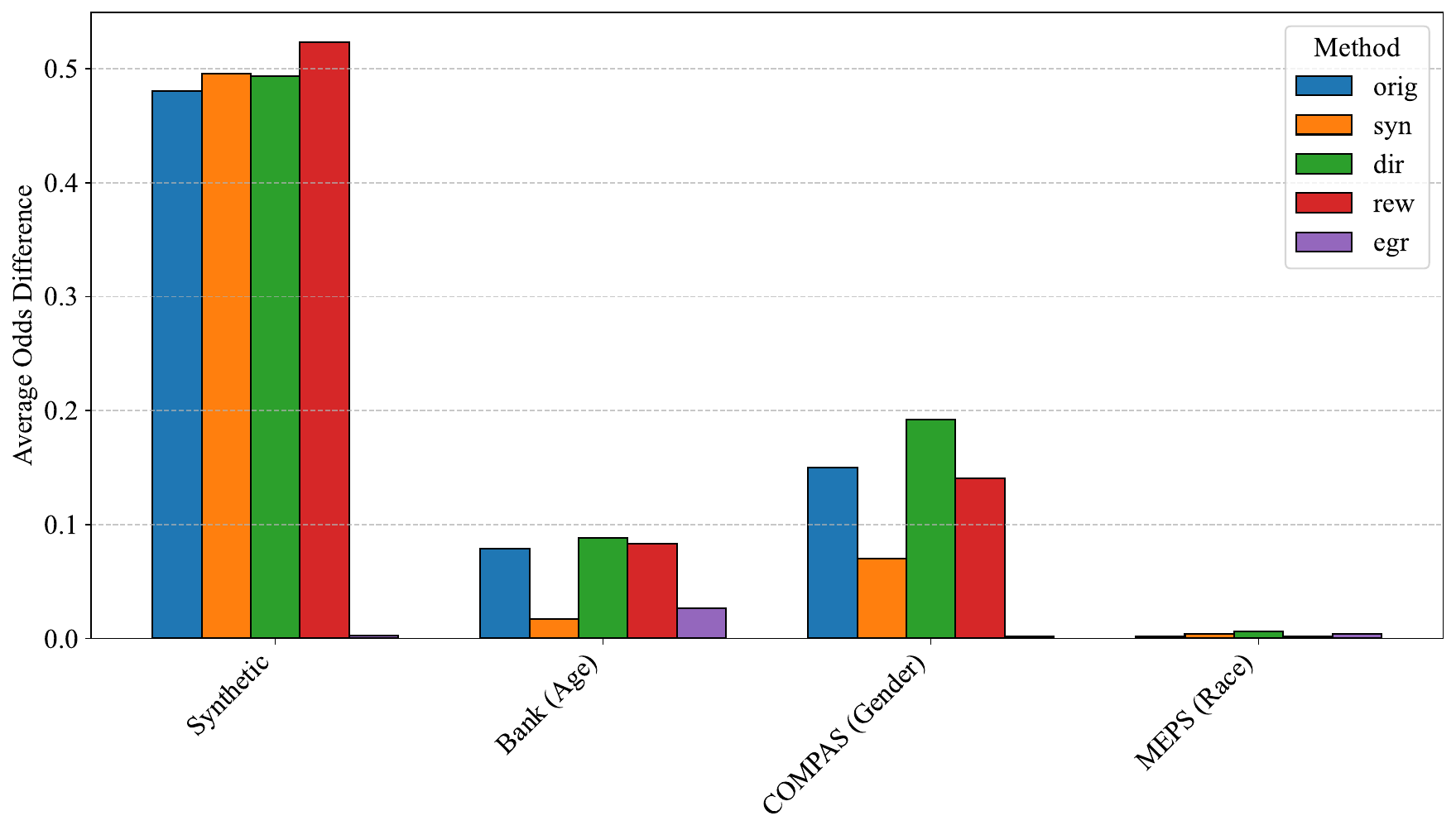}
        \label{fig:lira_fairness_dprf}
    }

    \caption{Fairness metric (average odds difference) results under LiRA attack. Lower values indicate better fairness.}
    \label{fig:lira_fairness_combined}
\end{figure*}

\begin{table*}[t]
\centering
\caption{
Subpopulation utility difference (difference between the test accuracy for the subpopulation before and after application of DP) and privacy risks for subpopulation after the application of DP (DPRF vs. RF) across fairness mitigators for LiRA setup. Rows show datasets with fairness mitigators. Asterisks (*) mark low-utility subgroups (< 5\%) whereas tick (\checkmark) means that the privacy risk is minimized for the given subpopulation (privacy risk goes down to 50\(\pm\)2\%).
}
\label{tab:utility_privacy_core_Lira}
\resizebox{\textwidth}{!}{%
\begin{tabular}{llrrrrrrrr}
\toprule
 & & \multicolumn{2}{c}{\(G_0^{-}\)} & \multicolumn{2}{c}{\(G_1^{-}\)} & \multicolumn{2}{c}{\(G_0^{+}\)} & \multicolumn{2}{c}{\(G_1^{+}\)} \\
\cmidrule(lr){3-4} \cmidrule(lr){5-6} \cmidrule(lr){7-8} \cmidrule(lr){9-10}
Dataset & Mitigator & Utility Diff. & Privacy & Utility Diff. & Privacy & Utility Diff. & Privacy & Utility Diff. & Privacy \\
\midrule
\multirow{5}{*}{Synthetic (G)} 
 & orig & -25.49\% &  & -0.75\% & \checkmark & 2.02\% & \checkmark & 1.60\% & \checkmark \\
\cline{2-10}
 & syn & -46.06\% &  & -1.50\% & \checkmark & 20.37\% & \checkmark & -2.97\% & \checkmark \\
\cline{2-10}
 & rew & -30.33\% & \checkmark & 1.67\% & \checkmark & 4.38\% &  & -5.06\% & \checkmark \\
\cline{2-10}
 & dir & -27.64\% &  & -0.53\% & \checkmark & 0.99\% & \checkmark & -0.19\% & \checkmark \\
\cline{2-10}
 & eg & -30.32\% &  & -69.24\% & \checkmark & -7.67\% & \checkmark & -7.46\% &  \\
\midrule
\multirow{5}{*}{MEPS (race)} & orig & 2.33\% & \checkmark & 5.16\% & \checkmark & -25.42\% & \checkmark & -33.14\% & \checkmark \\
\cline{2-10}
 & syn & 2.01\% & \checkmark & 4.44\% & \checkmark & -25.82\% & \checkmark & -31.82\%* & \checkmark \\
\cline{2-10}
 & rew & 2.25\% & \checkmark & 4.93\% & \checkmark & -24.28\% & \checkmark & -32.47\% & \checkmark \\
\cline{2-10}
 & dir & 2.09\% & \checkmark & 4.96\% & \checkmark & -23.82\% & \checkmark & -33.34\% & \checkmark \\
\cline{2-10}
 & eg & -2.67\% & \checkmark & -0.51\% & \checkmark & -25.67\% & \checkmark & -29.85\% & \checkmark \\
\midrule
\multirow{5}{*}{Bank (age)} & orig & 4.08\% & \checkmark & 1.55\% & \checkmark & -39.70\% & \checkmark & -35.04\% & \checkmark \\
\cline{2-10}
 & syn & 4.38\% &  & 1.45\% & \checkmark & -37.33\% & \checkmark & -34.35\%* & \checkmark \\
\cline{2-10}
 & rew & 2.92\% & \checkmark & 1.80\% & \checkmark & -33.71\% & \checkmark & -35.28\% & \checkmark \\
\cline{2-10}
 & dir & 4.62\% & \checkmark & 2.00\% & \checkmark & -40.86\% & \checkmark & -36.05\% & \checkmark \\
\cline{2-10}
 & eg & -9.42\% & \checkmark & -14.30\% & \checkmark & -26.90\% & \checkmark & -23.37\% & \checkmark \\
\midrule
\multirow{5}{*}{COMPAS (race)} & orig & 12.77\% & \checkmark & 13.93\% & \checkmark & -29.93\% & \checkmark & -36.70\% & \checkmark \\
\cline{2-10}
 & syn & 7.88\% & \checkmark & 14.82\% & \checkmark & -22.47\% & \checkmark & -40.39\%* & \checkmark \\
\cline{2-10}
 & rew & 14.72\% & \checkmark & 17.06\% & \checkmark & -31.40\% & \checkmark & -43.02\%* & \checkmark \\
\cline{2-10}
 & dir & 1.82\% & \checkmark & 13.84\% & \checkmark & -15.64\% & \checkmark & -37.84\%* & \checkmark \\
\cline{2-10}
 & eg & 24.71\% & \checkmark & 20.82\% & \checkmark & -54.44\%* & \checkmark & -43.40\%* & \checkmark \\
\bottomrule
\end{tabular}
}
\end{table*}

\begin{table*}[t]
\centering
\caption{
Subpopulation utility difference (difference between the test accuracy for the subpopulation before and after application of DP) and privacy risks for subpopulation after the application of DP (DPRF vs. RF) across fairness mitigators for OQTA setup. Rows show datasets with fairness mitigators. Asterisks (*) mark low-utility subgroups (< 5\%) whereas tick (\checkmark) means that the privacy risk is minimized for the given subpopulation (privacy risk goes down to 50\(\pm\)2\%). Values are averaged over 20 runs, while the standard deviation is in the interval [0;0.06].
}
\label{tab:utility_privacy_core_oqta}
\resizebox{\textwidth}{!}{
\begin{tabular}{llrrrrrrrr}
\toprule
 & & \multicolumn{2}{c}{\(G_0^{-}\)} & \multicolumn{2}{c}{\(G_1^{-}\)} & \multicolumn{2}{c}{\(G_0^{+}\)} & \multicolumn{2}{c}{\(G_1^{+}\)} \\
\cmidrule(lr){3-4} \cmidrule(lr){5-6} \cmidrule(lr){7-8} \cmidrule(lr){9-10}
Dataset & Mitigator & Utility Diff. & Privacy & Utility Diff. & Privacy & Utility Diff. & Privacy & Utility Diff. & Privacy \\
\midrule
\multirow{5}{*}{Synthetic (G)} 
 & orig & -23.76\% &  & 6.10\% &  & 3.86\% &  & -4.01\% & \checkmark \\
\cline{2-10}
 & syn & -44.29\% &  & 0.32\% & \checkmark & 28.34\% & \checkmark & 3.89\% & \checkmark \\
\cline{2-10}
 & rew & -27.08\% &  & 12.06\% &  & 5.13\% &  & -8.31\% & \checkmark \\
\cline{2-10}
 & dir & -14.13\% &  & 6.46\% &  & 1.29\% & \checkmark & -10.77\% &  \\
\cline{2-10}
 & eg & -30.06\% &  & -55.25\% & \checkmark & 12.70\% &  & 22.10\% &  \\
\midrule
\multirow{5}{*}{Bank (age)} & orig & 14.00\% & \checkmark & 3.39\% & \checkmark & -55.47\%* & \checkmark & -42.63\%* & \checkmark \\
\cline{2-10}
 & syn & 10.94\% & \checkmark & 2.91\% & \checkmark & -48.50\%* & \checkmark & -38.41\%* & \checkmark \\
\cline{2-10}
 & rew & 12.46\% & \checkmark & 3.40\% & \checkmark & -51.93\%* & \checkmark & -42.94\%* & \checkmark \\
\cline{2-10}
 & dir & 11.04\% & \checkmark & 3.04\% & \checkmark & -44.91\%* & \checkmark & -40.36\%* & \checkmark \\
\cline{2-10}
 & eg & 13.42\% & \checkmark & 3.46\% & \checkmark & -54.62\%* & \checkmark & -43.12\%* & \checkmark \\
\midrule
\multirow{5}{*}{COMPAS (race)} & orig & 11.57\% & \checkmark & 9.51\% & \checkmark & -34.82\% & \checkmark & -28.13\% & \checkmark \\
\cline{2-10}
 & syn & 11.57\% & \checkmark & 9.51\% & \checkmark & -34.82\% & \checkmark & -28.13\% & \checkmark \\
\cline{2-10}
 & rew & 10.17\% & \checkmark & 11.80\% & \checkmark & -33.49\% & \checkmark & -31.75\% & \checkmark \\
\cline{2-10}
 & dir & 11.58\% & \checkmark & 13.19\% & \checkmark & -37.42\% & \checkmark & -34.00\% & \checkmark \\
\cline{2-10}
 & eg & 22.15\% & \checkmark & 21.66\% & \checkmark & -53.32\%* & \checkmark & -45.51\%* & \checkmark \\
\midrule
\multirow{5}{*}{MEPS (race)} & orig & 3.26\% & \checkmark & 6.51\% & \checkmark & -31.66\%* & \checkmark & -38.48\%* & \checkmark \\
\cline{2-10}
 & syn & 4.02\% & \checkmark & 7.15\% & \checkmark & -34.58\%* & \checkmark & -39.79\%* & \checkmark \\
\cline{2-10}
 & rew & 3.31\% & \checkmark & 6.10\% & \checkmark & -32.11\%* & \checkmark & -37.88\%* & \checkmark \\
\cline{2-10}
 & dir & 3.11\% & \checkmark & 6.25\% & \checkmark & -31.11\%* & \checkmark & -36.77\%* & \checkmark \\
\cline{2-10}
 & eg & 3.28\% & \checkmark & 6.15\% & \checkmark & -30.96\%* & \checkmark & -38.01\%* & \checkmark \\
\bottomrule
\end{tabular}
}
\end{table*}

Post-processing methods also exhibit unique dynamics. For example, the \textbf{CPP} mitigator consistently yields privacy risks near 0.5, the level expected from random guessing in membership inference attacks, seemingly indicating strong privacy. However, our evaluation emphasizes the balance between utility, fairness, and privacy. CPP’s main limitation is its uneven and unpredictable impact on subgroup utility. As illustrated in Figure~\ref{fig:cpp_test_accuracies}, while test accuracies decline significantly, these declines are not concentrated among underrepresented groups. In several datasets (e.g., COMPAS (Gender), COMPAS (Race), LawSchoolGPA (Gender)), minority groups occasionally retain or even improve their accuracy, whereas majority groups decline. This suggests that CPP redistributes errors in a non-systematic manner, undermining overall utility and consistency. Fairness metrics similarly shift in dataset-specific ways without a clear trend of improvement (Please refer to Appendix~\ref{appendix:cpp_graphs}). Taken together, these results show that although CPP often achieves privacy levels close to random guessing, it does not reliably preserve subgroup utility. Detailed per-group test accuracies and fairness comparisons with the original model and CPP are provided in Appendix~\ref{appendix:cpp_graphs}.

Across all datasets and models, we observe a consistent relationship between subpopulation size and privacy vulnerability: \textbf{larger groups tend to exhibit \emph{lower} membership inference risk, whereas the smallest subpopulations are the most exposed.} The distribution of samples across subpopulations is shown in Figure~\ref{fig:subpopulation_sizes}. Measuring size by each subgroup’s relative representation, we find that majority groups, such as the Privileged Unfavorable group in Bank or the Unprivileged Favorable group in MEPS19, consistently exhibit lower privacy risk than small, underrepresented subpopulations. This pattern is robust across all attacks (see Figure~\ref{fig:dt_oqta_mia_1} and Appendix~\ref{appendix:pr_graphs}) and empirically supports the conclusion of Kulynych et al.~\cite{KulynychYCVT22} that minority groups face disproportionately higher vulnerability to membership inference. Taken together, these results demonstrate that \textbf{subpopulation prevalence remains a key determinant of individual privacy risk, even in the presence of fairness-enhancing algorithms designed to reduce disparities.}

\subsection{Impact of Differential Privacy}
\label{sec:dp_impact}
To the best of our knowledge, this is the first detailed analysis of how DP affects subpopulation-level privacy risks \textbf{in the presence of various
fairness-enhancing techniques}. Our analysis reveals that the application of DP is highly effective at reducing subpopulation privacy risks, as shown in Tables~\ref{tab:utility_privacy_core_Lira}--\ref{tab:utility_privacy_core_oqta}. This pattern holds across all fairness mitigators demonstrating that DP can offer strong protection against membership inference privacy risk across the subgroups when different fairness-enhancing techniques are applied.

However, this reduction in privacy risk comes at a significant cost to model utility at the subpopulation level.  As shown in Table~\ref{tab:utility_privacy_core_Lira}, while some subgroups exhibit only modest drops or even slight improvements in test accuracy, privileged subpopulations, those denoted as \(G_0^{+}\) and \(G_1^{+}\), suffer from drastic decreases in predictive performance. In many instances, the test accuracy for these groups approaches zero (those denoted with asterisk which shows test accuracy got less than 5\%), suggesting that the addition of DP noise severely compromises the model’s ability to learn from these examples. Utility and privacy risks for all other datasets and the OT attack, together with the results of DP-SGD are reported in Appendix~\ref{appendix:utility_pr_table}, where we observe the same pattern across datasets, further strengthening our empirical findings. Figure~\ref{fig:dprf_lira_test_accuracies} illustrates how test accuracy changes across different $\epsilon$ values (1, 10, 50).

When analyzing how DP interacts with fairness mitigators, we find that almost all mitigators benefit from reduced privacy risks, each achieving \checkmark-level risk levels across all subgroups. This suggests that DP’s privacy protection is largely invariant to the chosen fairness strategy. However, utility impacts vary: \textbf{REW} tends to preserve utility more effectively for unfavorable subgroups, while \textbf{EGR} generally causes the greatest accuracy drops for favorable ones. \textbf{SYN} and \textbf{DIR} fall in between, with a dataset-specific variation. Thus, the utility cost of DP interacts non-trivially with the chosen fairness method.
When it comes to the effect on fairness, most mitigators tend to exhibit improved fairness after DP is applied-except for \textbf{REW}, whose behavior is more variable. Interestingly, the uniform reduction in subgroup accuracies caused by DP can lead to numerically better fairness metrics, as the performance gap between groups narrows. However, this comes at a cost: the improvements in fairness often reflect overall degradation, not actual performance gains. In other words, while the model may appear fairer on paper, it becomes less useful in practice.

This trade-off underscores that in certain applications, DP can significantly reduce privacy risks while preserving the model's overall performance. However, this benefit is not uniformly distributed: while most subpopulations retain acceptable levels of accuracy, certain subgroups suffer dramatic reductions in performance, sometimes to the point of exhibiting almost no learning. This divergence highlights that even when DP is tuned to achieve a favorable balance between utility and privacy at an aggregate level, its impact on certain subpopulations can be disproportionately negative.

Importantly, fairness-enhancing algorithms do not affect DP-induced utility loss uniformly. Instead, their interaction with DP is mitigator-dependent. Consistent with our analysis, \textbf{REW} tends to preserve subgroup utility more effectively for unfavorable subgroups, whereas \textbf{EGR} generally causes the largest accuracy drops for favorable subgroups. \textbf{SYN} and \textbf{DIR} typically exhibit intermediate, dataset-dependent behavior. Overall, we do not observe a uniform trend in which fairness mitigators consistently improve or consistently worsen utility under DP. Rather, the effect varies across mitigators, subgroups, and datasets, reinforcing that the interaction between DP and fairness methods is non-trivial and must be evaluated at the subpopulation level.

Additional experiments at intermediate privacy budgets ($\epsilon \in \{3, 5\}$) confirm these patterns while revealing a compounded disadvantage for underrepresented subgroups (details in Table \ref{tab:utility_privacy_eps3_eps5}): in COMPAS (Race), the underrepresented favorable group ($G_1^{+}$) retains \checkmark-level privacy under only 1 of 5 mitigators at $\epsilon = 3$ and 0 of 5 at $\epsilon = 5$, while simultaneously suffering utility drops of up to 49\%, whereas other subgroups maintain privacy protection across all $\epsilon$ values.

\subsection{Impact of Machine Learning Models} 
\label{sec:model_impact}
Our analysis shows that the privacy impact of fairness mitigation techniques is closely related to the model architecture. In particular, more complex models tend to increase privacy risks. For example, across subpopulations in the Bank and MEPS19 datasets, privacy risk generally rises in the order of Neural Networks, Decision Trees, and then Random Forests (Figures~\ref{fig:all_models_attacks},~\ref{fig:dt_oqta_mia_1}, \ref{fig:dt_lira_mia}--\ref{fig:nn_oqta_mia_1}). This trend highlights the role of model complexity and overfitting in driving membership inference vulnerability. Neural Networks, in particular, display mixed outcomes: techniques like synthetic data generation amplify subgroup risks in some cases, while other mitigators show inconsistent patterns of protection. These findings challenge the idea that fairness interventions always deteriorates privacy and instead reveal that the interplay between model choice, complexity, and mitigation strategy can create new privacy-fairness tradeoffs.

\subsection{Discussion}
\label{sec:discussion}

Our results across Sections~\ref{sec:fairness_impact}--\ref{sec:model_impact} reveal several cross-cutting patterns that merit joint consideration.

\textbf{Fairness mitigators and privacy risk.}
No single fairness-enhancing method consistently reduces privacy risk across all models and datasets. \textbf{EGR} achieves the strongest privacy-risk reductions for RF but performs poorly on simpler models (DT, NN), while \textbf{REW} offers modest but stable privacy improvements alongside reliable fairness gains. \textbf{SYN} is effective on specific datasets (e.g., Bank, COMPAS) but lacks generalizability, and \textbf{DIR} behaves inconsistently, sometimes increasing privacy risk. Post-processing via \textbf{CPP} achieves near-random-guessing privacy levels but at the cost of unpredictable subgroup utility redistribution, undermining its practical reliability. Crucially, all mitigators preserve overall accuracy within $\pm$1--3\%, yet subpopulation-level utility shifts can reach up to $\pm$33\%, highlighting that aggregate metrics mask significant subgroup disparities.

\textbf{The role of subpopulation size.}
Across all experiments, subpopulation size emerges as a persistent driver of privacy vulnerability: smaller subgroups consistently face higher membership inference risk, regardless of the mitigator or attack applied. This pattern persists even after fairness interventions, indicating that current mitigation strategies do not neutralize the structural disadvantage of underrepresentation.

\textbf{Differential Privacy: protection with uneven costs.}
DP reliably suppresses membership inference risk across all mitigators and subgroups, often to near-random-guessing levels. However, its utility cost is distributed unevenly: certain subgroups, particularly privileged and underrepresented favorable groups, experience accuracy collapses to near zero. The interaction between DP and fairness mitigators is non-trivial: \textbf{REW} tends to preserve subgroup utility more effectively under DP, while \textbf{EGR} causes the largest accuracy drops for favorable subgroups. Furthermore, apparent fairness improvements under DP often reflect uniform accuracy degradation rather than genuine gains, as performance gaps narrow primarily because all groups lose utility. Intermediate privacy budgets ($\epsilon \in \{3, 5\}$) reveal compounded disadvantages for underrepresented subgroups, who lose both privacy protection and utility simultaneously.

\textbf{Model complexity as a mediating factor.}
Privacy risk generally increases with model complexity, and this ordering interacts with the choice of mitigator. Fairness interventions that reduce privacy risk for complex models (e.g., \textbf{EGR} on RF) may amplify it for simpler ones, challenging the assumption that fairness interventions uniformly affect privacy in a single direction.

\textbf{Practical recommendations.}
Given the dataset-, model-, and subgroup-dependent nature of these trade-offs, practitioners should select strategies based on their primary objective:
\begin{itemize}
    \item \textbf{If privacy is paramount:} combine DP with lower-complexity models and conservative mitigators such as \textbf{REW}, while monitoring subgroup utility for disproportionate degradation.
    \item \textbf{If fairness is the priority:} \textbf{EGR} and \textbf{REW} offer strong fairness gains, with the caveat that \textbf{EGR} may increase privacy risk for simpler model classes.
    \item \textbf{If utility preservation is critical:} avoiding DP and relying on pre- or in-processing mitigators helps maintain predictive performance, though subpopulation-level accuracy should still be verified.
\end{itemize}
In all cases, evaluation at the subpopulation level is essential, as aggregate metrics consistently fail to capture the disparities observed across our experiments.

\section{Conclusions}
\label{sec:conclusions}

In this work, we presented a comprehensive subpopulation-level analysis of the trade-offs between fairness, privacy, and utility. By evaluating a wide range of model families, membership inference attacks, and fairness mitigation strategies, we demonstrate that aggregate metrics consistently obscure critical disparities that emerge at the subgroup level.

Our research makes three primary contributions to the field. First, we provide the first detailed assessment of the interaction between Differential Privacy (DP) and fairness-enhancing algorithms through a subpopulation lens, revealing that privacy protection often imposes an uneven utility cost on the very groups fairness interventions aim to protect. Second, we introduced a subpopulation-level adaptation of the Likelihood Ratio Attack (LiRA), enabling a more granular audit of membership inference vulnerability. Third, our findings empirically confirm that underrepresented groups bear a disproportionate privacy risk, a structural disadvantage that existing fairness mitigators do not inherently resolve.

The central takeaway of our study is that there is no universal solution for balancing fairness, privacy, and utility. These trade-offs are not static; they are deeply dependent on dataset characteristics, model architecture, and subgroup definitions. Consequently, we argue that these objectives must be evaluated jointly and at a granular level to ensure that fairness interventions do not inadvertently introduce new forms of privacy-related harm. Our framework provides a rigorous foundation for future research into optimizing these three pillars of trustworthy AI simultaneously, ensuring that the benefits of machine learning are distributed equitably across all populations.

\section{Ethical and Potential Harm Considerations}

This study follows EuroS\&P guidelines on proactive harm prevention. All experiments use publicly available benchmark datasets containing no personally identifiable information, and the attacks (LiRA, OQTA, OTA) run only in controlled, simulated settings; no deployed systems or real users are targeted. We also note that fairness interventions may unintentionally reduce subgroup utility, and our analysis discusses these risks to help practitioners avoid harmful misapplications. Our work aims to promote safer and more equitable ML systems and introduces no actions that could directly harm individuals or groups.

\section{Open Science and Data Availability}

In accordance with EuroS\&P’s Open Science expectations, we aim to make our work as transparent and reproducible as possible. All datasets used in this study are publicly available benchmark datasets, and we provide full details of preprocessing and experimental configuration to enable exact replication. The source code used to run all experiments, generate figures, and reproduce the results in this paper is released under an open license and can be accessed at: \url{https://anonymous.4open.science/r/eurosp-2026-submission-395/}. No proprietary, sensitive, or restricted data are used, and all artifacts required for reproducibility are openly shared.

\section{Acknowledgments}

Umid Suleymanov and Murat Kantarcioglu were supported in part by The Commonwealth cyber initiative grants. The authors also thank ADA University’s Center for Data Analytics Research (CeDAR) for providing computing resources.

We used AI assistance solely for improving grammar, clarity, and La\TeX{} formatting of text and mathematical expressions, in accordance with conference policies. In Section~\ref{sec:theory_synthetic}, large language models were used to help check and refine intermediate steps of some mathematical derivations; all final formulations and arguments were independently verified by the authors. The system was not used to generate scientific ideas, experimental designs, analyses, results, or claims presented in this paper.

\bibliographystyle{plain}
\bibliography{references}
\appendices
\begingroup
\raggedbottom
\setlength{\textfloatsep}{6pt plus 2pt minus 2pt}
\setlength{\floatsep}{6pt plus 2pt minus 2pt}
\setlength{\intextsep}{6pt plus 2pt minus 2pt}
\setlength{\abovecaptionskip}{2pt}
\setlength{\belowcaptionskip}{0pt}
\captionsetup[figure]{skip=2pt}
\section{Privacy Risk Graphs}
\label{appendix:pr_graphs}

\begin{figure}[H]
    \captionsetup{font=footnotesize}
    \centering
    \includegraphics[width=0.95\columnwidth]{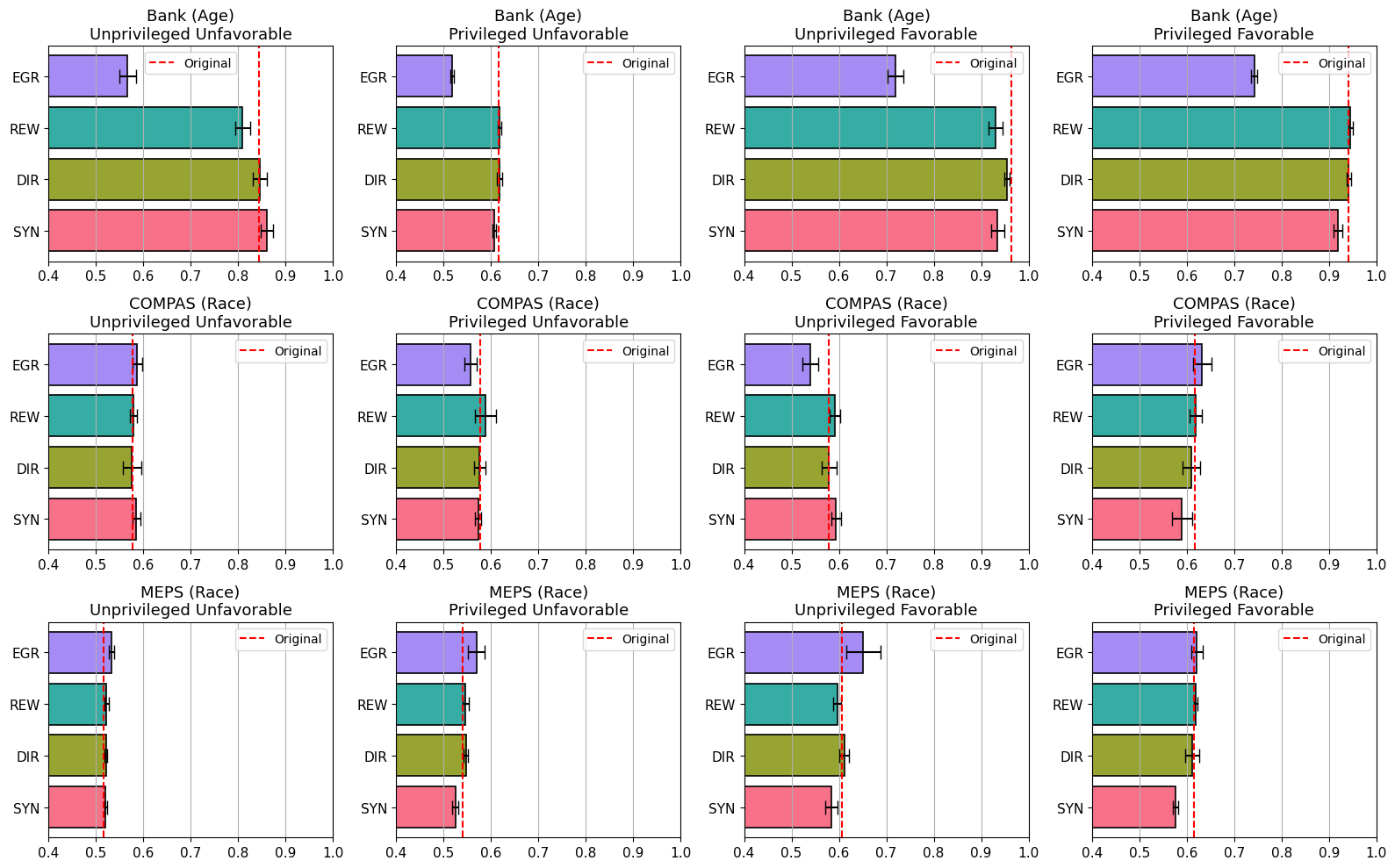} 
    \caption{Decision Tree privacy risk results under LiRA attack for: Bank (Age), COMPAS (Race), and MEPS (Race) datasets. 
    Visualizes subpopulation privacy risks across different fairness mitigation techniques. The red dashed line indicates baseline risk from the original unmitigated model. Values are averaged over 20 runs, while the standard deviation is shown with error bars. The horizontal range is dynamic for better visibility.}
    \label{fig:dt_lira_mia}
\end{figure}

\begin{figure}[H]
    \captionsetup{font=footnotesize}
    \centering
    \includegraphics[width=0.95\columnwidth]{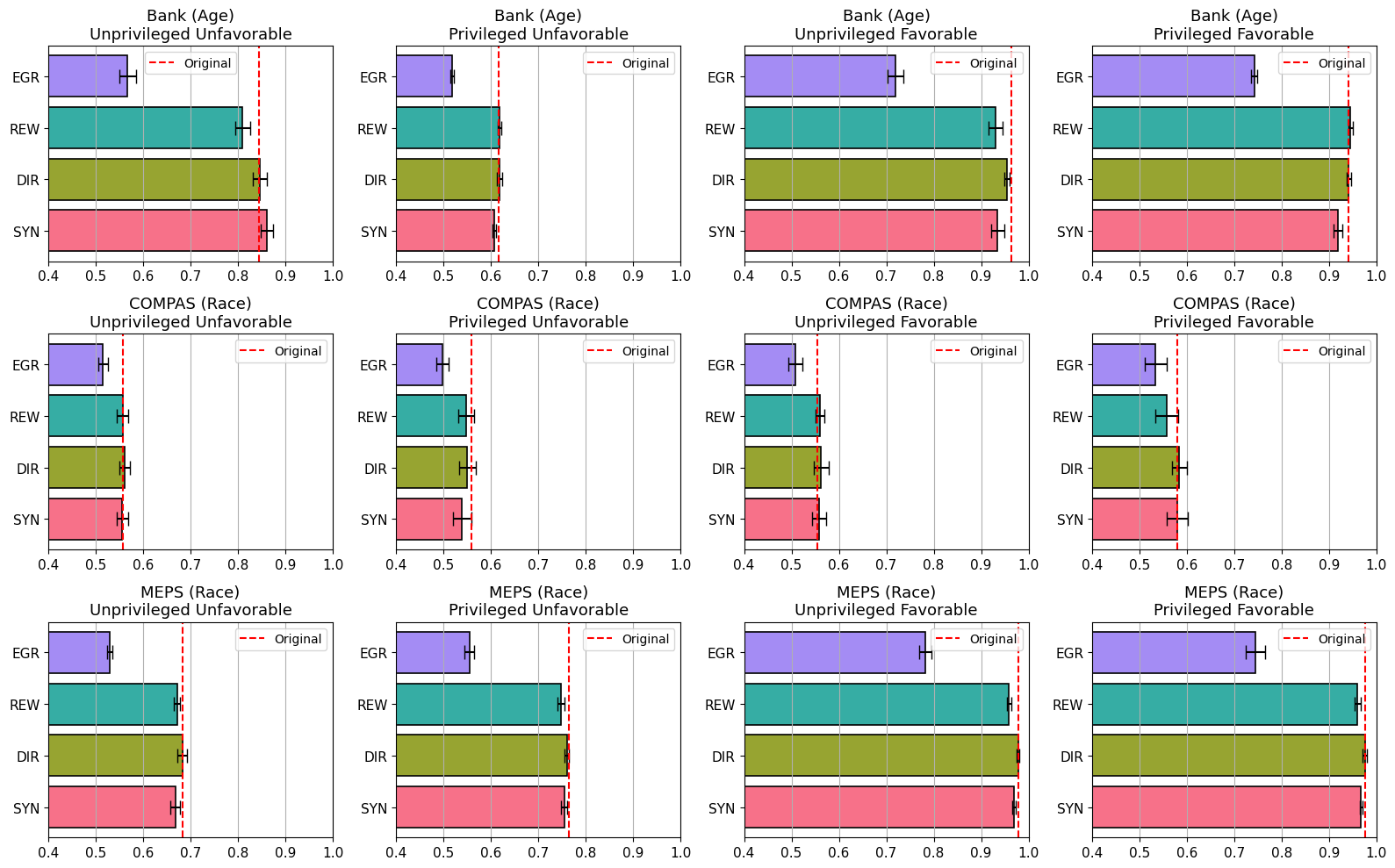}
    \caption{Random Forest privacy risk results under LiRA attack for: Bank (Age), COMPAS (Race), and MEPS (Race) datasets. 
    Visualizes subpopulation privacy risks across different fairness mitigation techniques. The red dashed line indicates baseline risk from the original unmitigated model.Values are averaged over 20 runs, while the standard deviation is shown with error bars. The horizontal range is dynamic for better visibility.}
    \label{fig:rf_lira_mia}
\end{figure}

\begin{figure}[H]
    \captionsetup{font=footnotesize}
    \centering
    \includegraphics[width=0.95\columnwidth]{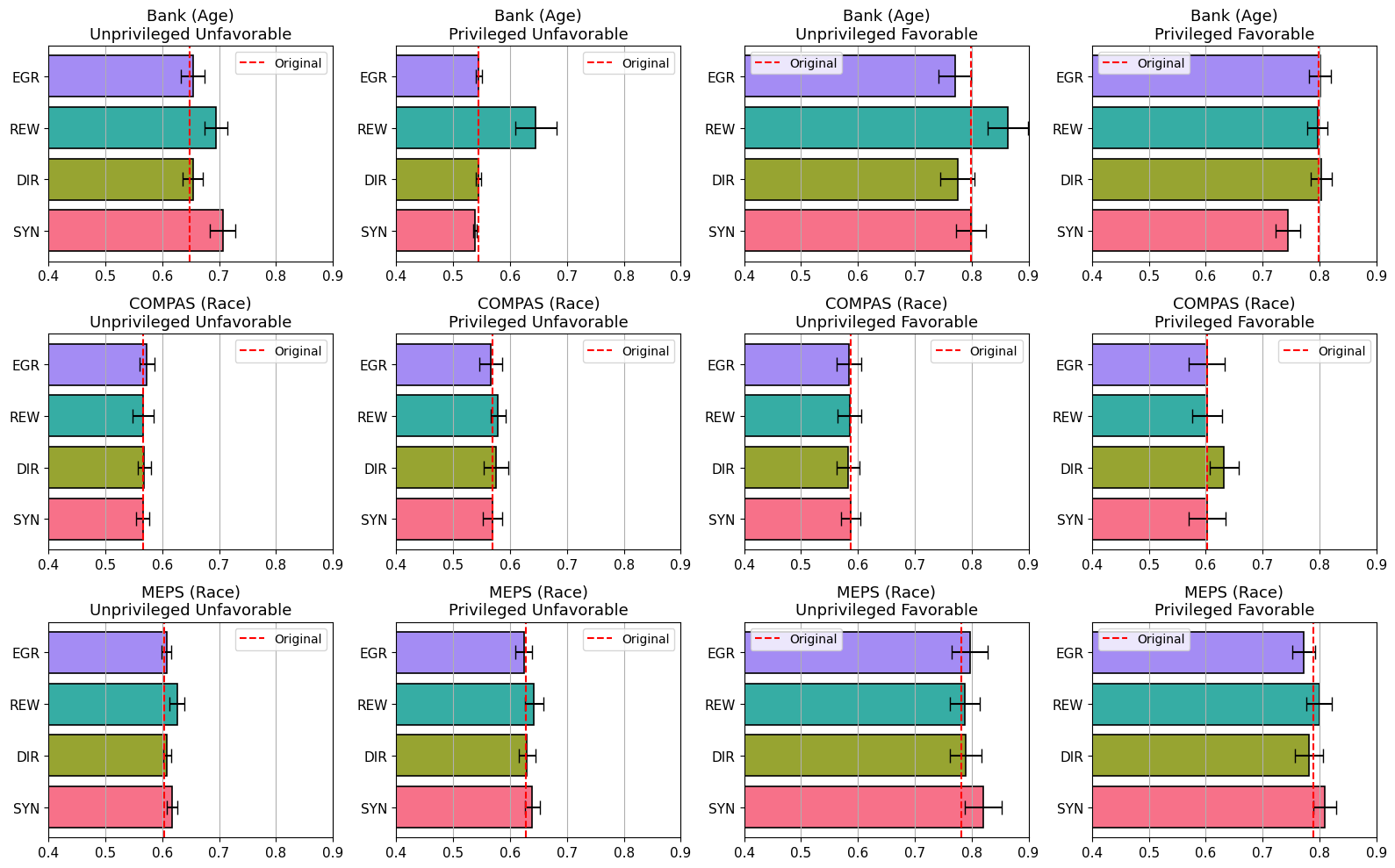} 
    \caption{Random Forest privacy risk results under OQTA for: Bank (Age), COMPAS (Race), and MEPS (Race) datasets. 
    Visualizes subpopulation privacy risks across different fairness mitigation techniques. The red dashed line indicates baseline risk from the original unmitigated model. Values are averaged over 20 runs, while the standard deviation is shown with error bars. The horizontal range is dynamic for better visibility.}
    \label{fig:rf_oqta_mia_1}
\end{figure}

\begin{figure}[H]
    \captionsetup{font=footnotesize}
    \centering
    \includegraphics[width=0.95\columnwidth]{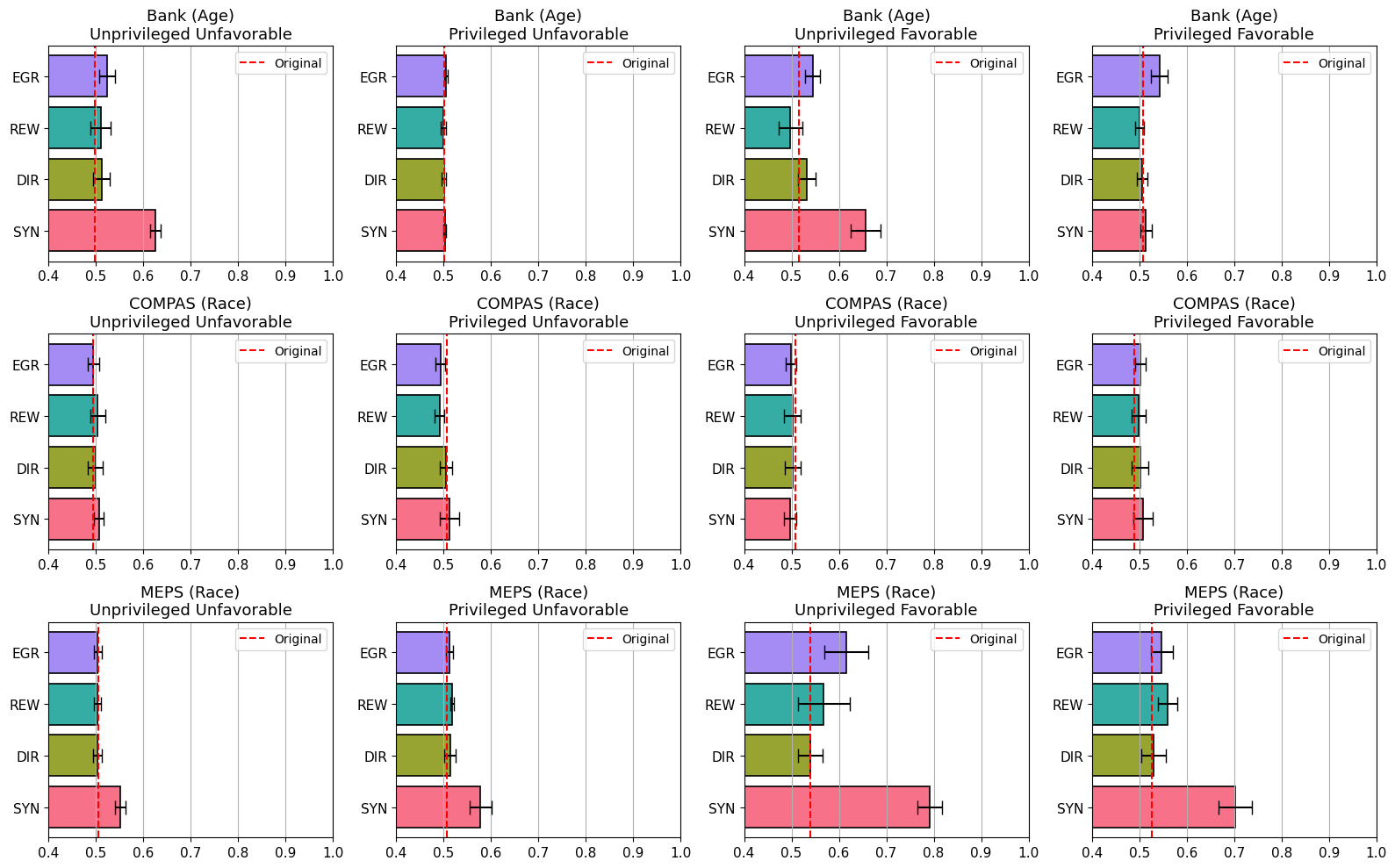}
    \caption{Neural Network privacy risk results under LiRA attack for: Bank (Age), COMPAS (Race), and MEPS (Race) datasets. 
    Visualizes subpopulation privacy risks across different fairness mitigation techniques. The red dashed line indicates baseline risk from the original unmitigated model.Values are averaged over 20 runs, while the standard deviation is shown with error bars. The horizontal range is dynamic for better visibility.}
    \label{fig:nn_lira_mia}
\end{figure}

\begin{figure}[H]
    \captionsetup{font=footnotesize}
    \centering
    \includegraphics[width=0.95\columnwidth]{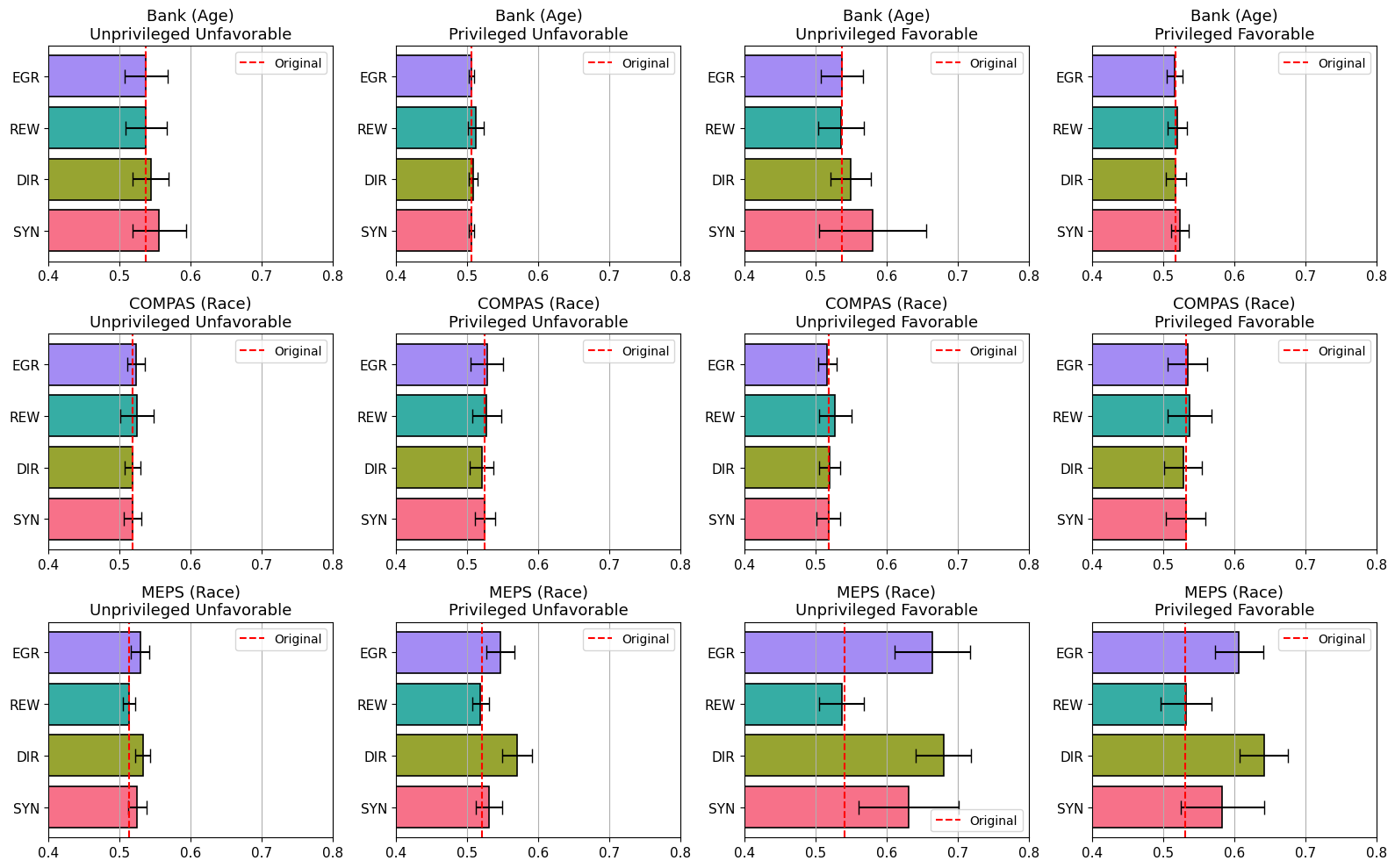} 
    \caption{Neural Network privacy risk results under OQTA for: Bank (Age), COMPAS (Race), and MEPS (Race) datasets. 
    Visualizes subpopulation privacy risks across different fairness mitigation techniques. The red dashed line indicates baseline risk from the original unmitigated model. Values are averaged over 20 runs, while the standard deviation is shown with error bars. The horizontal range is dynamic for better visibility.}
    \label{fig:nn_oqta_mia_1}
\end{figure}






\begin{figure}[H]
    \captionsetup{font=footnotesize}
    \centering
    \includegraphics[width=0.95\columnwidth]{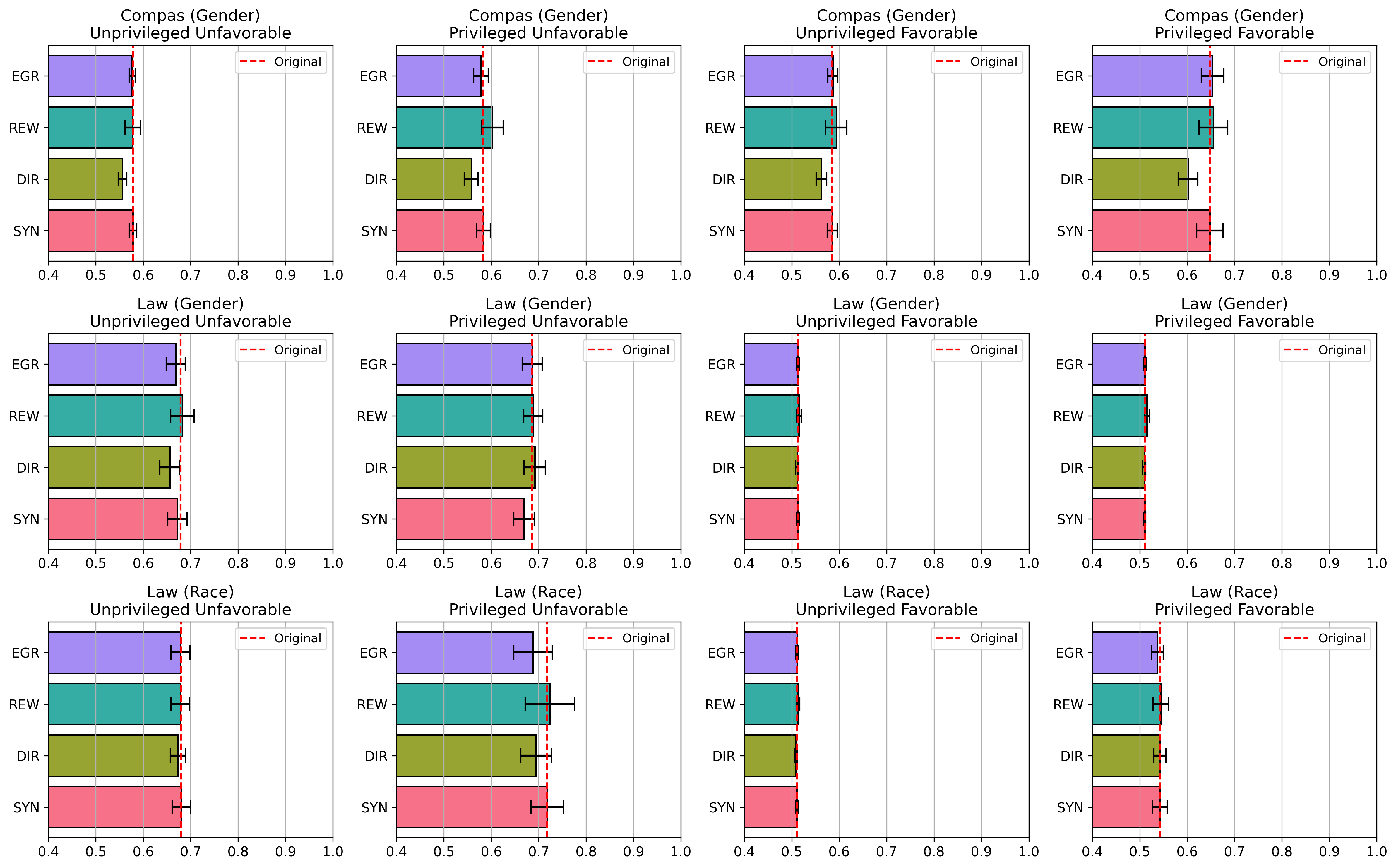} 
    \caption{Decision Tree privacy risk results under OT attack for: COMPAS (Gender), Law (Gender), and Law (Race) datasets. Visualizes subpopulation privacy risks across different fairness mitigation techniques. The red dashed line indicates baseline risk from the original unmitigated model. Values are averaged over 20 runs, while the standard deviation is shown with error bars. The horizontal range is dynamic for better visibility.}
    \label{fig:dt_ota_mia_part1}
\end{figure}






\begin{figure}[H]
    \captionsetup{font=footnotesize}
    \centering
    \includegraphics[width=0.95\columnwidth]{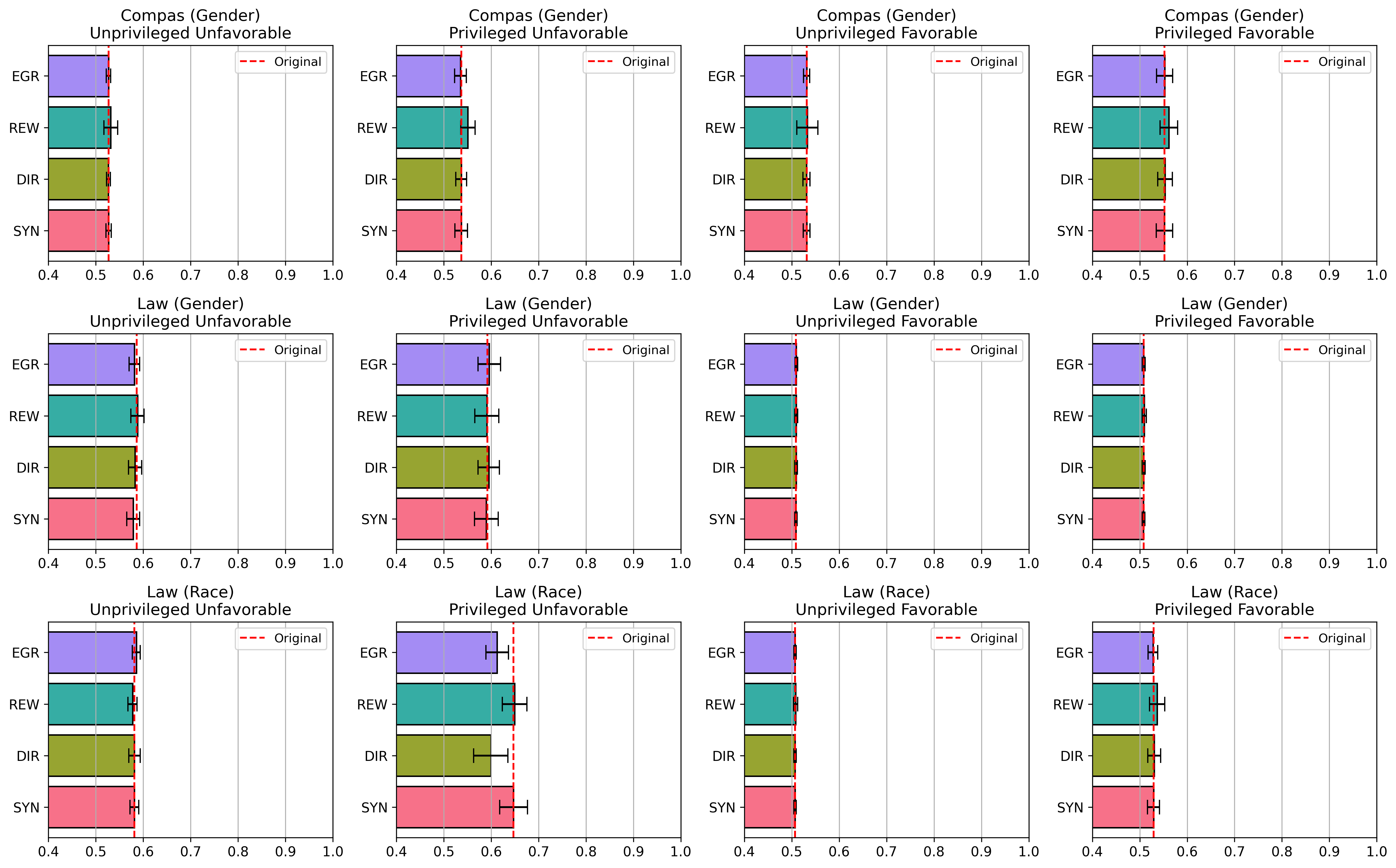} 
    \caption{Random Forest privacy risk results under OT attack for: COMPAS (Gender), Law (Gender), and Law (Race) datasets.
    Visualizes subpopulation privacy risks across different fairness mitigation techniques. The red dashed line indicates baseline risk from the original unmitigated model. Values are averaged over 20 runs, while the standard deviation is shown with error bars. The horizontal range is dynamic for better visibility.}
    \label{fig:rf_ota_mia_part1}
\end{figure}






\begin{figure}[H]
    \captionsetup{font=footnotesize}
    \centering
    \includegraphics[width=0.95\columnwidth]{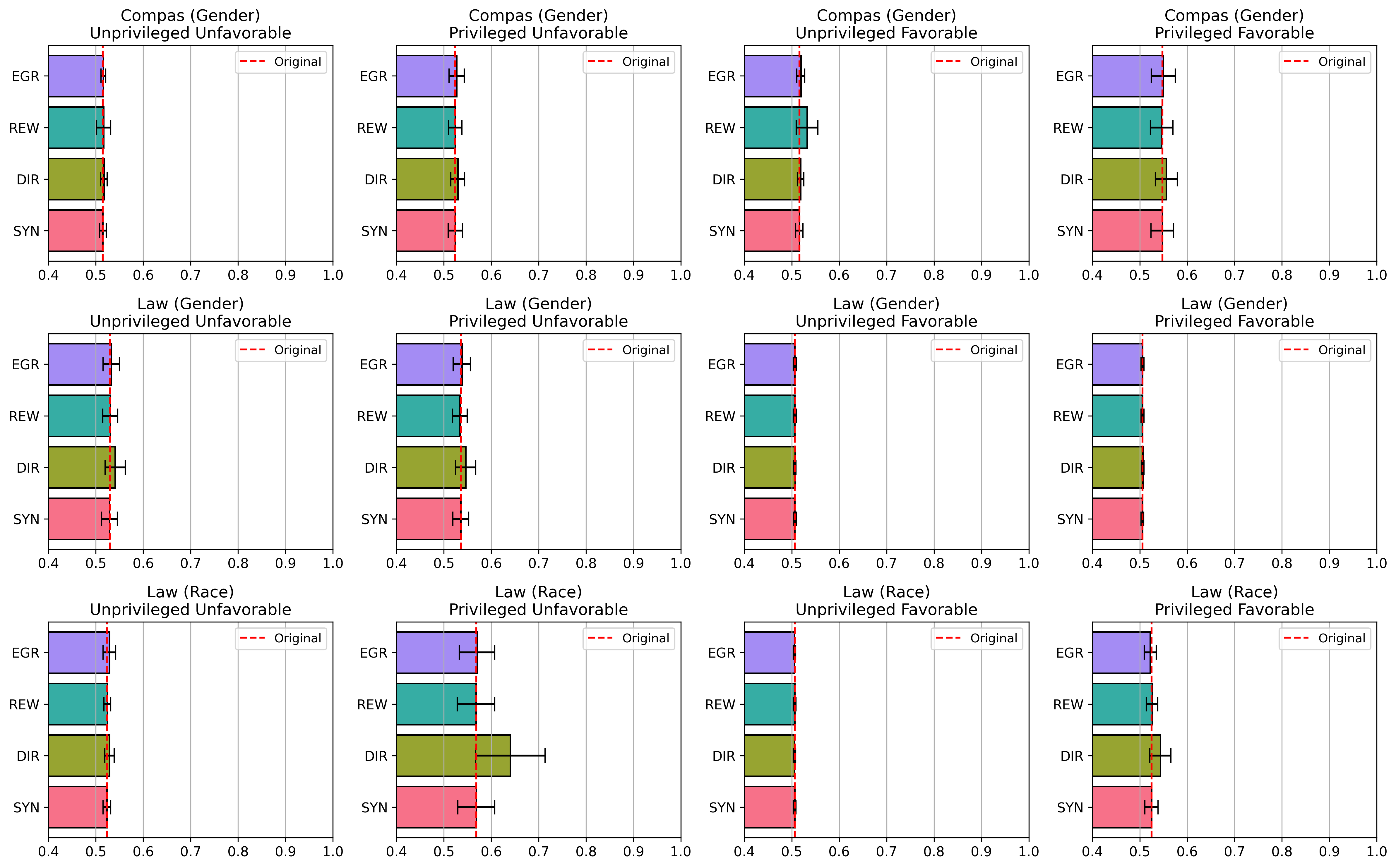} 
    \caption{Neural Network privacy risk results under OT attack for: COMPAS (Gender), Law (Gender), and Law (Race) datasets.
    Visualizes subpopulation privacy risks across different fairness mitigation techniques. The red dashed line indicates baseline risk from the original unmitigated model. Values are averaged over 20 runs, while the standard deviation is shown with error bars.  The horizontal range is dynamic for better visibility.}
    \label{fig:nn_ota_mia_part1}
\end{figure}



\begin{figure}[H]
    \captionsetup{font=footnotesize}
    \centering
    \includegraphics[width=0.95\columnwidth]{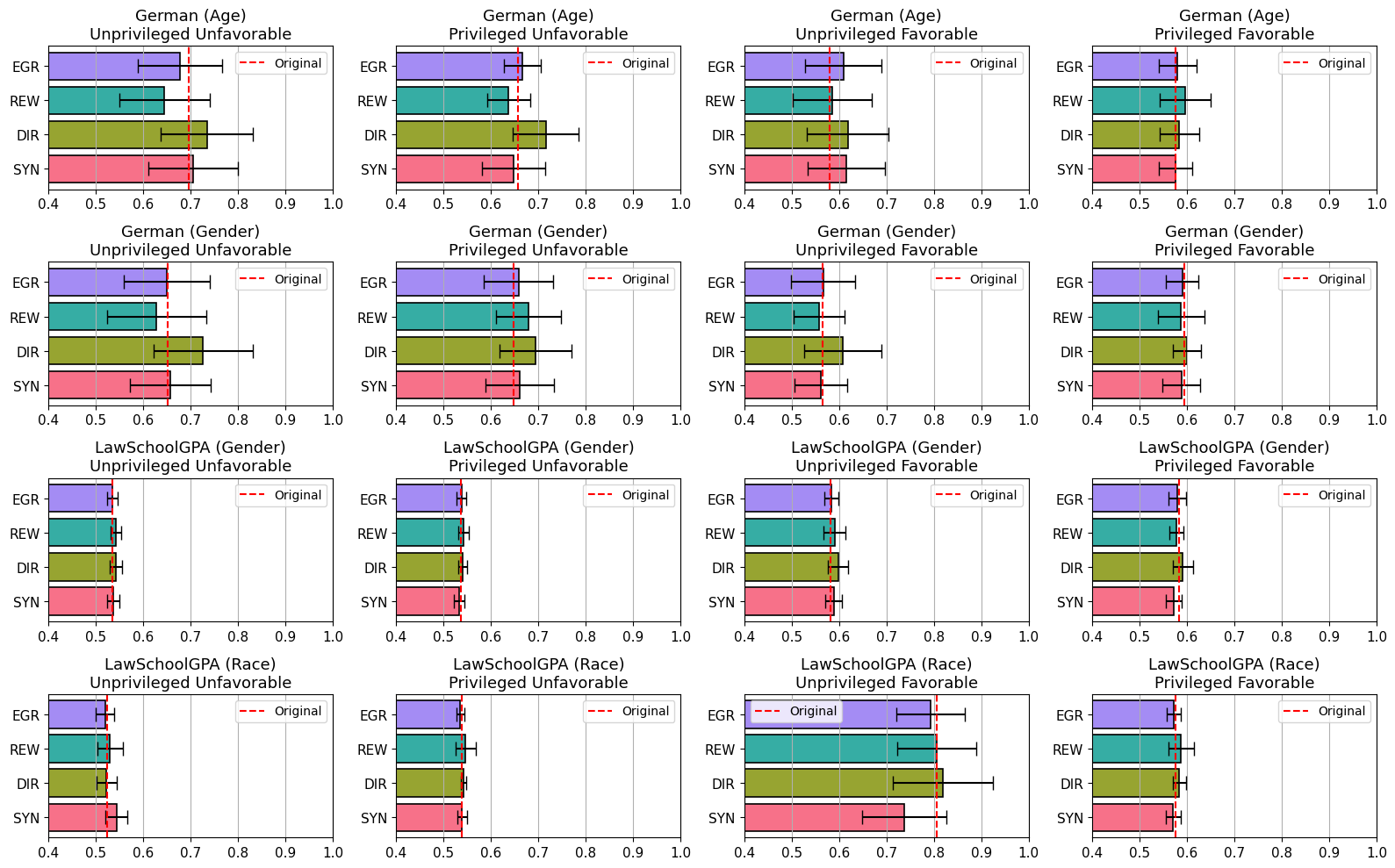}
    \caption{Decision Tree privacy risk results under OQTA for: German (Age), German (Gender), Law School GPA (Gender), and Law School GPA (Race) datasets. Visualizes subpopulation privacy risks across different fairness mitigation techniques.The red dashed line indicates baseline risk from the original unmitigated model. Values are averaged over 20 runs, while the standard deviation is shown with error bars. The horizontal range is dynamic for better visibility.}
    \label{fig:dt_oqta_mia_part2}
\end{figure}

\begin{figure}[H]
    \captionsetup{font=footnotesize}
    \centering
    \includegraphics[width=0.95\columnwidth]{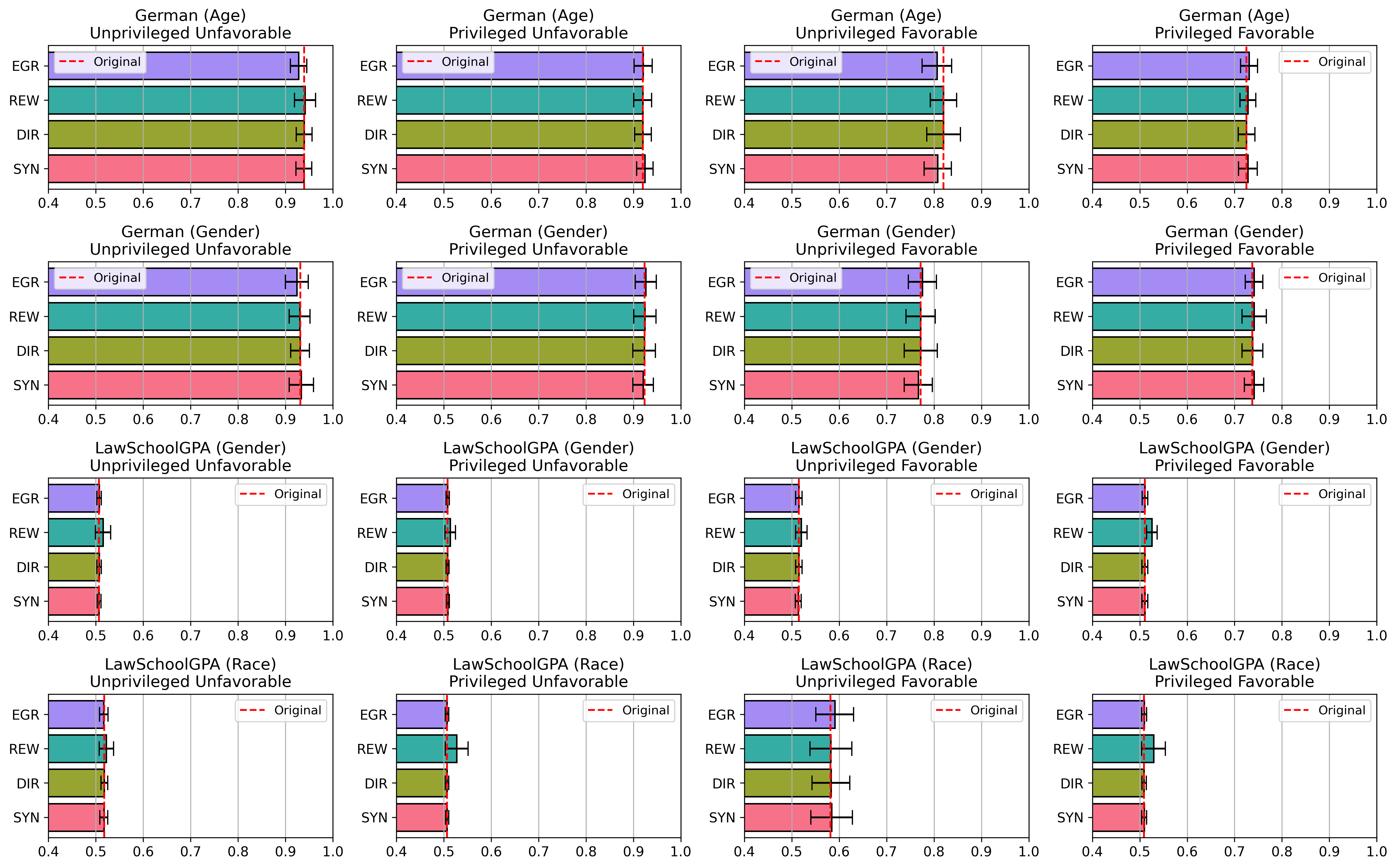} 
    \caption{Random Forest privacy risk results under OT attack for: German (Age), German (Gender), Law School GPA (Gender), and Law School GPA (Race) datasets.
    Visualizes subpopulation privacy risks across different fairness mitigation techniques. The red dashed line indicates baseline risk from the original unmitigated model. Values are averaged over 20 runs, while the standard deviation is shown with error bars. The horizontal range is dynamic for better visibility.}
    \label{fig:rf_ota_mia_part2}
\end{figure}

\begin{figure}[H]
    \captionsetup{font=footnotesize}
    \centering
    \includegraphics[width=0.95\columnwidth]{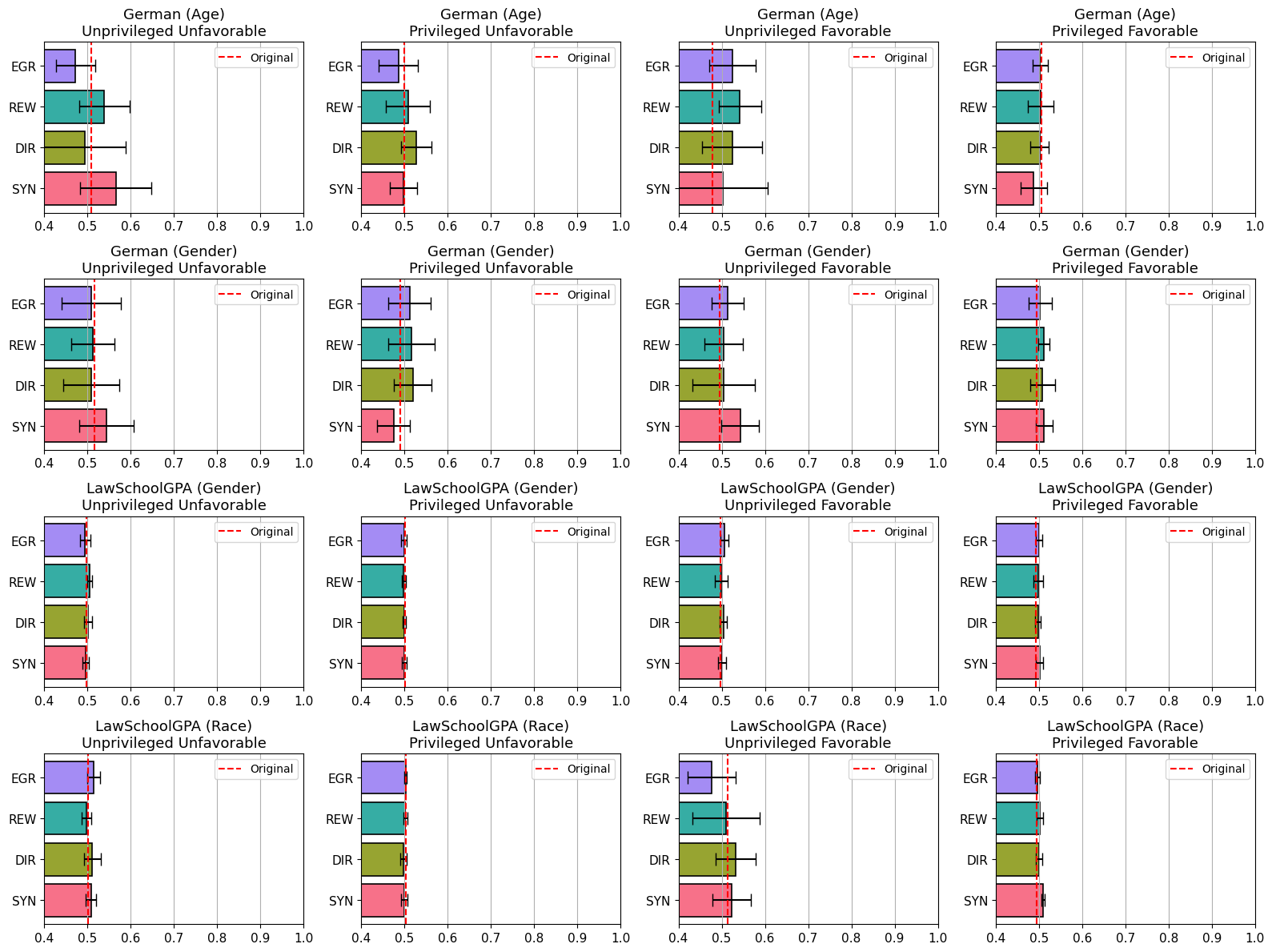} 
    \caption{Neural Network privacy risk results under LiRA attack for: German (Age), German (Gender), Law School GPA (Gender), and Law School GPA (Race) datasets.
    Visualizes subpopulation privacy risks across different fairness mitigation techniques. The red dashed line indicates baseline risk from the original unmitigated model.Values are averaged over 20 runs, while the standard deviation is shown with error bars. The horizontal range is dynamic for better visibility.}
    \label{fig:nn_lira_mia_part2}
\end{figure}

\FloatBarrier
\clearpage
\onecolumn
\section{DP Utility and Privacy Risk Graphs }
\setlength{\textfloatsep}{4pt plus 1pt minus 1pt}
\setlength{\intextsep}{4pt plus 1pt minus 1pt}
\setlength{\floatsep}{4pt plus 1pt minus 1pt}
\setlength{\abovecaptionskip}{3pt}
\setlength{\belowcaptionskip}{2pt}
\label{appendix:utility_pr_table}

\begin{table}[H] 
\centering
\renewcommand{\arraystretch}{0.95}
\setlength{\tabcolsep}{4pt}
\caption{%
Subpopulation utility difference (difference between the test accuracy for the subpopulation before and after application of DP) and privacy risks for subpopulation after the application of DP (DP - SGD vs. NN) across fairness mitigators for LiRA setup. Rows show datasets with fairness mitigators. Asterisks (*) mark low-utility subgroups (< 5\%) whereas tick (\checkmark) means that the privacy risk is minimized for the given subpopulation (privacy risk goes down to 50$\pm$2\%). Values are averaged over 20 runs, while the standard deviation is in the interval [0;0.06].
}
\label{tab:utility_privacy_sgd_LiRA}
\resizebox{\textwidth}{!}{%
\begin{tabular}{llrrrrrrrr}
\toprule
& & \multicolumn{2}{c}{$G_0^{-}$} & \multicolumn{2}{c}{$G_1^{-}$} & \multicolumn{2}{c}{$G_0^{+}$} & \multicolumn{2}{c}{$G_1^{+}$} \\
\cmidrule(lr){3-4} \cmidrule(lr){5-6} \cmidrule(lr){7-8} \cmidrule(lr){9-10}
Dataset & Mitigator & Utility Diff. & Privacy & Utility Diff. & Privacy & Utility Diff. & Privacy & Utility Diff. & Privacy \\

\midrule
\multirow{5}{*}{Synthetic (G)} 
& orig 
& -66.20\%* &  
& -71.26\% & \checkmark 
& 65.38\% & \checkmark 
& 20.60\% & \checkmark \\
\cline{2-10}

& syn 
& -34.61\% & \checkmark 
& -17.41\% & \checkmark 
& 19.68\% & \checkmark 
& -11.84\% & \checkmark \\
\cline{2-10}

& dir 
& -83.33\%* &  
& -60.66\% & \checkmark 
& 83.33\% & \checkmark 
& 31.53\% & \checkmark \\
\cline{2-10}

& rew 
& -100.00\%* & \checkmark 
& -93.39\%* & \checkmark 
& 100.00\% & \checkmark 
& 23.91\% & \checkmark \\
\cline{2-10}

& eg 
& -6.42\% &  
& -0.61\% & \checkmark 
& 0.52\% & \checkmark 
& -0.46\% & \checkmark \\

\midrule

\multirow{5}{*}{Bank (age)}

& orig
& \num[round-mode=places,round-precision=2]{\fpeval{(0.9997136311569301 - 0.8704114664271302)*100}}\% & \checkmark
& \num[round-mode=places,round-precision=2]{\fpeval{(0.9980936366432928 - 0.9580950782496679)*100}}\% & \checkmark
& \num[round-mode=places,round-precision=2]{\fpeval{(0.002367830434744933 - 0.48905513703819786)*100}}\% & \checkmark
& \num[round-mode=places,round-precision=2]{\fpeval{(0.04129061559578678 - 0.47739210538664395)*100}}\% & \checkmark \\
\cline{2-10}

& syn
& \num[round-mode=places,round-precision=2]{\fpeval{(0.8671524446040398 - 0.8596214454962899)*100}}\% & \checkmark
& \num[round-mode=places,round-precision=2]{\fpeval{(0.9578115120937546 - 0.9538168429600301)*100}}\% & \checkmark
& \num[round-mode=places,round-precision=2]{\fpeval{(0.493444013292706 - 0.5041746187949467)*100}}\%* & \checkmark
& \num[round-mode=places,round-precision=2]{\fpeval{(0.4946555867497528 - 0.4970009154715731)*100}}\%* & \checkmark \\
\cline{2-10}

& dir
& \num[round-mode=places,round-precision=2]{\fpeval{(0.8631939939889266 - 0.9064859477385524)*100}}\% & \checkmark
& \num[round-mode=places,round-precision=2]{\fpeval{(0.9719062841967271 - 0.9827595787881224)*100}}\% & \checkmark
& \num[round-mode=places,round-precision=2]{\fpeval{(0.5751130137819208 - 0.4614131267491915)*100}}\% & \checkmark
& \num[round-mode=places,round-precision=2]{\fpeval{(0.4002011223339233 - 0.26296492775932007)*100}}\% & \checkmark \\
\cline{2-10}
& rew
& \num[round-mode=places,round-precision=2]{\fpeval{(0.9535960311873305 - 0.9547360364609419)*100}}\% & \checkmark
& \num[round-mode=places,round-precision=2]{\fpeval{(0.9656954325800159 - 0.9610296665379913)*100}}\% & \checkmark
& \num[round-mode=places,round-precision=2]{\fpeval{(0.30640650426469535 - 0.32562929180431904)*100}}\% & \checkmark
& \num[round-mode=places,round-precision=2]{\fpeval{(0.46011873467738607 - 0.4701907936256982)*100}}\% & \checkmark \\
\cline{2-10}
& eg
& \num[round-mode=places,round-precision=2]{\fpeval{(0.8495095287133169 - 0.9547360364609419)*100}}\% & \checkmark
& \num[round-mode=places,round-precision=2]{\fpeval{(0.9459309495265688 - 0.9610296665379913)*100}}\% & \checkmark
& \num[round-mode=places,round-precision=2]{\fpeval{(0.48755391827455896 - 0.32562929180431904)*100}}\% & \checkmark
& \num[round-mode=places,round-precision=2]{\fpeval{(0.47908944402137243 - 0.4701907936256982)*100}}\% & \checkmark \\

\midrule
\multirow{5}{*}{COMPAS (race)}

& orig
& \num[round-mode=places,round-precision=2]{\fpeval{(0.7079245633593461 - 0.7366135628228535)*100}}\% & \checkmark
& \num[round-mode=places,round-precision=2]{\fpeval{(1.0 - 0.9144417062087489)*100}}\% & \checkmark
& \num[round-mode=places,round-precision=2]{\fpeval{(0.5297452584204271 - 0.6055848003748835)*100}}\% & \checkmark
& \num[round-mode=places,round-precision=2]{\fpeval{(0.0 - 0.262714894857636)*100}}\%* & \checkmark \\
\cline{2-10}

& syn
& \num[round-mode=places,round-precision=2]{\fpeval{(0.7079245633593461 - 0.7132847429366121)*100}}\% & \checkmark
& \num[round-mode=places,round-precision=2]{\fpeval{(1.0 - 0.8962500117331529)*100}}\% & \checkmark
& \num[round-mode=places,round-precision=2]{\fpeval{(0.5297452584204271 - 0.6103936243347395)*100}}\% & \checkmark
& \num[round-mode=places,round-precision=2]{\fpeval{(0.0 - 0.3122924817903414)*100}}\%* & \checkmark \\
\cline{2-10}

& dir
& \num[round-mode=places,round-precision=2]{\fpeval{(0.6967979721822065 - 0.7009592192237535)*100}}\% & \checkmark
& \num[round-mode=places,round-precision=2]{\fpeval{(0.8912912009622073 - 0.9106520275150273)*100}}\% & \checkmark
& \num[round-mode=places,round-precision=2]{\fpeval{(0.614802447814888 - 0.6187858193541654)*100}}\% & \checkmark
& \num[round-mode=places,round-precision=2]{\fpeval{(0.31294869640384565 - 0.2451812510691654)*100}}\% & \checkmark \\
\cline{2-10}

& rew
& \num[round-mode=places,round-precision=2]{\fpeval{(0.7739353069245515 - 0.804863018926728)*100}}\% & \checkmark
& \num[round-mode=places,round-precision=2]{\fpeval{(0.7570950167605792 - 0.8265492157033743)*100}}\% & \checkmark
& \num[round-mode=places,round-precision=2]{\fpeval{(0.5350662862384978 - 0.5016346560352442)*100}}\% & \checkmark
& \num[round-mode=places,round-precision=2]{\fpeval{(0.4681610225029824 - 0.399108028001078)*100}}\% & \checkmark \\
\cline{2-10}
& eg
& \num[round-mode=places,round-precision=2]{\fpeval{(0.9677404494740643 - 0.7421240691513695)*100}}\% & \checkmark
& \num[round-mode=places,round-precision=2]{\fpeval{(0.9728466632949385 - 0.7639953253352662)*100}}\% & \checkmark
& \num[round-mode=places,round-precision=2]{\fpeval{(0.050860916072636216 - 0.43293388903643315)*100}}\%* & \checkmark
& \num[round-mode=places,round-precision=2]{\fpeval{(0.03538839182986866 - 0.7639953253352662)*100}}\%* & \checkmark \\

\bottomrule
\end{tabular}%
}
\end{table}%

\begin{table}[H]
\centering
\renewcommand{\arraystretch}{0.95}
\setlength{\tabcolsep}{4pt}
\caption{%
Subpopulation utility difference (difference between the test accuracy for the subpopulation before and after application of DP) and privacy risks for subpopulations after the application of DP (DP-SGD vs. NN) across fairness mitigators for OQTA setup. Rows show datasets with fairness mitigators. Asterisks (*) mark low-utility subgroups (< 5\%) whereas tick (\checkmark) means that the privacy risk is minimized for the given subpopulation (privacy risk goes down to 50\(\pm\)2\%). Values are averaged over 20 runs, while the standard deviation is in the interval [0;0.06].
}
\label{tab:utility_privacy_oqta_dpsgd_2}
\resizebox{\textwidth}{!}{%
\begin{tabular}{@{}l l r r r r r r r r@{}} 
\toprule
 & & \multicolumn{2}{c}{\(G_0^{-}\)} & \multicolumn{2}{c}{\(G_1^{-}\)} & \multicolumn{2}{c}{\(G_0^{+}\)} & \multicolumn{2}{c}{\(G_1^{+}\)} \\
\cmidrule(lr){3-4} \cmidrule(lr){5-6} \cmidrule(lr){7-8} \cmidrule(lr){9-10}
Dataset & Mitigator & Utility Diff. & Privacy & Utility Diff. & Privacy & Utility Diff. & Privacy & Utility Diff. & Privacy \\
\midrule

\multirow{5}{*}{Law (race)} & orig & +0.00\%* & \checkmark & +0.00\%* &  & +0.00\% & \checkmark & +0.00\% & \checkmark \\
\cline{2-10}
  & syn & +0.00\%* &  & +0.00\%* &  & +0.00\% & \checkmark & +0.00\% & \checkmark \\
\cline{2-10}
  & rew & +0.00\%* &  & +0.00\%* &  & +0.00\% & \checkmark & +0.00\% & \checkmark \\
\cline{2-10}
  & dir & +0.00\%* & \checkmark & +0.00\%* &  & +0.00\% & \checkmark & +0.00\% & \checkmark \\
\cline{2-10}
  & eg & +0.00\%* &  & +0.00\%* &  & +0.00\% & \checkmark & +0.00\% & \checkmark \\
\midrule

\multirow{5}{*}{Law (gender)} & orig & +0.00\%* &  & +0.00\%* &  & +0.00\% & \checkmark & +0.00\% & \checkmark \\
\cline{2-10}
  & syn & +0.00\%* &  & +0.00\%* &  & +0.00\% & \checkmark & +0.00\% & \checkmark \\
\cline{2-10}
  & rew & +0.00\%* &  & +0.00\%* &  & +0.00\% & \checkmark & +0.00\% & \checkmark \\
\cline{2-10}
  & dir & +0.00\%* &  & +0.00\%* &  & +0.00\% & \checkmark & +0.00\% & \checkmark \\
\cline{2-10}
  & eg & +0.00\%* &  & +0.00\%* &  & +0.00\% & \checkmark & +0.00\% & \checkmark \\
\midrule

\multirow{5}{*}{MEPS (race)} & orig & +2.31\% & \checkmark & +9.51\% & \checkmark & -27.12\%* & \checkmark & -48.80\%* & \checkmark \\
\cline{2-10}
  & syn & +7.10\% & \checkmark & +10.77\% & \checkmark & -43.60\%* &  & -47.40\%* & \checkmark \\
\cline{2-10}
  & rew & +6.46\% & \checkmark & +10.18\% & \checkmark & -39.53\%* & \checkmark & -40.17\%* & \checkmark \\
\cline{2-10}
  & dir & +2.36\% & \checkmark & +9.38\% & \checkmark & -28.09\%* & \checkmark & -48.41\%* & \checkmark \\
\cline{2-10}
  & eg & +7.25\% & \checkmark & +11.41\% & \checkmark & -35.55\%* & \checkmark & -43.91\%* & \checkmark \\

\bottomrule
\end{tabular}}
\end{table}%

\begin{table}[H]
\centering
\renewcommand{\arraystretch}{0.95}
\setlength{\tabcolsep}{4pt}
\caption{%
Subpopulation utility difference (difference between the test accuracy for the subpopulation before and after application of DP) and privacy risks for subpopulation after the application of DP (DPRF vs. RF) across fairness mitigators for OTA setup. Rows show datasets with fairness mitigators. Asterisks (*) mark low-utility subgroups (< 5\%) whereas tick (\checkmark) means that the privacy risk is minimized for the given subpopulation (privacy risk goes down to 50\(\pm\)2\%). Values are averaged over 20 runs, while the standard deviation is in the interval [0;0.06].
}
\label{tab:utility_privacy_extended_ota}

\resizebox{\textwidth}{!}{%
\begin{tabular}{@{}l l r r r r r r r r@{}}
\toprule
 & & \multicolumn{2}{c}{\(G_0^{-}\)} & \multicolumn{2}{c}{\(G_1^{-}\)} & \multicolumn{2}{c}{\(G_0^{+}\)} & \multicolumn{2}{c}{\(G_1^{+}\)} \\
\cmidrule(lr){3-4} \cmidrule(lr){5-6} \cmidrule(lr){7-8} \cmidrule(lr){9-10}
Dataset & Mitigator & Utility Diff. & Privacy & Utility Diff. & Privacy & Utility Diff. & Privacy & Utility Diff. & Privacy \\
\midrule
\multirow{5}{*}{Synthetic (G)} & orig & -26.55\% &  & 2.16\% & \checkmark & 1.02\% &  & -10.43\% & \checkmark \\
\cline{2-10}
 & syn & -35.42\% &  & -1.61\% & \checkmark & 13.77\% & \checkmark & -15.41\% & \checkmark \\
\cline{2-10}
 & dir & -16.86\% &  & 1.13\% & \checkmark & 1.59\% & \checkmark & -9.45\% & \checkmark \\
\cline{2-10}
 & rew & -29.16\% &  & 3.44\% & \checkmark & 3.14\% &  & -11.72\% & \checkmark \\
\cline{2-10}
 & eg & -39.78\%* &  & -88.15\%* & \checkmark & 8.27\% &  & 8.80\% & \checkmark \\

\midrule
\multirow{5}{*}{Bank (age)} & orig & 10.62\% & \checkmark & 2.49\% & \checkmark & -48.10\% & \checkmark & -39.00\%* & \checkmark \\
\cline{2-10}
 & syn & 6.99\% & \checkmark & 1.94\% & \checkmark & -39.86\%* & \checkmark & -33.95\%* & \checkmark \\
\cline{2-10}
 & dir & 12.02\% &  & 2.38\% &  & -50.37\% &  & -37.43\%* & \\
\cline{2-10}
 & rew & 7.74\% & \checkmark & 2.61\% & \checkmark & -43.30\% & \checkmark & -40.20\%* & \checkmark \\
\cline{2-10}
 & eg & 8.05\% & \checkmark & 1.83\% & \checkmark & -44.52\%* & \checkmark & -41.64\%* & \checkmark \\
\midrule

\multirow{5}{*}{COMPAS (race)} & orig & 8.19\% & \checkmark & 10.86\% & \checkmark & -27.48\% & \checkmark & -33.29\% & \checkmark \\
\cline{2-10}
 & syn & 8.19\% & \checkmark & 10.86\% & \checkmark & -27.48\% & \checkmark & -33.29\% & \checkmark \\
\cline{2-10}
 & dir & 7.73\% & \checkmark & 13.92\% & \checkmark & -28.70\% & \checkmark & -37.64\% & \checkmark \\
\cline{2-10}
 & rew & 6.14\% & \checkmark & 13.80\% & \checkmark & -25.25\% & \checkmark & -37.38\% & \checkmark \\
\cline{2-10}
 & eg & 23.69\% & \checkmark & 24.78\% & \checkmark & -57.67\%* & \checkmark & -54.28\%* & \checkmark \\


\bottomrule
\end{tabular}}
\end{table}%

\begin{table}[!t]
\centering
\renewcommand{\arraystretch}{0.95}
\setlength{\tabcolsep}{4pt}
\caption{
Subpopulation utility difference and privacy risks for DPRF vs.\ RF at intermediate privacy budgets ($\epsilon \in \{3, 5\}$) across fairness mitigators (OQTA attack). Asterisks (*) mark low-utility subgroups ($< 5\%$); tick (\checkmark) indicates minimized privacy risk ($50\pm2\%$).
}
\label{tab:utility_privacy_eps3_eps5}

\resizebox{\textwidth}{!}{%
\begin{tabular}{lllrrrrrrrr}
\toprule
 & & & \multicolumn{2}{c}{$G_0^{-}$} & \multicolumn{2}{c}{$G_1^{-}$} & \multicolumn{2}{c}{$G_0^{+}$} & \multicolumn{2}{c}{$G_1^{+}$} \\
\cmidrule(lr){4-5} \cmidrule(lr){6-7} \cmidrule(lr){8-9} \cmidrule(lr){10-11}
Dataset & $\epsilon$ & Mitigator & Utility Diff. & Privacy & Utility Diff. & Privacy & Utility Diff. & Privacy & Utility Diff. & Privacy \\
\midrule
\multirow{10}{*}{COMPAS (Race)}
 & \multirow{5}{*}{3} & orig & +11.15\% & \checkmark & +9.73\% &  & -32.35\% & \checkmark & -29.05\% &  \\
\cline{3-11}
 & & syn & +11.15\% & \checkmark & +9.73\% &  & -32.35\% & \checkmark & -29.05\% &  \\
\cline{3-11}
 & & rew & +9.74\% & \checkmark & +12.01\% &  & -31.02\% & \checkmark & -32.66\% &  \\
\cline{3-11}
 & & dir & +10.33\% & \checkmark & +14.02\% &  & -34.75\% & \checkmark & -35.15\% & \checkmark \\
\cline{3-11}
 & & eg & +24.01\% & \checkmark & +24.31\% &  & -56.24\%* & \checkmark & -48.68\%* &  \\
\cmidrule(l){2-11}
 & \multirow{5}{*}{5} & orig & +8.81\% & \checkmark & +9.56\% & \checkmark & -30.81\% & \checkmark & -27.84\% &  \\
\cline{3-11}
 & & syn & +8.81\% & \checkmark & +9.56\% & \checkmark & -30.81\% & \checkmark & -27.84\% &  \\
\cline{3-11}
 & & rew & +7.40\% & \checkmark & +11.84\% & \checkmark & -29.47\% & \checkmark & -31.45\% &  \\
\cline{3-11}
 & & dir & +10.26\% & \checkmark & +12.88\% & \checkmark & -34.47\% & \checkmark & -32.93\% &  \\
\cline{3-11}
 & & eg & +20.78\% & \checkmark & +21.39\% & \checkmark & -52.47\%* & \checkmark & -45.16\%* &  \\
\midrule
\multirow{10}{*}{Synthetic (G)}
 & \multirow{5}{*}{3} & orig & -25.76\%* &  & +4.08\% &  & +3.52\% &  & -0.63\% & \checkmark \\
\cline{3-11}
 & & syn & -41.75\% &  & -1.59\% & \checkmark & +28.86\% & \checkmark & +5.49\% & \checkmark \\
\cline{3-11}
 & & rew & -29.69\%* &  & +10.05\% &  & +4.79\% &  & -4.93\% & \checkmark \\
\cline{3-11}
 & & dir & -20.62\%* &  & +8.58\% &  & +1.87\% & \checkmark & -12.46\% &  \\
\cline{3-11}
 & & eg & -30.06\%* &  & -55.25\%* & \checkmark & +12.70\% &  & +22.10\% & \checkmark \\
\cmidrule(l){2-11}
 & \multirow{5}{*}{5} & orig & -25.76\%* &  & +4.20\% &  & +3.52\% &  & -0.60\% & \checkmark \\
\cline{3-11}
 & & syn & -41.75\% &  & -1.55\% & \checkmark & +28.86\% & \checkmark & +5.49\% & \checkmark \\
\cline{3-11}
 & & rew & -29.69\%* &  & +10.17\% &  & +4.79\% &  & -4.90\% & \checkmark \\
\cline{3-11}
 & & dir & -20.62\%* &  & +8.58\% &  & +1.87\% & \checkmark & -12.46\% &  \\
\cline{3-11}
 & & eg & -30.06\%* &  & -55.25\%* & \checkmark & +12.70\% &  & +22.10\% & \checkmark \\
\bottomrule
\end{tabular}%
}
\end{table}%

\FloatBarrier

\twocolumn
\section{Fairness Graphs}

\begin{figure}[H]
    \captionsetup{font=footnotesize}
    \centering
    \includegraphics[width=0.95\columnwidth]{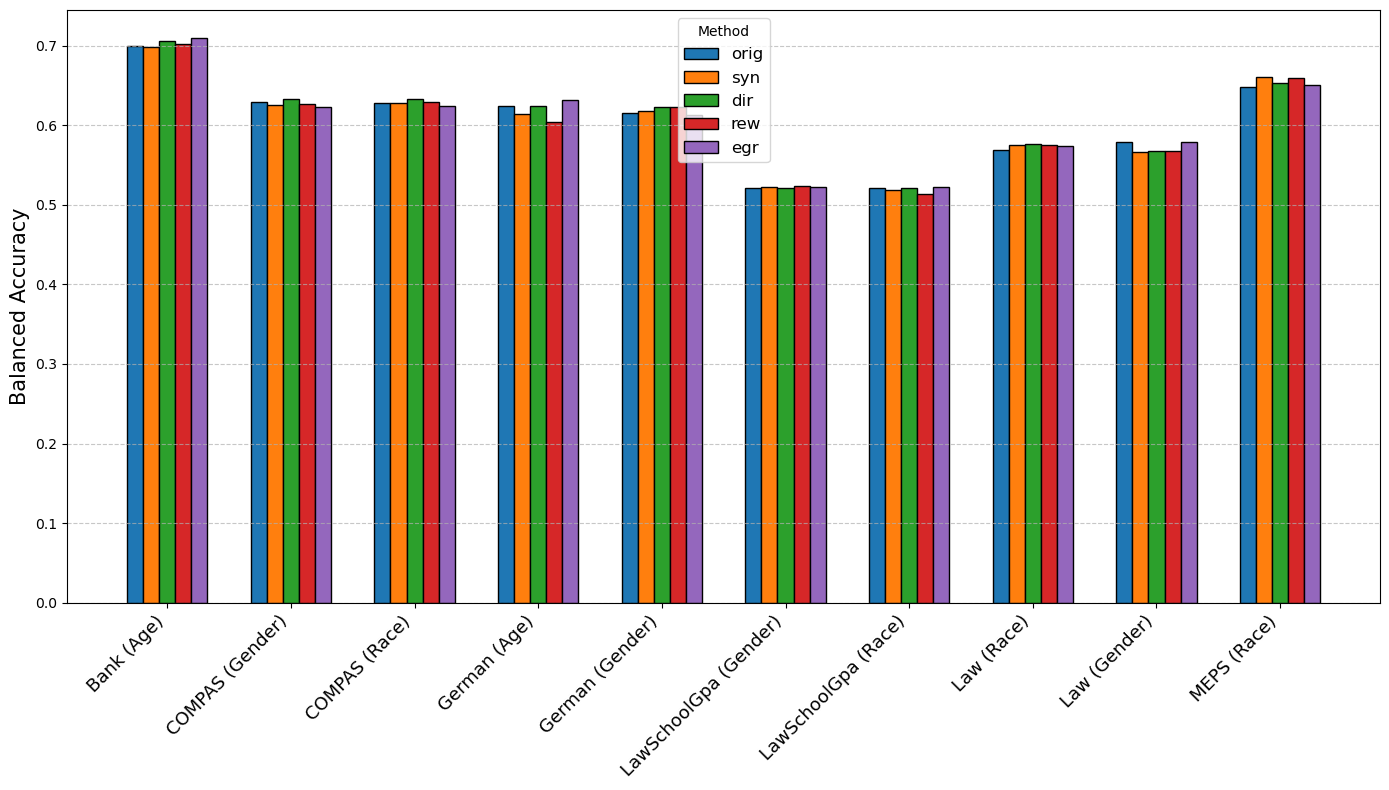} 
    \caption{Decision Tree fairness metric (balanced accuracy) results under LiRA attack for all datasets.}
    \label{fig:dt_lira_fairness_ba}
\end{figure}

\begin{figure}[H]
    \captionsetup{font=footnotesize}
    \centering
    \includegraphics[width=0.95\columnwidth]{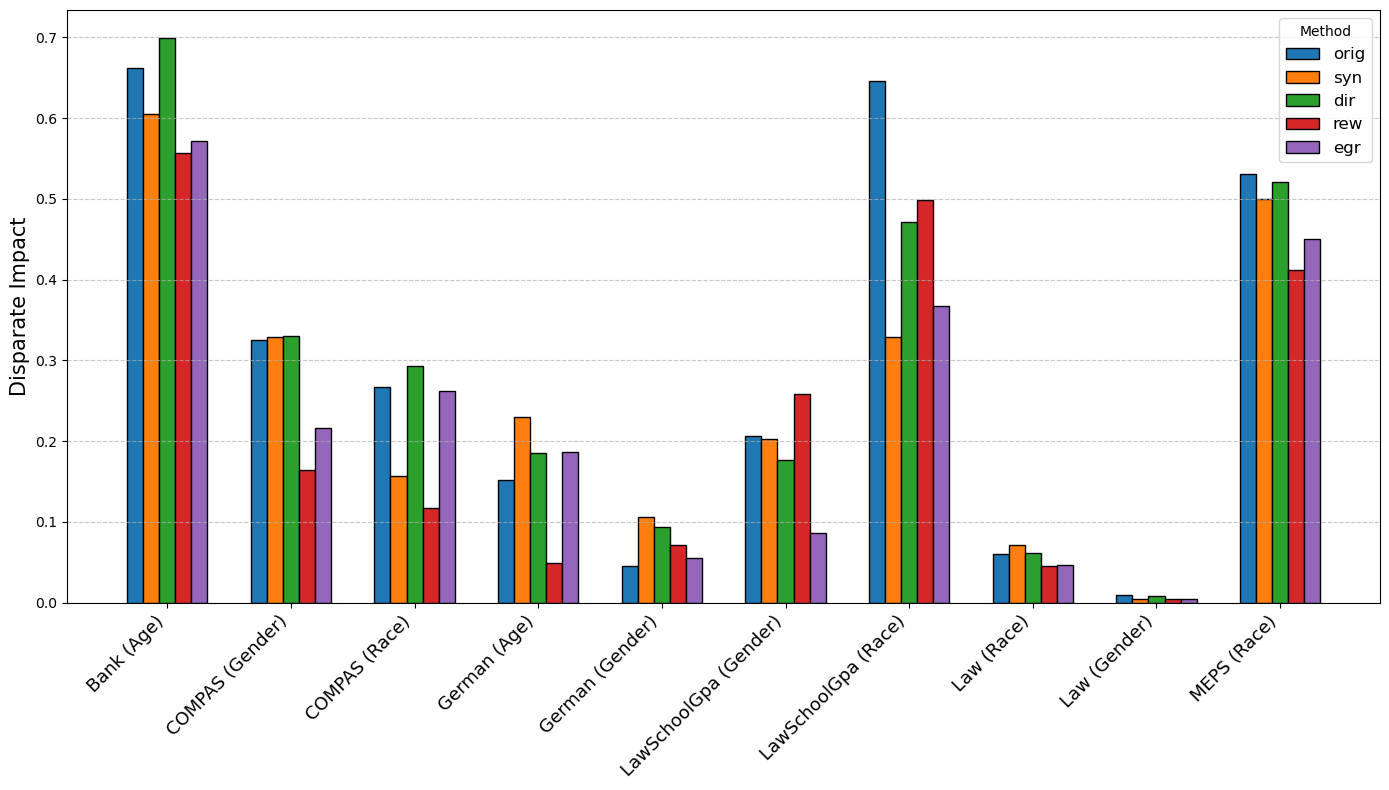} 
    \caption{Decision Tree fairness metric (disparate impact) results under LiRA attack for all datasets. Lower values indicate better fairness.}
    \label{fig:dt_lira_fairness_di}
\end{figure}

\begin{figure}[H]
    \captionsetup{font=footnotesize}
    \centering
    \includegraphics[width=0.95\columnwidth]{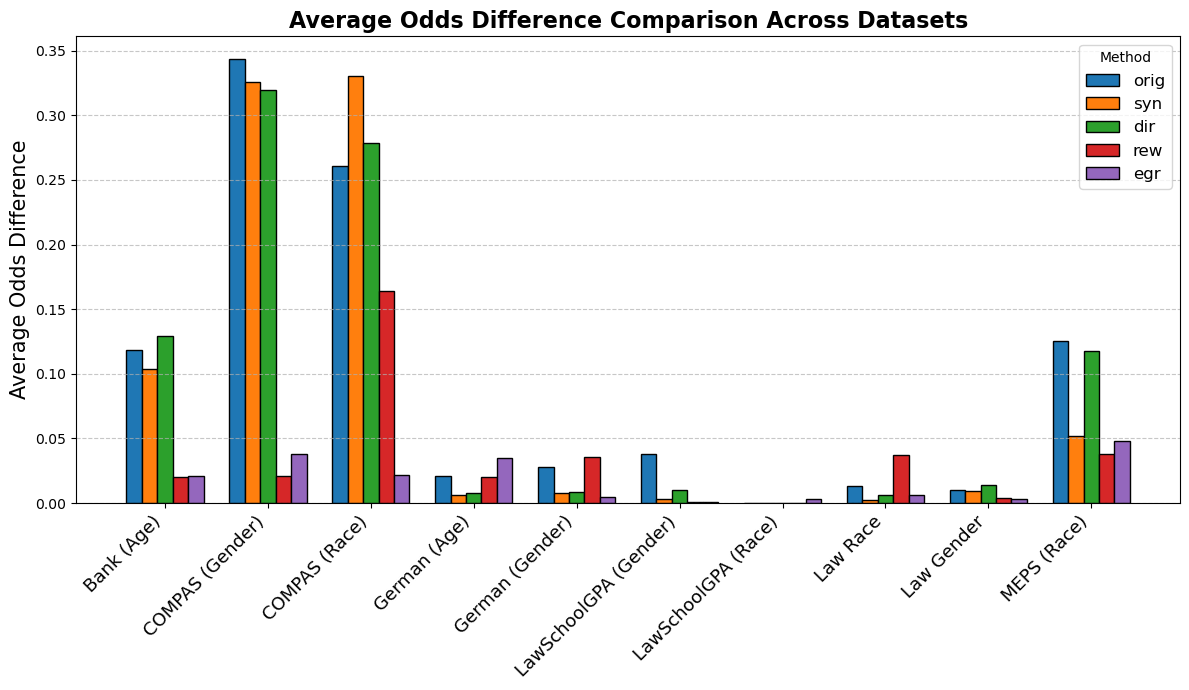} 
    \caption{Neural Network fairness metric (average odds difference) results under LiRA attack for all datasets. Lower values indicate better fairness.}
    \label{fig:nn_lira_fairness_aod}
\end{figure}

\begin{figure}[H]
    \captionsetup{font=footnotesize}
    \centering
    \includegraphics[width=0.95\columnwidth]{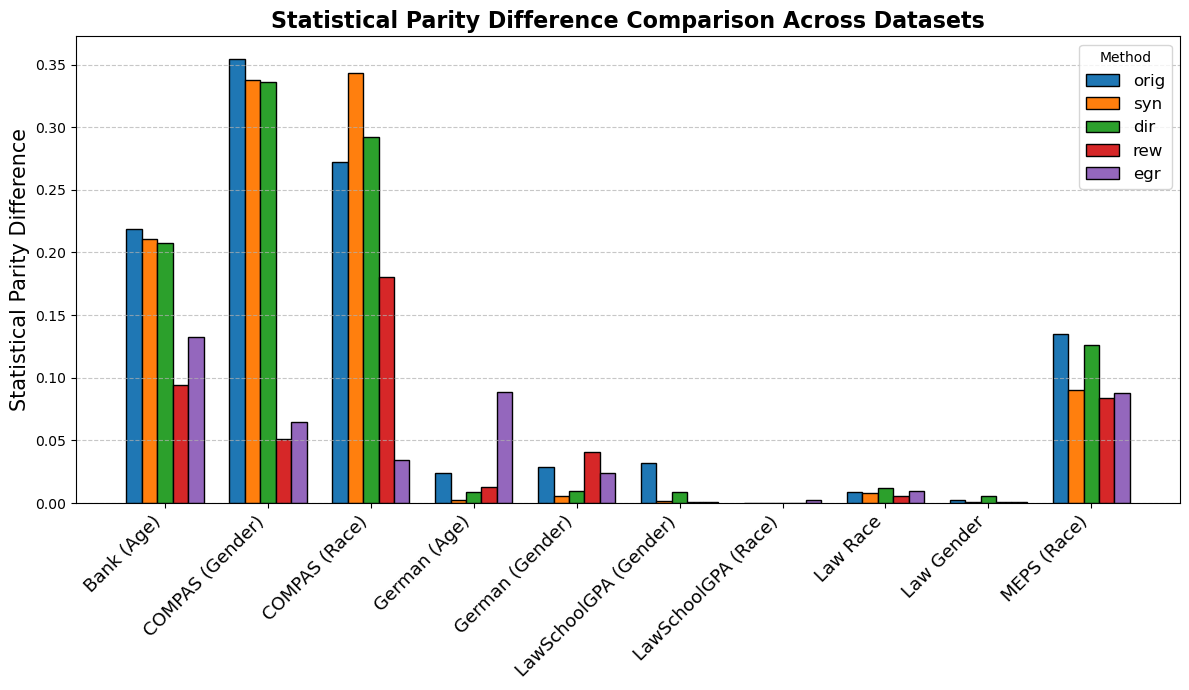} 
    \caption{Neural Network fairness metric (statistical parity difference) results under LiRA attack for all datasets. Lower values indicate better fairness.}
    \label{fig:nn_lira_fairness_spd}
\end{figure}

\begin{figure}[H]
    \captionsetup{font=footnotesize}
    \centering
    \includegraphics[width=0.95\columnwidth]{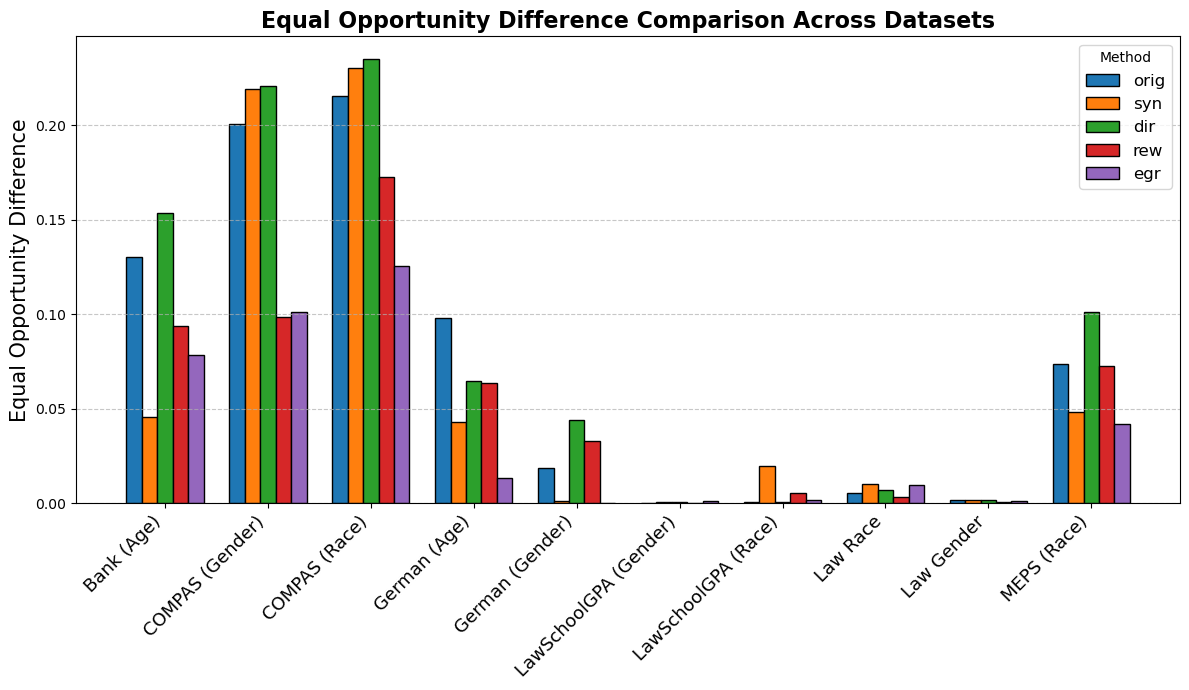} 
    \caption{Random Forest fairness metric (equal opportunity difference) results under LiRA attack for all datasets. Lower values indicate better fairness.}
    \label{fig:rf_lira_fairness_eod}
\end{figure}

\begin{figure}[H]
    \captionsetup{font=footnotesize}
    \centering
    \includegraphics[width=0.95\columnwidth]{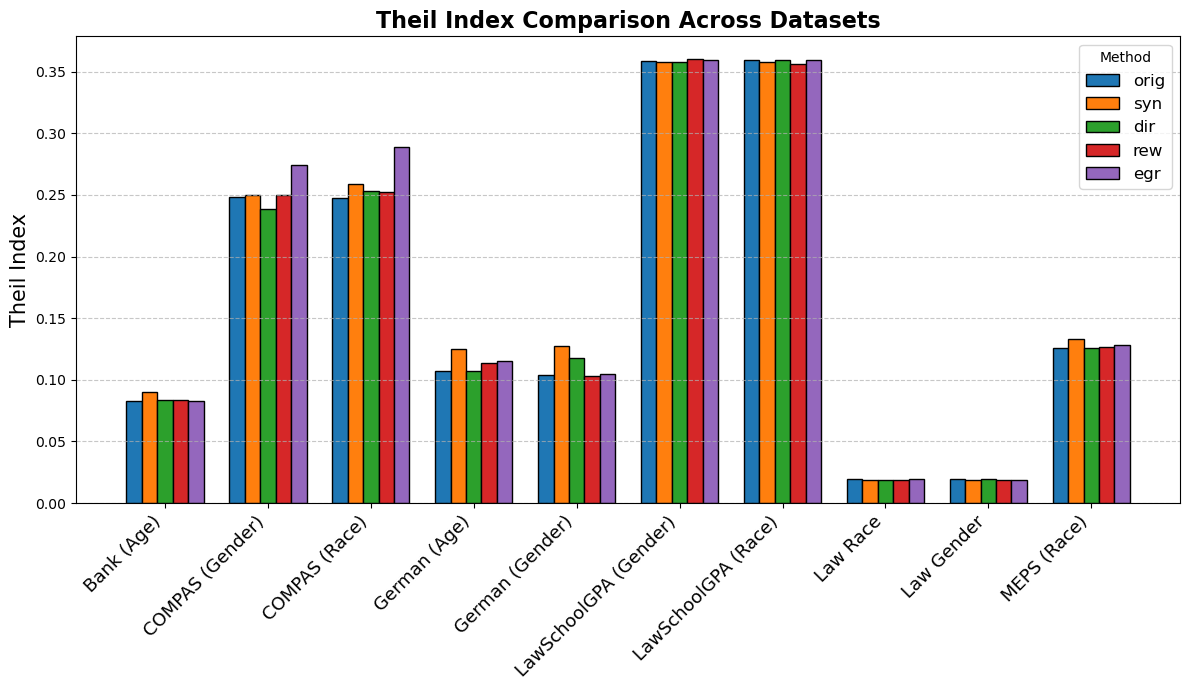} 
    \caption{Random Forest fairness metric (theil index) results under LiRA attack for all datasets. Lower values indicate better fairness.}
    \label{fig:rf_lira_fairness_ti}
\end{figure}

\begin{figure}[H]
    \captionsetup{font=footnotesize}
    \centering
    \includegraphics[width=0.95\columnwidth]{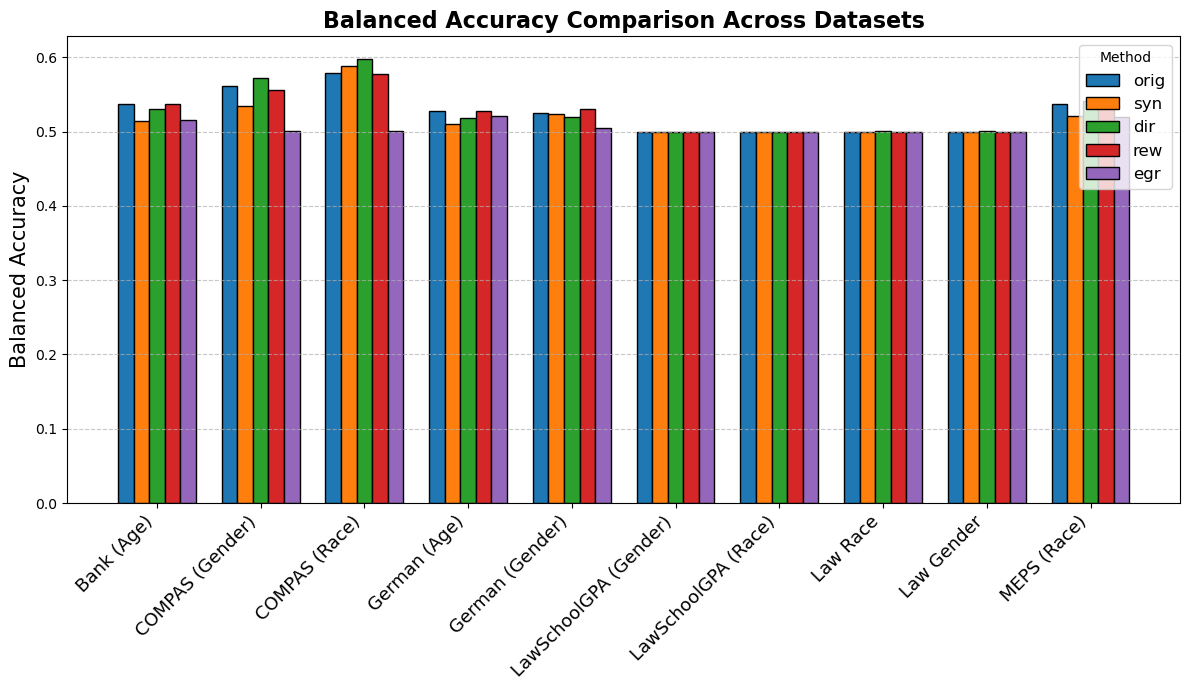} 
    \caption{DPRF fairness metric (balanced accuracy) results under LiRA attack for all datasets.}
    \label{fig:dprf_e1_lira_fairness_ba}
\end{figure}

\begin{figure}[H]
    \captionsetup{font=footnotesize}
    \centering
    \includegraphics[width=0.95\columnwidth]{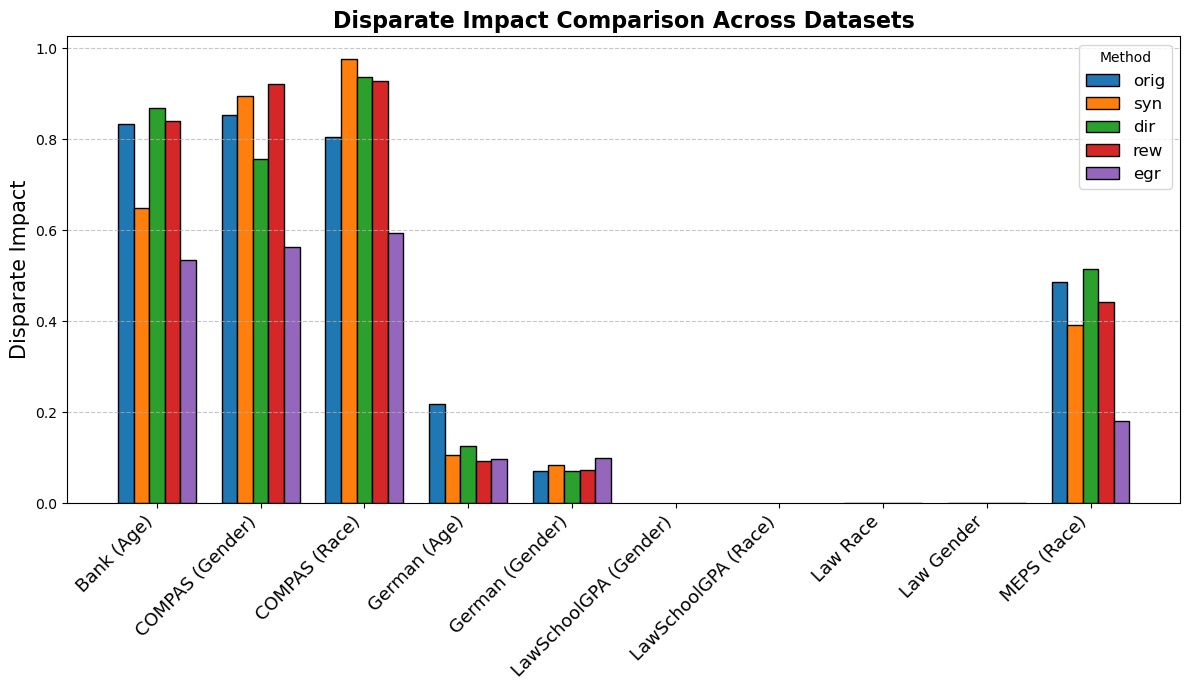} 
    \caption{DPRF fairness metric (disparate impact) results under LiRA attack for all datasets. Lower values indicate better fairness.}
    \label{fig:dprf_e1_lira_fairness_di}
\end{figure}

\twocolumn
\section{CPP Graphs}
\label{appendix:cpp_graphs}
\FloatBarrier
\subsection{CPP Fairness Figures (OTA)}

\begin{figure}[H]
  \centering
  \includegraphics[width=\columnwidth]{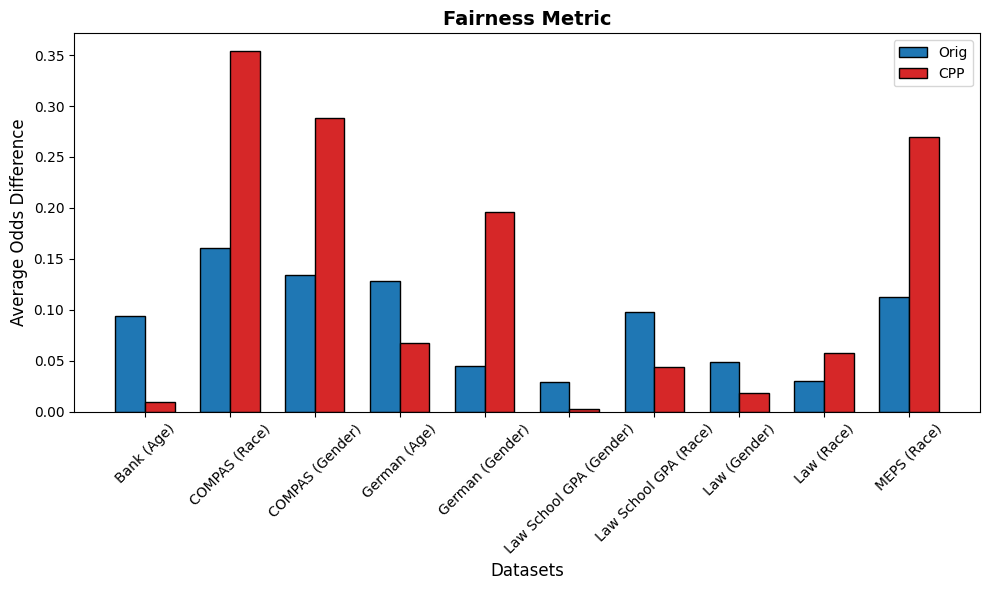}
  \caption{\textbf{Average Odds Difference (absolute).} Lower is better. Bars compare the original model (Orig) vs.\ the post-processed model (CPP) across datasets.}
  \label{fig:cpp-fairness-aod}
\end{figure}

\begin{figure}[H]
  \centering
  \includegraphics[width=\columnwidth]{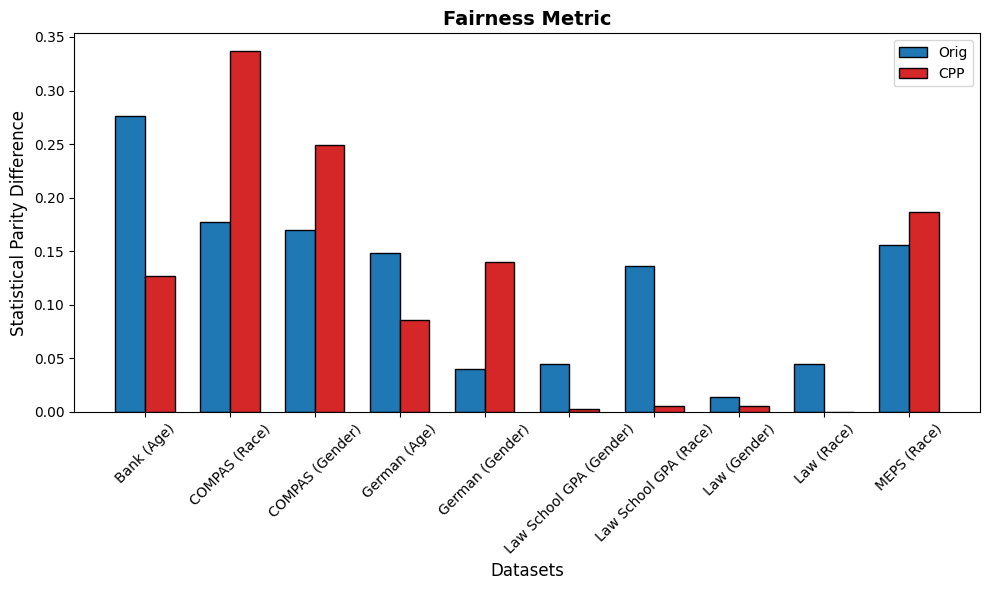}
  \caption{\textbf{Statistical Parity Difference (absolute).} Lower is better. Absolute gaps are shown for quick comparison; signed values (bias direction) are available on request or can be added to the appendix if needed.}
  \label{fig:cpp-fairness-spd}
\end{figure}

\begin{figure}[H]
  \centering
  \includegraphics[width=\columnwidth]{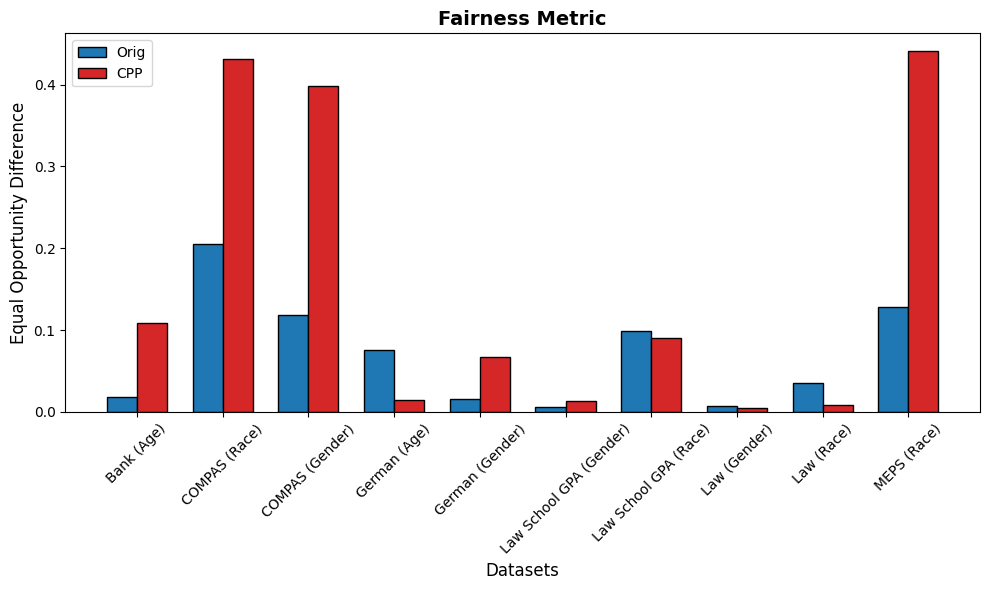}
  \caption{\textbf{Equal Opportunity Difference (absolute).} Lower is better.}
  \label{fig:cpp-fairness-eod}
\end{figure}

\begin{figure}[H]
\vspace{11mm}
  \centering
  \includegraphics[width=\columnwidth]{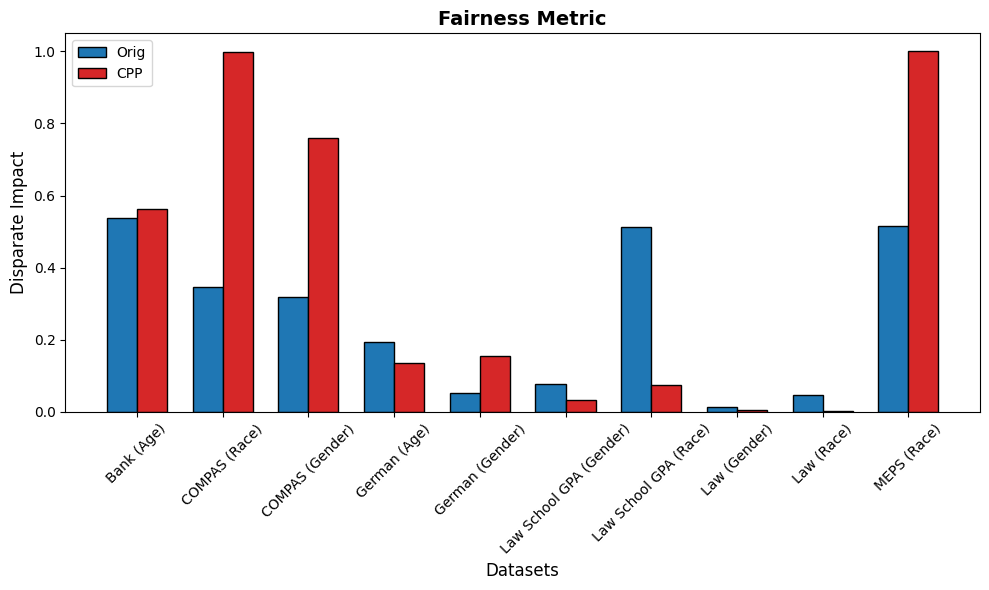}
  \caption{\textbf{Disparate Impact (absolute, transformed).} Lower is better (0 indicates parity under our definition).}
  \label{fig:cpp-fairness-di}
\end{figure}

\begin{figure}[H]
\vspace{2mm}
  \centering
  \includegraphics[width=\columnwidth]{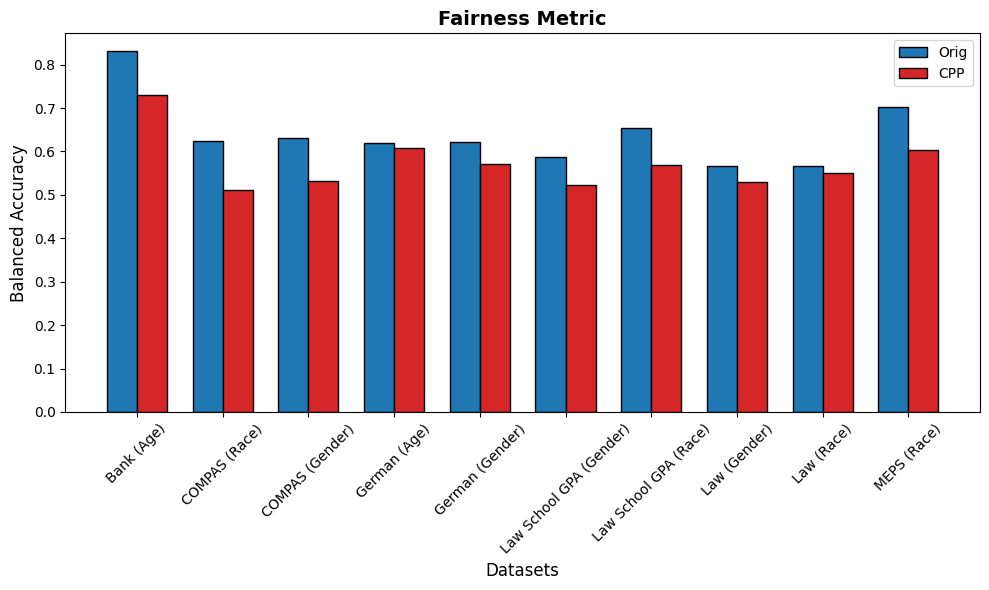}
  \caption{\textbf{Balanced Accuracy (raw).} Higher is better. Shows the utility impact of CPP at the aggregate level.}
  \label{fig:cpp-fairness-balacc}
\end{figure}

\begin{figure}[H]
\vspace{6mm}
  \centering
  \includegraphics[width=\columnwidth]{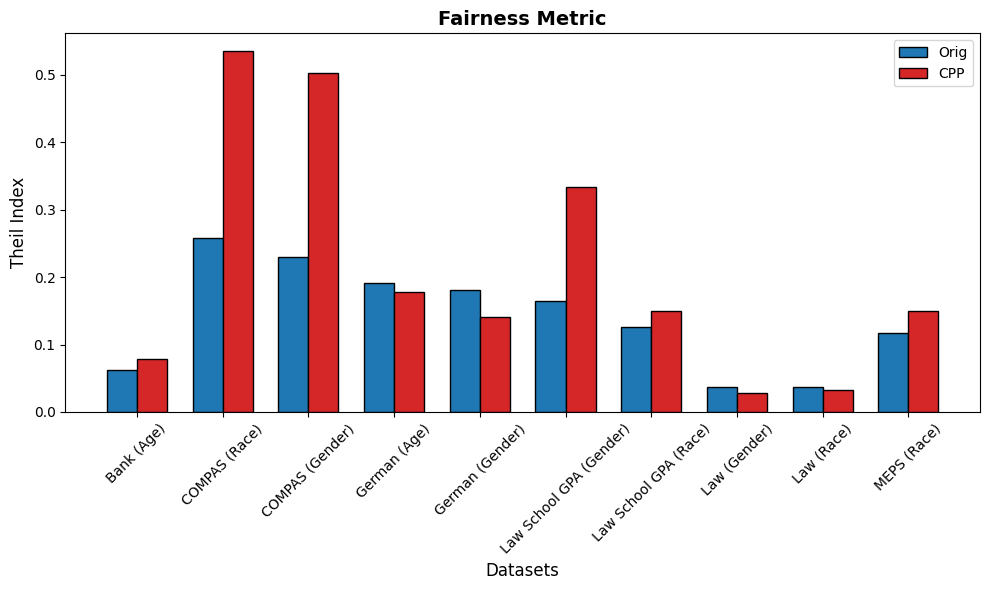}
  \caption{\textbf{Theil Index.} Lower is better; measures inequality in predicted outcomes across groups.}
  \label{fig:cpp-fairness-theil}
\end{figure}

\clearpage
\FloatBarrier
\subsection{CPP Utility Figures (Subpopulation Test Accuracies) under OTA setup}

\begin{figure}[H]
  \centering
  \includegraphics[width=\columnwidth]{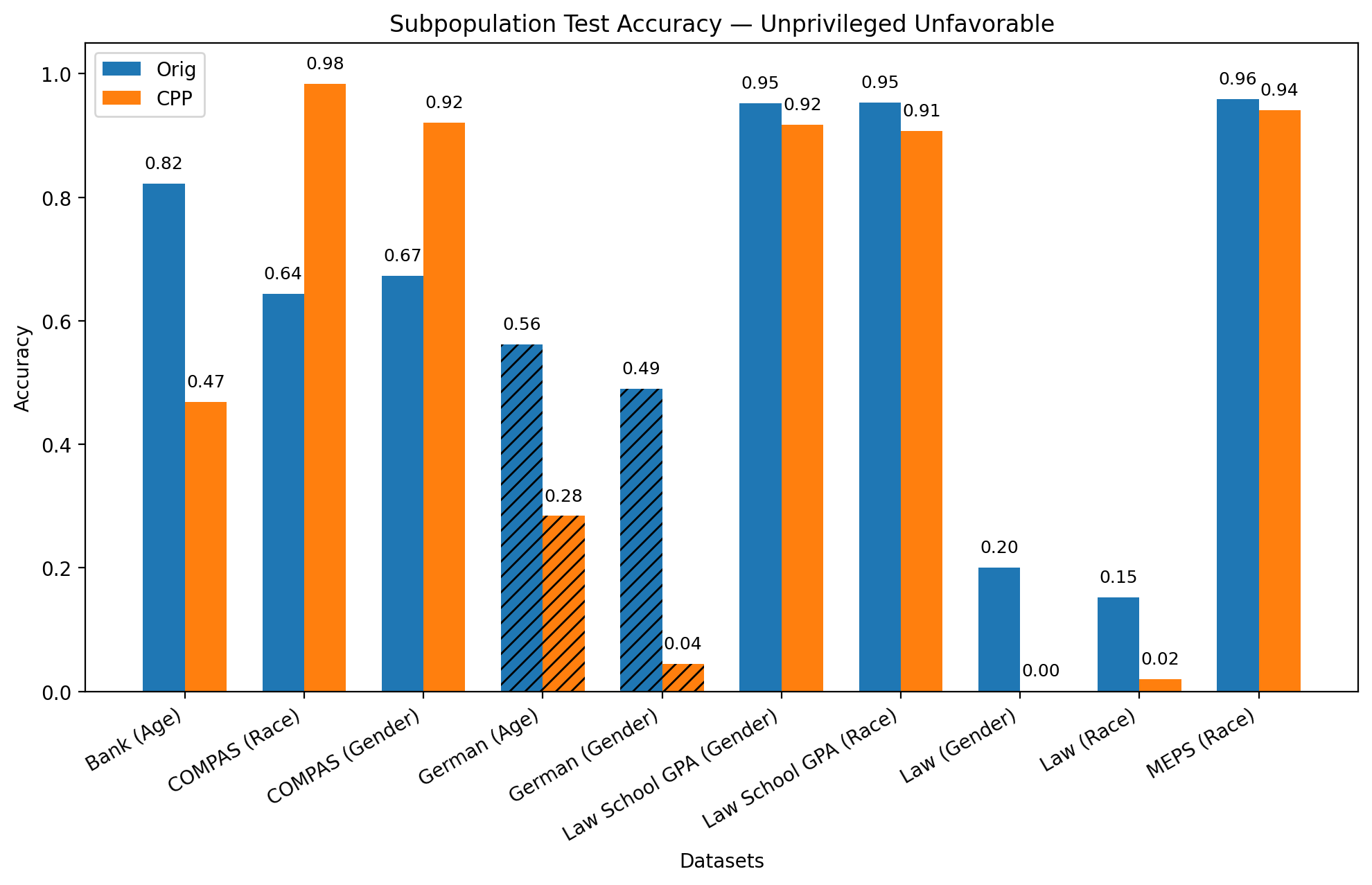}
  \caption{\textbf{Subpopulation: Unprivileged Unfavorable.} Test accuracies by dataset for Orig vs.\ CPP. Hatched bars mark datasets where this subpopulation is underrepresented.}
  \label{fig:cpp-util-0minus}
\end{figure}

\begin{figure}[H]
  \centering
  \includegraphics[width=\columnwidth]{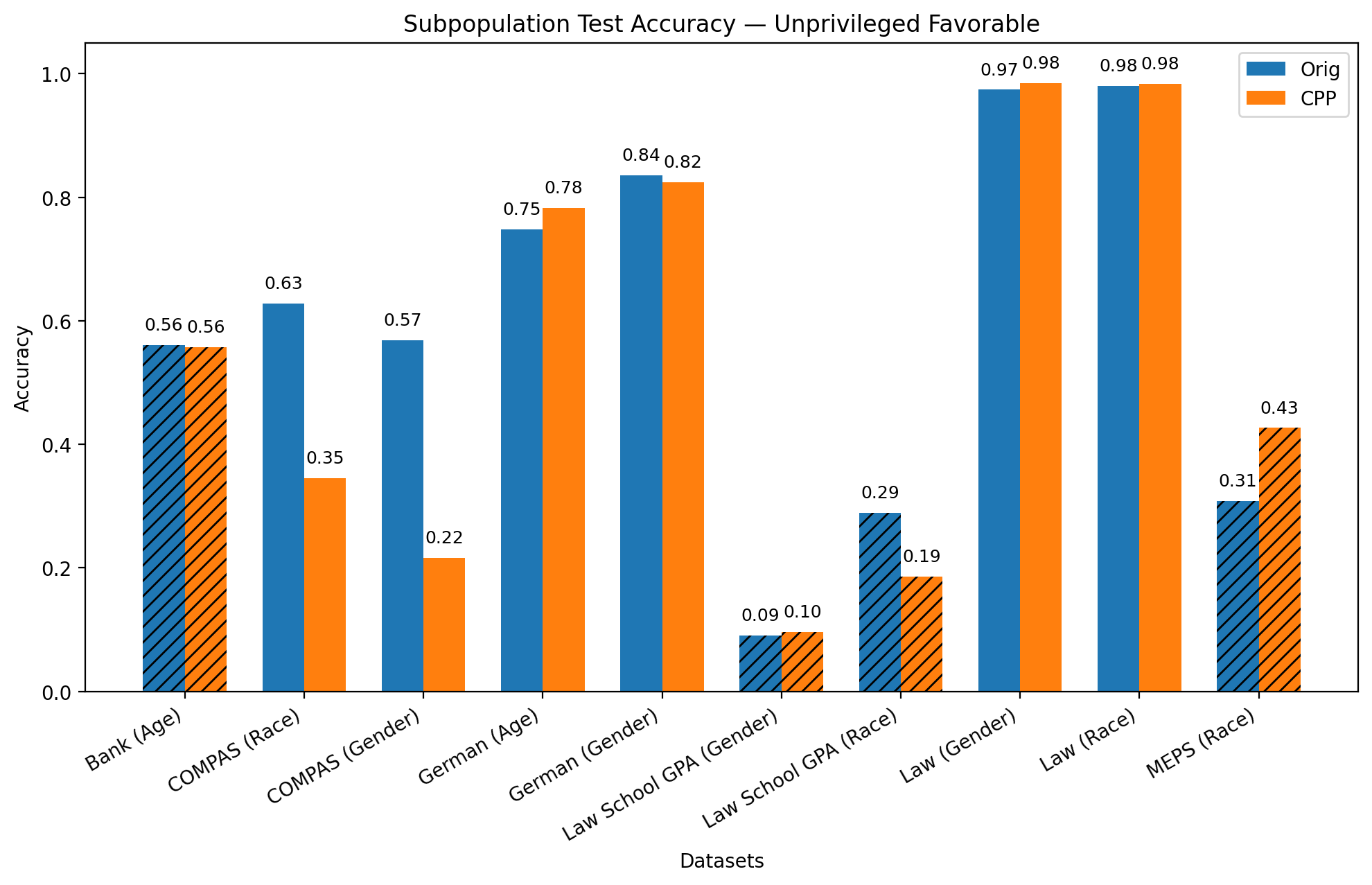}
  \caption{\textbf{Subpopulation: Unprivileged Favorable.} Test accuracies by dataset for Orig vs.\ CPP. Hatched bars mark datasets where this subpopulation is underrepresented.}
  \label{fig:cpp-util-0plus}
\end{figure}

\begin{figure}[H]
  \centering
  \includegraphics[width=\columnwidth]{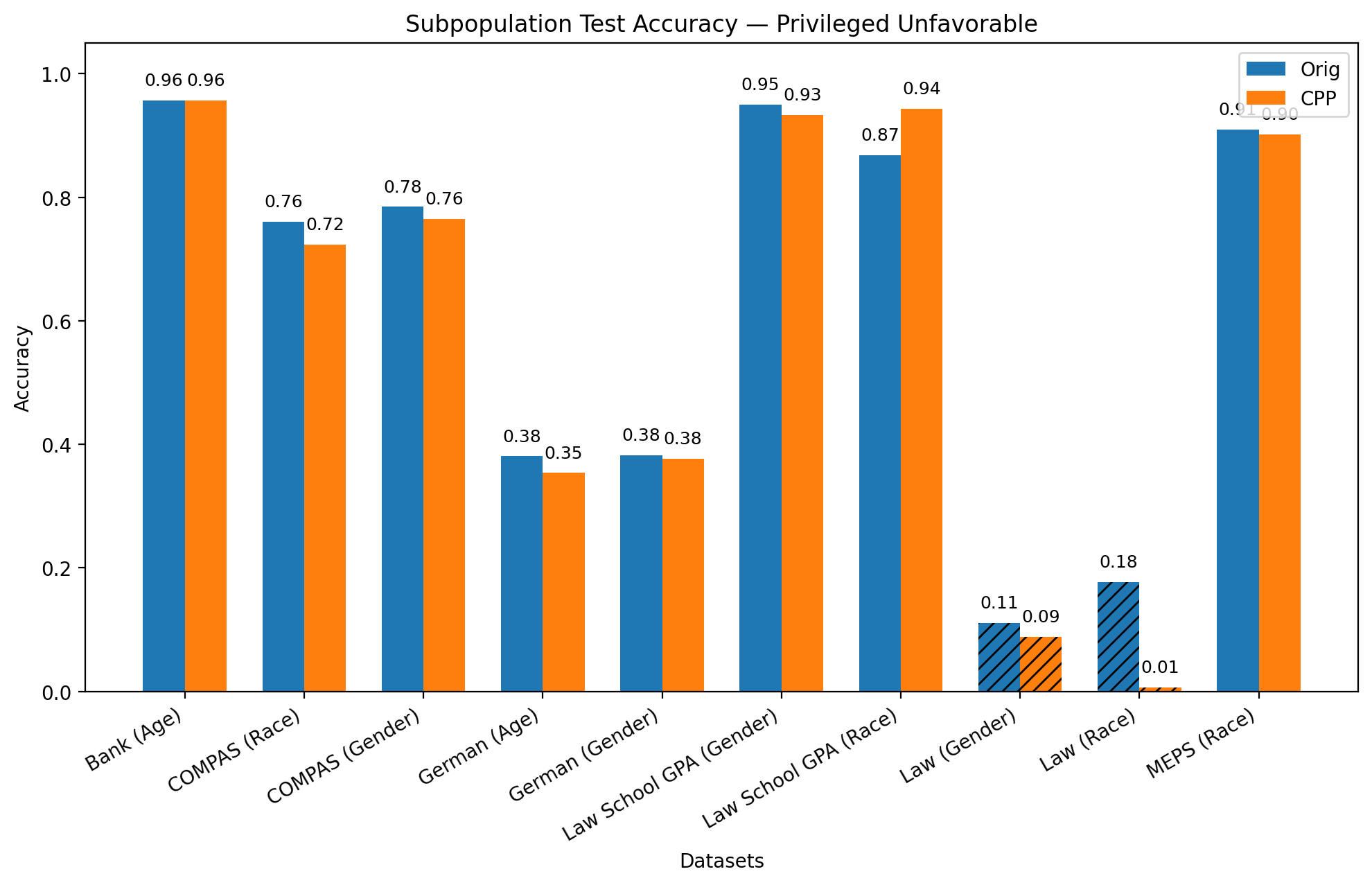}
  \caption{\textbf{Subpopulation: Privileged Unfavorable.} Test accuracies by dataset for Orig vs.\ CPP. Hatched bars mark datasets where this subpopulation is underrepresented.}
  \label{fig:cpp-util-1minus}
\end{figure}

\begin{figure}[H]
\vspace{16mm}
  \centering
  \includegraphics[width=\columnwidth]{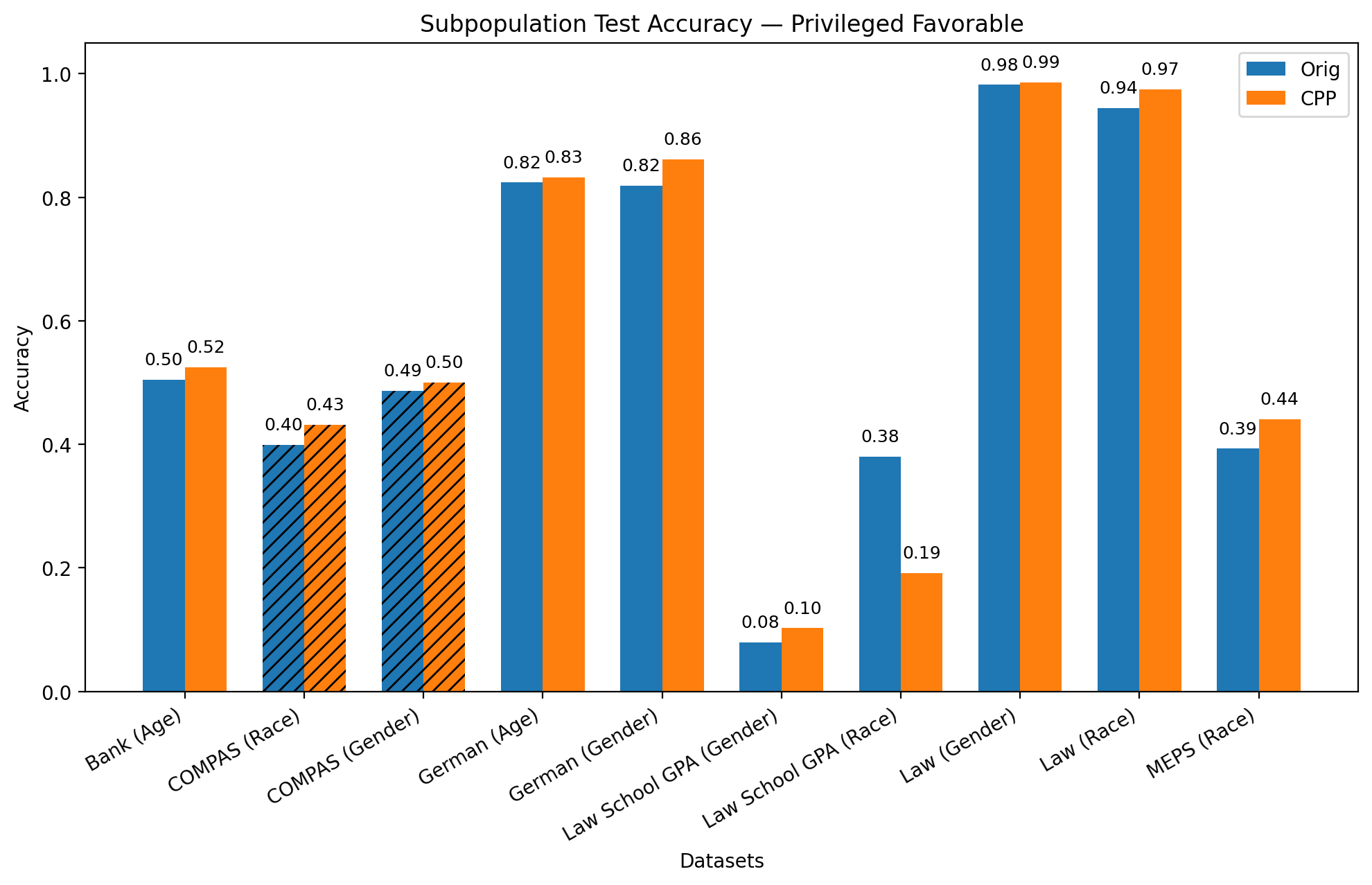}
  \caption{\textbf{Subpopulation: Privileged Favorable.} Test accuracies by dataset for Orig vs.\ CPP. Hatched bars mark datasets where this subpopulation is underrepresented.}
  \label{fig:cpp-util-1plus}
\end{figure}
\newpage
\clearpage
\section{OQTA Algorithm Details}
\label{appendix:oqta_algorithm}

\subsection{Algorithm}
The full pseudocode for OQTA is shown in Algorithm~\ref{alg:oqta}.
\begin{figure}[htbp]
\centering
\caption{Optimized Quantile Threshold Attack}
\label{alg:oqta} 
\begin{algorithmic}[1]
    \State \textbf{Input:} Target model \(A\), target dataset \(\mathcal{D}_{\text{target}}\), population dataset \(\mathcal{D}_{\text{population}}\), quantiles \(\mathbf{Q}\) (e.g., \texttt{logspace}(-5,0,100))
    \State \textbf{Output:} Privacy risk metrics for all subgroups
    \State \textbf{Split} \(\mathcal{D}_{\text{target}}\) into 
           \(\texttt{TARGET\_MEMBER}\) (train subset) 
           and \(\texttt{TARGET\_NON\_MEMBER}\) (test subset). \label{alg:split-data}
    \State \textbf{Use} \(\mathcal{D}_{\text{population}}\) as \texttt{REFERENCE\_MEMBER}. \label{alg:reference-member}
    \State \textbf{Extract} subgroup labels \(g[i]\) based on sensitive attributes and true labels. \label{alg:start-signals}
    \State \textbf{Initialize} metrics for each subgroup.

    \ForAll{\(G_g^y\) in subgroups}
        \State \textbf{Compute loss distribution from population data:}
        \Statex \quad 
          \(\mathrm{LD}_{g,y}^{(\mathrm{pop})} 
            \gets \{\,\ell(A,z)\mid z \in \mathcal{D}_{\text{population}},\,z \in G_g^y\}\)
        \State \textbf{Compute loss distribution from target data:}
        \Statex \quad 
          \(\mathrm{LD}_{g,y}^{(\mathrm{target})} 
            \gets \{\,\ell(A,z)\mid z \in (\texttt{TARGET\_MEMBER}\cup\texttt{TARGET\_NON\_MEMBER}),\,z \in G_g^y\}\) \label{alg:end-signals}
        
        \State \textbf{Initialize} \(\boldsymbol{\tau}_{g,y}\gets\emptyset\) \Comment{\textcolor{blue}{\emph{List of candidate thresholds}}} \label{alg:start-thresholds}
        \ForAll{\(\alpha\in \mathbf{Q}\)}
            \State \(\tau \gets \texttt{quantile\_threshold}\bigl(\mathrm{LD}_{g,y}^{(\mathrm{pop})},\,\alpha\bigr)\)
            \State \(\boldsymbol{\tau}_{g,y}\gets \boldsymbol{\tau}_{g,y}\cup\{\tau\}\)
        \EndFor
        
        \State \textbf{Initialize} \(\textit{best\_accuracy} \gets 0\);
               \(\tau^{(g,y)} \gets \emptyset\)
        \ForAll{\(\tau \in \boldsymbol{\tau}_{g,y}\)}
            \State \textbf{Compute accuracy on} \(\mathrm{LD}_{g,y}^{(\mathrm{target})}\)
            \If{\(\mathrm{accuracy} > \textit{best\_accuracy}\)}
                \State \(\textit{best\_accuracy} \gets \mathrm{accuracy}\)
                \State \(\tau^{(g,y)} \gets \tau\)
            \EndIf
        \EndFor
        
        \State \Comment{\textcolor{blue}{\emph{Threshold derived from population data, 
               selected via target data’s best accuracy}}} 
    \EndFor \label{alg:end-thresholds}

    \State \textbf{Perform membership inference:} \label{alg:start-mia}
    \ForAll{\(z=(x,g,y)\)\, in \,\(\texttt{TARGET\_MEMBER} \cup \texttt{TARGET\_NON\_MEMBER}\)}
        \State \(\ell_z \gets \ell(A,z)\)
        \State \(\mathcal{A}(z)\gets 
        \begin{cases}
          1, & \text{if } \ell_z < \tau^{(g,y)},\\
          0, & \text{otherwise}
        \end{cases}\)
    \EndFor
    
    \ForAll{\(G_g^y\) in subgroups}
        \State \textbf{Compute} Accuracy and privacy risk (PR) for \(G_g^y\)
    \EndFor
    
    \State \Return Aggregated privacy metrics for all subgroups \label{alg:end-mia}
\end{algorithmic}
\end{figure}

\newpage 
\subsection{Subgroup Encodings}
Group indices \(g[i]\) are computed using:
\[
g[i] = y_{\text{true}}[i] + (s_{\text{features}}[i] + 1) \cdot 2
\]
where \(s_{\text{features}}[i]\) indicates whether the sample is from the privileged (1) or unprivileged (0) group, and \(y_{\text{true}}[i]\) is the ground truth label. This results in four subgroups:
\begin{itemize}
    \item \(G_0^-\): Unprivileged, Unfavorable (\(g=2\))
    \item \(G_0^+\): Unprivileged, Favorable (\(g=3\))
    \item \(G_1^-\): Privileged, Unfavorable (\(g=4\))
    \item \(G_1^+\): Privileged, Favorable (\(g=5\))
\end{itemize}

\subsection{Threshold Computation Details}
Thresholds are computed from the population loss distribution using log-spaced quantiles:
\[
\text{Quantiles} = logspace(-5, 0, 100)
\]
where \(logspace(a=-5, b=0, n=100)\) generates \(n\) values evenly spaced between \(10^a\) and \(10^b\) on a logarithmic scale.

Linear interpolation is applied between cumulative distribution values to compute each candidate threshold \(\tau\). The final \(\tau^{(g,y)}\) is the threshold that achieves the highest balanced accuracy when applied to the target set.

\newpage
\clearpage
\section{Subpopulation Utility Under Fairness Mitigators}
\label{app:subpop-utility}
\setlength{\floatsep}{14pt plus 2pt minus 2pt}
\setlength{\textfloatsep}{14pt plus 2pt minus 2pt}

\begin{table}[H]
\centering
\caption{LiRA (DT): subpopulation test accuracy under fairness mitigators.}
\label{tab:lira-dt-subgroup-acc}
\small
\setlength{\tabcolsep}{4pt}
\begin{tabular}{llccccc}
\toprule
Dataset & Subgroup & Orig & SYN & DIR & REW & EGR \\
\midrule
\multirow{4}{*}{Synthetic (G)}
 & $G_0^-$ & 0.33 & 0.50 & 0.26 & 0.35 & 0.45 \\
 & $G_0^+$ & 0.94 & 0.77 & 0.95 & 0.91 & 0.88 \\
 & $G_1^-$ & 0.88 & 0.89 & 0.89 & 0.85 & 0.83 \\
 & $G_1^+$ & 0.87 & 0.86 & 0.87 & 0.88 & 0.87 \\
\midrule
\multirow{4}{*}{Bank (Age)}
 & $G_0^-$ & 0.85 & 0.90 & 0.82 & 0.89 & 0.88 \\
 & $G_0^+$ & 0.50 & 0.33 & 0.51 & 0.34 & 0.41 \\
 & $G_1^-$ & 0.96 & 0.96 & 0.96 & 0.96 & 0.96 \\
 & $G_1^+$ & 0.46 & 0.44 & 0.45 & 0.45 & 0.46 \\
\midrule
\multirow{4}{*}{COMPAS (Race)}
 & $G_0^-$ & 0.65 & 0.67 & 0.70 & 0.72 & 0.66 \\
 & $G_0^+$ & 0.62 & 0.59 & 0.58 & 0.55 & 0.57 \\
 & $G_1^-$ & 0.76 & 0.76 & 0.77 & 0.72 & 0.73 \\
 & $G_1^+$ & 0.44 & 0.44 & 0.42 & 0.49 & 0.46 \\
\midrule
\multirow{4}{*}{MEPS (Race)}
 & $G_0^-$ & 0.96 & 0.95 & 0.96 & 0.94 & 0.95 \\
 & $G_0^+$ & 0.34 & 0.33 & 0.33 & 0.38 & 0.35 \\
 & $G_1^-$ & 0.91 & 0.91 & 0.90 & 0.91 & 0.91 \\
 & $G_1^+$ & 0.40 & 0.40 & 0.42 & 0.40 & 0.38 \\
\bottomrule
\end{tabular}
\end{table}

\begin{table}[H]
\centering
\caption{OQTA (DT): subpopulation test accuracy under fairness mitigators.}
\label{tab:oqta-dt-subgroup-acc}
\small
\setlength{\tabcolsep}{4pt}
\begin{tabular}{llccccc}
\toprule
Dataset & Subgroup & Orig & SYN & DIR & REW & EGR \\
\midrule
\multirow{4}{*}{Synthetic (G)}
 & $G_0^-$ & 0.20 & 0.52 & 0.06 & 0.54 & 0.42 \\
 & $G_0^+$ & 0.93 & 0.66 & 0.99 & 0.80 & 0.79 \\
 & $G_1^-$ & 0.65 & 0.66 & 0.64 & 0.56 & 0.58 \\
 & $G_1^+$ & 0.68 & 0.65 & 0.71 & 0.77 & 0.75 \\
\midrule
\multirow{4}{*}{Bank (Age)}
 & $G_0^-$ & 0.84 & 0.83 & 0.84 & 0.89 & 0.87 \\
 & $G_0^+$ & 0.54 & 0.46 & 0.47 & 0.44 & 0.48 \\
 & $G_1^-$ & 0.95 & 0.96 & 0.95 & 0.95 & 0.95 \\
 & $G_1^+$ & 0.47 & 0.44 & 0.47 & 0.50 & 0.48 \\
\midrule
\multirow{4}{*}{COMPAS (Race)}
 & $G_0^-$ & 0.65 & 0.65 & 0.67 & 0.65 & 0.66 \\
 & $G_0^+$ & 0.58 & 0.58 & 0.55 & 0.58 & 0.56 \\
 & $G_1^-$ & 0.72 & 0.72 & 0.68 & 0.67 & 0.66 \\
 & $G_1^+$ & 0.46 & 0.46 & 0.47 & 0.49 & 0.50 \\
\midrule
\multirow{4}{*}{MEPS (Race)}
 & $G_0^-$ & 0.94 & 0.95 & 0.94 & 0.93 & 0.93 \\
 & $G_0^+$ & 0.34 & 0.31 & 0.32 & 0.35 & 0.35 \\
 & $G_1^-$ & 0.89 & 0.91 & 0.88 & 0.89 & 0.89 \\
 & $G_1^+$ & 0.40 & 0.35 & 0.41 & 0.38 & 0.37 \\
\bottomrule
\end{tabular}
\end{table}

\begin{table}[H]
\centering
\caption{OTA (DT): subpopulation test accuracy under fairness mitigators.}
\label{tab:ota-dt-subgroup-acc}
\small
\setlength{\tabcolsep}{4pt}
\begin{tabular}{llcccccc}
\toprule
Dataset & Subgroup & Orig & SYN & DIR & REW & EGR & CPP \\
\midrule
\multirow{4}{*}{Synthetic (G)}
 & $G_0^-$ & 0.34 & 0.47 & 0.32 & 0.36 & 0.54 & 0.05 \\
 & $G_0^+$ & 0.93 & 0.74 & 0.93 & 0.90 & 0.84 & 0.95 \\
 & $G_1^-$ & 0.88 & 0.89 & 0.88 & 0.87 & 0.87 & 0.05 \\
 & $G_1^+$ & 0.87 & 0.86 & 0.87 & 0.87 & 0.86 & 0.95 \\
\midrule
\multirow{4}{*}{Bank (Age)}
 & $G_0^-$ & 0.82 & 0.90 & 0.79 & 0.89 & 0.89 & 0.47 \\
 & $G_0^+$ & 0.58 & 0.35 & 0.56 & 0.43 & 0.46 & 0.56 \\
 & $G_1^-$ & 0.95 & 0.96 & 0.93 & 0.95 & 0.96 & 0.96 \\
 & $G_1^+$ & 0.51 & 0.51 & 0.53 & 0.57 & 0.51 & 0.52 \\
\midrule
\multirow{4}{*}{COMPAS (Race)}
 & $G_0^-$ & 0.69 & 0.69 & 0.67 & 0.71 & 0.70 & 0.98 \\
 & $G_0^+$ & 0.60 & 0.60 & 0.60 & 0.57 & 0.60 & 0.35 \\
 & $G_1^-$ & 0.77 & 0.77 & 0.74 & 0.69 & 0.70 & 0.72 \\
 & $G_1^+$ & 0.44 & 0.43 & 0.46 & 0.51 & 0.53 & 0.43 \\
\midrule
\multirow{4}{*}{MEPS (Race)}
 & $G_0^-$ & 0.95 & 0.95 & 0.95 & 0.94 & 0.94 & 0.94 \\
 & $G_0^+$ & 0.35 & 0.33 & 0.35 & 0.38 & 0.37 & 0.43 \\
 & $G_1^-$ & 0.89 & 0.91 & 0.89 & 0.90 & 0.92 & 0.90 \\
 & $G_1^+$ & 0.43 & 0.40 & 0.44 & 0.40 & 0.37 & 0.44 \\
\bottomrule
\end{tabular}
\end{table}

\end{document}